\documentclass[10pt,twocolumn]{article}
\usepackage[letterpaper,portrait,margin=1in,columnsep=10pt]{geometry}
\usepackage{graphicx,amsmath,amssymb,booktabs,tabularx,array}
\usepackage[hyphens]{url}

\usepackage{microtype}
\usepackage{placeins}
\usepackage{caption}
\usepackage{titlesec}
\titlespacing*{\section}{0pt}{2ex plus .5ex minus .2ex}{.8ex plus .2ex}
\titlespacing*{\subsection}{0pt}{1.5ex plus .5ex minus .2ex}{.6ex plus .2ex}
\usepackage{float,needspace,etoolbox,stfloats}
\usepackage[colorlinks=true,urlcolor=blue,linkcolor=black]{hyperref}
\usepackage{xcolor}
\definecolor{paperblue}{HTML}{355C82}
\hypersetup{urlcolor=paperblue,pdfpagelayout=SinglePage,pdfpagemode=UseNone,pdfstartview=Fit}

\title{\textbf{Beyond Job Titles: Interpreting Early-Career Job Advertisements with Sparse Autoencoders and Semantic Clustering}}

\author{Yisong (Ethan) Cao, Leah Kim, Brian Wright, and Ali Rivera\\Active Learning Lab\\School of Data Science, University of Virginia}

\date{}

\renewenvironment{abstract}{\section*{\abstractname}}{\par}
\begin{document}
\maketitle

\begin{abstract}
Job titles and job descriptions alone provide limited insight into how different technical roles, such as Data Scientists, Data Analysts and Software Engineers, evolve and change in the market. Traditional quantitative methods such as skill counting or recording matched keywords in large textual datasets do not fully explain these differences. 

To address this, we design and compare two methods based on deep learning and transformer representations to discover insights from large-scale textual data: 
\begin{itemize}
    \item \textbf{An embedding-based clustering analysis of semantic vectors with respect to occupation type and month of posting}
    \item \textbf{A pretrained sparse autoencoder (SAE) for feature extraction from the same set of clustered semantic vectors}
\end{itemize}
We use 36,081 early-career job postings scraped from LinkedIn between December 2025 and October 2026. During preprocessing, we apply GPT-6 Luna to classify passages into Team context, Role task and Qualifications without rewriting the source text corpus for each job description. We then examine seven individual and combined segmentations of those passages through \textbf{PERC}, a workflow comprising \textbf{Preprocessing, Embedding, Reduction and Clustering}. To examine occupational and monthly differences, we adopt Maximum Mean Discrepancy (MMD) to rank differences between embedding distributions and ask GPT-6.1 Terra to explain the selected cluster comparisons using complete advertisements. 

On the other hand, we also use a pretrained Goodfire sparse autoencoder to extract features from layer-19 representations of Llama-3.1-8B-Instruct. To compare what these representations contribute, GPT-6 Luna generates hypotheses from the same descriptions using PERC clusters, SAE-extracted features or text alone, and we subsequently employ GPT-6.1 Terra to check the resulting claims in advertisements from held-out companies. 

In our occupational difference comparison analysis, we observe significant differences across the roles of Data Scientist, Data Analyst, Machine Learning Engineer, Business Analyst and Software Engineer in the feature subjects extracted through Llama-8B's feature space, such as \textit{``developing machine learning models,'' ``data reporting,'' ``AI-system implementation,'' ``software development'' and ``business analysis''}. In our monthly distributional analysis, we further observe how Data Scientist and Business Analyst roles exhibit different responsibilities and tasks in the descriptions.

Although these monthly comparison findings could be limited by the scale of our dataset and by outages of our scraping machine, which led to missing data between May and August 2026, the methodologies we examine could be applied to larger-scale datasets in a robust fashion. 
\end{abstract}

\section{Introduction}
\label{sec:intro}
What do job titles and job descriptions tell us at first glance? Do they inform us about the distinct nature of a role or what we should expect from similar roles across the job market?

With job postings arriving daily, if not hourly, this pool of messy and unstructured text makes it difficult to assess differences in the work being advertised. Amid the changing job market, a natural question for early-career job seekers is: what roles are changing with the advent of AI, and what tools can help us examine the work, the teams and the qualifications that employers describe?

Intuitively, we could track how often a skill is mentioned or how many positions are posted. However, descriptions can share keywords and still refer to different work. Python might be used to develop ``machine learning models'', maintain an application or automate a business report. Skill counts alone do not show how the tools are used.

To examine this context, we first apply \textbf{PERC}, short for \textbf{Preprocessing, Embedding, Reduction and Clustering}. This framework converts descriptions into vectors and groups postings with similar meaning. We then use Maximum Mean Discrepancy (MMD) to identify large distributional differences, and GPT-6.1 Terra reads the full descriptions from the selected clusters to explain what differs in their work, team context and qualifications.

Meanwhile, we apply a pretrained 537-million-parameter \textbf{sparse autoencoder (SAE)} from Goodfire to the layer-19 representations of Llama-3.1-8B-Instruct. The SAE extracts features that can activate together within a description. GPT-6 Luna uses those features to propose differences between occupations or months, and GPT-6.1 Terra checks whether the proposed characteristics also appear in postings from held-out companies.

Together, these methods let us ask what grouping descriptions and examining their features each reveal, how their findings compare with GPT-6 Luna reading the text without either representation, and whether they identify occupational or monthly differences. We examine \textbf{Data Science} (DS), \textbf{Data Analytics} (DA), \textbf{Software Engineering} (SWE), \textbf{Machine Learning Engineering} (MLE) and \textbf{Business Analysis} (BA) using early-career job postings collected through our hourly scraper from December 2025 to October 2026.

\section{Literature Review}
The point of this research is to examine whether methods that group job descriptions or identify features within them can help us understand the work underneath a job title. Namely, if two companies are both hiring a data scientist, we want to examine whether those positions involve similar responsibilities, serve different applications, or change in what they describe from month to month.

Prior studies use NLP tools to extract skills, cluster job postings and describe the roles they find. We draw from these studies with particular attention to the text each method examines and the insights it can produce.

\subsection{Precedents in Textual Analytics and Topic Modeling}
Lukauskas et al. \hyperlink{ref-1}{[1]} present a methodology closely related to PERC, examining more than 500,000 Lithuanian job postings to construct groups of similar jobs. They extract requirements, represent the text with sentence embeddings and compare reduction and clustering methods. GPT-4 then describes these groups from their keywords and human experts assess the descriptions.

We adopt a similar sequence and examine seven segmentation text inputs formed from Team context, Role task and Qualifications. Separating these passages lets us distinguish what the employer requests from what the applicant would do. For example, an ad can request a statistics degree without asking the applicant to build statistical models. For the final cluster comparisons, we give GPT-6.1 Terra every full job description in the selected groups, allowing it to inspect applications and exceptions beyond a list of keywords.

Other studies examine several kinds of work beneath the data-science title. Michalczyk et al. \hyperlink{ref-2}{[2]} combine topic modeling, clustering and expert assessment of 25,104 postings to identify six roles. Gunklach et al. \hyperlink{ref-3}{[3]} use BERTopic and K-medoids on 16,348 unique postings, representing each through its topic distribution and identifying nine roles through manual assessment.

\subsection{Precedents in Skill Theme Extraction}
We also examine how embeddings and topic models help compare requested skills, given that Qualifications is one of our segmentation text inputs. Skill Scanner \hyperlink{ref-4}{[4]} applies Sentence-BERT, UMAP, DBSCAN and K-means to compare skills across job postings, CVs and a data-science curriculum. Gurcan et al. \hyperlink{ref-5}{[5]} use a MiniLM-based BERTopic model to examine skill themes in 3,156 big-data postings across Analyst, Engineer, Developer and Architect categories. Their categories come from job-title terms, so the BERTopic analysis examines differences within already-defined occupations. Our occupation labels serve a similar purpose.

On top of identifying skills, researchers examine how to connect different expressions to a common category. SkiLLens \hyperlink{ref-6}{[6]} uses embeddings and language models to map candidate skills to ESCO, the European classification of skills and occupations, with human review supporting multilingual classification. TaxonomyBuilder \hyperlink{ref-7}{[7]} examines how filtering postings affects AI-skill and task taxonomies produced through clustering and language-model labeling. The reported improvement in domain coverage illustrates how filtering can change the resulting taxonomy. In our study, we examine which text to include within each advertisement by comparing Team context, Role task and Qualifications separately and in combination.

Sentence-BERT \hyperlink{ref-12}{[12]} provides a foundation for comparing sentences with similar meaning even when their words differ. Colombo et al.'s VEUCTOR \hyperlink{ref-8}{[8]} examines training, selecting and aligning representations using 4.5 million postings across 28 languages, finding that embedding model choices affect occupational skill groupings. We examine that issue through joint configuration searches over embedding, reduction and clustering methods.

\subsection{Precedents in Occupation and Monthly Comparisons}
Kauttonen et al. \hyperlink{ref-9}{[9]} compare 107,335 ICT job postings with 18,254 thesis abstracts over twelve years using subject indexing, TF-IDF, truncated SVD and K-means, followed by UMAP visualization. Their study distinguishes the representation used for grouping from the projection used to display it. We make the same distinction: two-dimensional maps show part of the vector variation, and MMD uses the original embeddings.

Labussière and Bol \hyperlink{ref-10}{[10]} use topic models of extracted skills and MMD to examine whether occupation labels describe the skill content of individual jobs. Their findings include differences within occupations and similarities across occupation boundaries. This provides a direct precedent for applying MMD to labor-market data.

We adopt that strategy to compare collections of job vectors, such as DS task descriptions from two posting months. Gretton et al. \hyperlink{ref-11}{[11]} provide the kernel two-sample framework underlying MMD. We use the score to select large observed differences for inspection, then ask GPT-6.1 Terra to explain them from the source descriptions.

\subsection{Precedents in Adopting SAEs for Data Analysis}
Cho et al. \hyperlink{ref-13}{[13]} already apply SAEs to labor-market text, examining the internal representations supporting contextual skill substitution in a masked language model. They examine whether a skill can replace another in a particular work context, whereas we examine which characteristics distinguish occupations or months.

To examine those differences, we adopt the dataset-comparison procedure introduced by Jiang et al. \hyperlink{ref-14}{[14]}. The authors rank features by activation-frequency differences, interpret them using examples and combine related features into hypotheses about properties that differ between datasets. A language-model judge then checks those hypotheses against documents.

For example, a feature might activate on passages about developing ``machine learning models''. A hypothesis defines which activities count as ``developing machine learning models'' and proposes that they occur more often in one of the two groups of advertisements. GPT-6.1 Terra checks that definition in job descriptions from companies excluded from hypothesis generation, returning present, absent or unclear with supporting quotes. The percentage of advertisements in which Terra identifies the characteristic is therefore distinct from the percentage in which the SAE feature activates.

Jiang et al. use Llama-3.3-70B and a Goodfire SAE at layer 50 in their main experiments. We use Llama-3.1-8B-Instruct with its layer-19 Goodfire SAE following the toolkit's configuration for the 8B-parameter Llama model. We also examine the full dataset with Gemma 2 2B and Gemma Scope \hyperlink{ref-15}{[15]}. This separate analysis reveals sensitivity to description length and formatting, motivating those checks in the comparison.

\subsection{The Contribution of This Research}
With these precedents, we combine segmentation, PERC, MMD ranking and language-model inspection into a reproducible analysis of early-career technical jobs. We compare that workflow with pretrained SAE analysis and GPT-6 Luna on the same selected descriptions, then examine what the resulting findings explain about DS, DA, MLE, BA and SWE work.

We compare which distinctions the methods share and which additional characteristics they propose, then examine how formatting of the original descriptions, repeated recruitment and small monthly samples affect the findings. We also connect the representations by examining SAE feature activations within PERC clusters selected for comparison using MMD.

\section{Methodology}
\label{sec:methodology}
To examine what the job postings tell us about the nature of the work, we first apply PERC to group descriptions with similar meaning and then apply Maximum Mean Discrepancy (MMD) to identify which distributions of job description vectors differ the most. After this, we employ GPT-6.1 Terra to read every full job description from the selected cluster groups and explain differences in Team context, Role task and Qualifications.

Meanwhile, we apply a pretrained sparse autoencoder from Goodfire with approximately 537 million parameters and a dictionary of 65,536 features to examine the activations at layer 19 of Llama-3.1-8B-Instruct as it reads each job-ad section. We then use GPT-6 Luna to turn those features into proposed hypotheses about differences between occupations or months. Finally, GPT-6.1 Terra checks whether the same characteristics also appear in job descriptions from held-out companies.

\begin{figure*}[!tb]\centering
\includegraphics[width=\textwidth,height=.62\textheight,keepaspectratio]{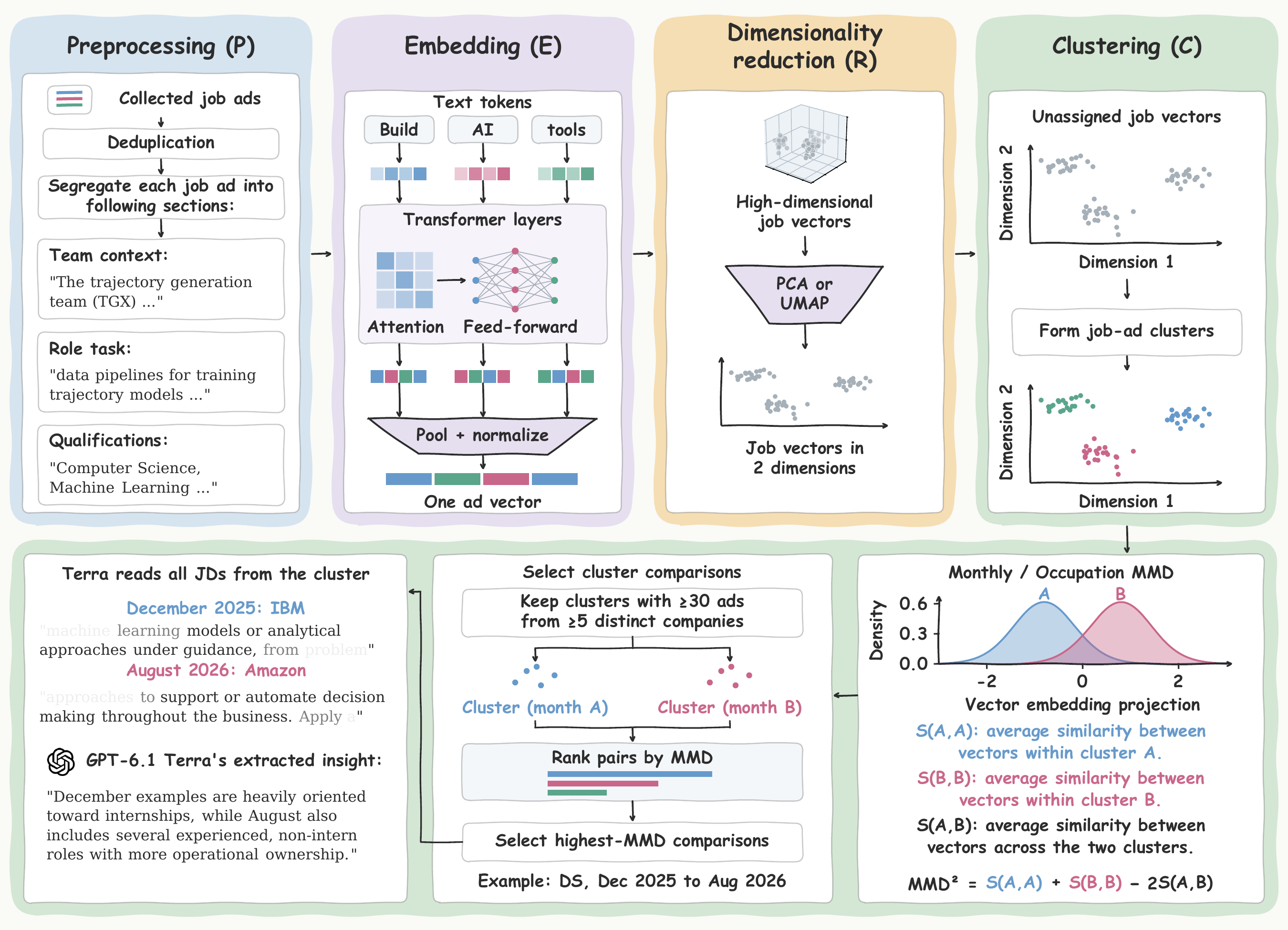}
\caption{A PERC workflow diagram that follows job advertisements from preparation to cluster comparison and language-model interpretation. The upper row demonstrates our design for how a typical job receives preprocessing, section segmentation, embedding, dimensionality reduction and clustering. The lower row compares two cluster distributions and ranks their MMD scores before GPT-6.1 Terra reads the selected advertisements. The narrative example shown in the figure contrasts December 2025 descriptions of more \textit{``supervised work''} with August 2026 descriptions of \textit{``greater operational responsibility''} generated via an LLM verifier at the end.}
\label{fig:clustering_pipeline}\end{figure*}

\subsection{Preprocessing (P): Preparing and Separating Job Descriptions}
We collect 41,040 records through our hourly scraper and retain 36,081 job postings after checking dates and text, removing copies posted on the same date, identifying company names and selecting early-career positions. By early career, we refer to internships, co-ops, junior and new-graduate positions, as well as descriptions requesting zero to two years of experience. The retained sample contains 35,511 internships and only 570 other early-career positions. This implies that our findings would primarily reflect the narratives of internship advertisements rather than those of full-time early-career roles. (Appendix~\ref{app:data} provides more information about the filtering rules and the number of postings retained at each step.)

Posting dates span 29 observed months from August 2023 to October 2026. Our scraper began collecting in December 2025 and could retrieve older postings that remained on the board, so we group jobs by posting date and separate older collected postings from fresh observations in the method comparison.

After this, we divide each job description into short passages and use GPT-6 Luna to classify what each passage describes. We retain the passages labeled as Team context, Role task or Qualifications and group them into the following sections without changing their original wording:
\begin{itemize}
\item \textbf{Team context:} the team that the applicant would join and the problem it addresses, such as turning scene understanding into autonomous-vehicle behavior.
\item \textbf{Role task:} what the applicant would do within that team, such as improving data pipelines used to train ``machine learning models'' for trajectory prediction.
\item \textbf{Qualifications:} the background that the employer requests, such as enrollment in a master's program in computer science, machine learning or robotics.
\end{itemize}
To examine which parts provide useful insights, we compare seven \textbf{segmentation text inputs}: the three individual sections, Team context + Role task, Team context + Qualifications, Role task + Qualifications, and all three joined together. We retain all job types during preprocessing, then use occupation names in titles or role introductions to identify DS, DA, MLE, BA and SWE for occupation comparisons.

\begin{figure*}[!tb]\centering
\includegraphics[width=\textwidth,height=.62\textheight,keepaspectratio]{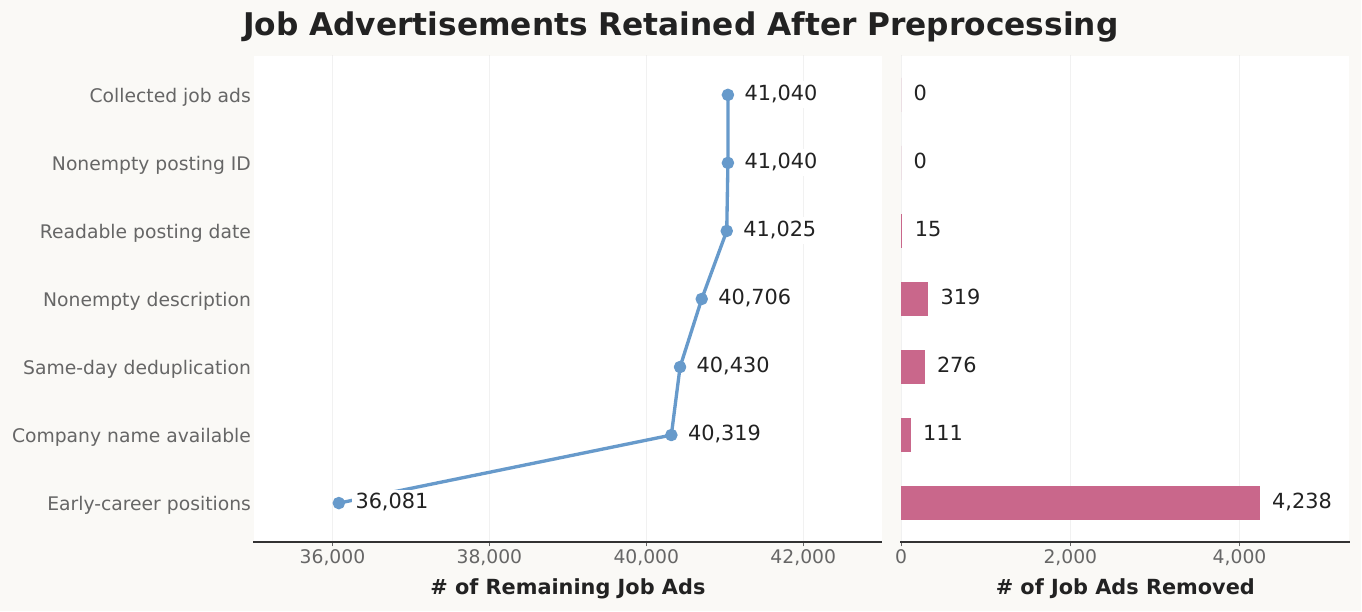}
\caption{A line chart and a horizontal bar chart that show how many job advertisements remain after each preprocessing step and how many advertisements each step removes, respectively. When we apply the preprocessing steps, the number of job advertisements falls from 41,040 to 36,081. The largest decrease occurs when we select early-career positions and filter out 4,238 advertisements, compared with 319 advertisements removed for empty descriptions and 276 advertisements removed through same-day deduplication.}
\label{fig:preprocessing_counts}\end{figure*}

\subsection{Embedding (E): Turning Job Descriptions into Vectors}
Once we have separated the descriptions, we join the passages belonging to each segmentation text input before embedding. Each posting therefore contributes one vector per available input, so four task passages become one Role task vector. We compare job vectors formed from the same input.

We examine MiniLM, MPNet, BGE, E5, EmbeddingGemma 2 and text-embedding-3-small. Text that exceeds an embedding model's input limit is embedded in chunks and combined into one vector. (Appendix~\ref{app:embeddings} provides more information about the embedding model settings and how we combine chunk embeddings into one vector.)

\begin{figure*}[!tb]\centering
\includegraphics[width=\textwidth,height=.62\textheight,keepaspectratio]{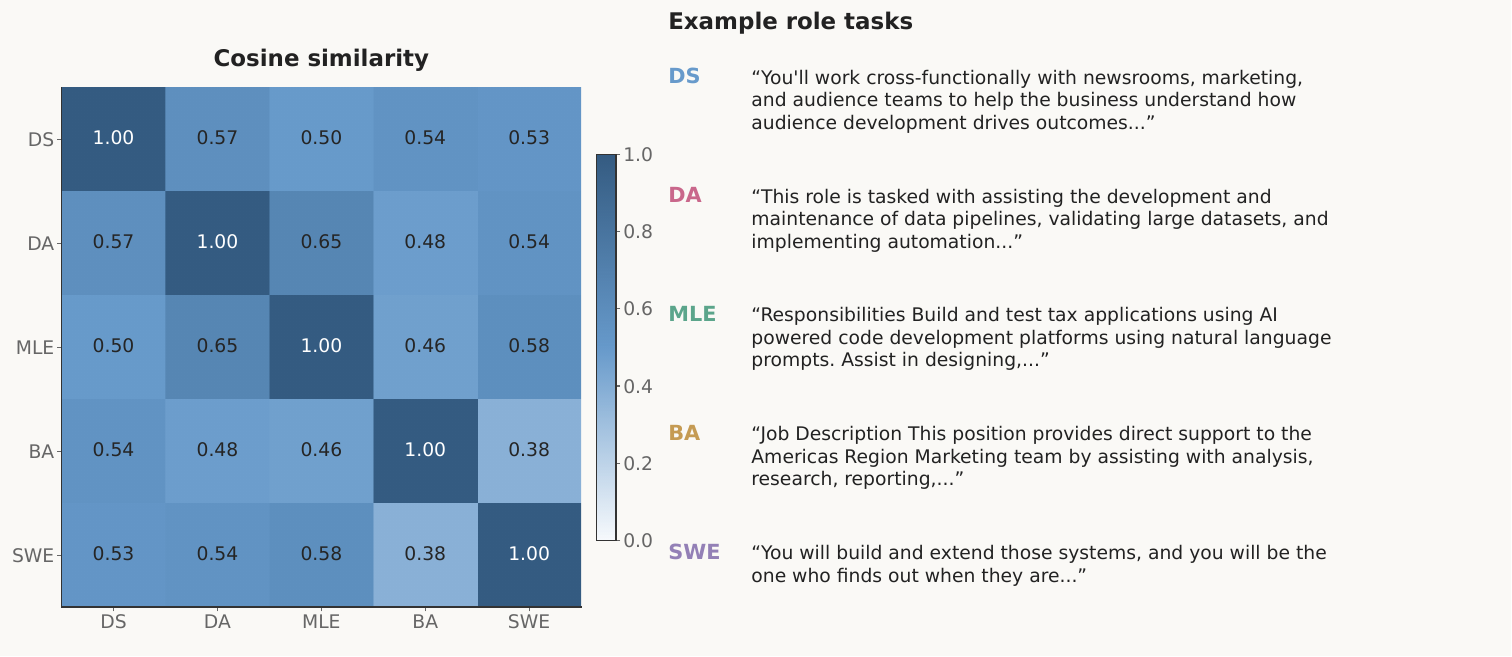}
\caption{A cosine-similarity matrix comparing five Role task advertisements embedded with text-embedding-3-small. The adjacent excerpts show the tasks described in those advertisements. Among these five Role task examples, Data Analyst and Machine Learning Engineer have the highest cosine similarity at 0.65 and Business Analyst and Software Engineer have the lowest at 0.38. The excerpts contrast data pipelines and application development with marketing analysis and reporting, showing how we can compare the work described beneath different occupation names.}
\label{fig:embedding_similarity_examples}\end{figure*}

\subsection{Reduction (R), Clustering (C) and Framework Config Grid Search}
We randomly split company names into training (60\%), validation (20\%) and test (20\%) sets. All postings from one company stay together, allowing us to examine whether a framework fitted to one set of employers also groups descriptions from held-out employers.

For each segmentation text input, we compare PCA and UMAP alongside the original vectors and then compare K-means, Gaussian mixtures, hierarchical agglomerative clustering, spectral clustering and HDBSCAN. We first require an Adjusted Rand Index (ARI) of at least 0.60 across runs and assignment of at least 60\% of validation descriptions to clusters. Every cluster must include at least ten training companies, five validation companies and ten validation advertisements, with no training company contributing more than 20\% of its descriptions. Among configurations meeting these criteria, we select the highest positive TF-IDF cosine silhouette, which measures how distinctly the clusters use words. The selected framework must use PCA or UMAP. We also report silhouette on the original embedding vectors. (Appendix~\ref{app:clustering} provides more information about the criteria for selecting the right PERC framework configuration and how we calculate each measurement.)

After evaluating the selected framework on test companies, we use all job advertisements retained after preprocessing for monthly and occupational analytics. Figure~\ref{fig:clustering_pipeline} shows the complete PERC workflow.

\begin{figure*}[!tb]\centering
\includegraphics[width=\textwidth,height=.62\textheight,keepaspectratio]{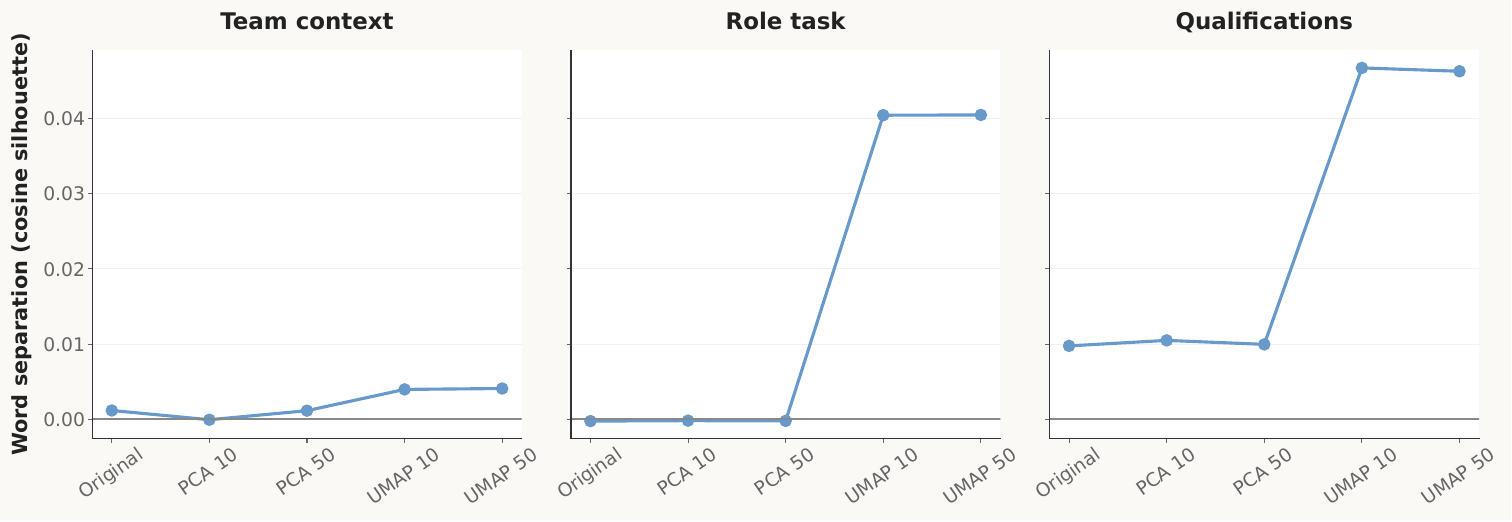}
\caption{A set of line charts comparing TF-IDF cosine silhouette for the original MPNet vectors, PCA and UMAP across three job-ad sections. Each setting uses four K-means clusters. With UMAP, the Role task cosine silhouette rises from approximately zero with the original vectors or PCA to 0.0405. For Qualifications, UMAP reaches about 0.047 compared with about 0.010 for PCA and the original vectors, whereas Team context reaches only about 0.004. Increasing UMAP from 10 to 50 dimensions makes little difference in any panel and PCA provides little improvement over the original vectors. The higher UMAP scores indicate more similar word use within clusters relative to word use between clusters, with the largest observed increases in cosine silhouette for Role task and Qualifications.}
\label{fig:reduction_comparison_sections}\end{figure*}

\begin{figure*}[!tb]\centering
\includegraphics[width=\textwidth,height=.62\textheight,keepaspectratio]{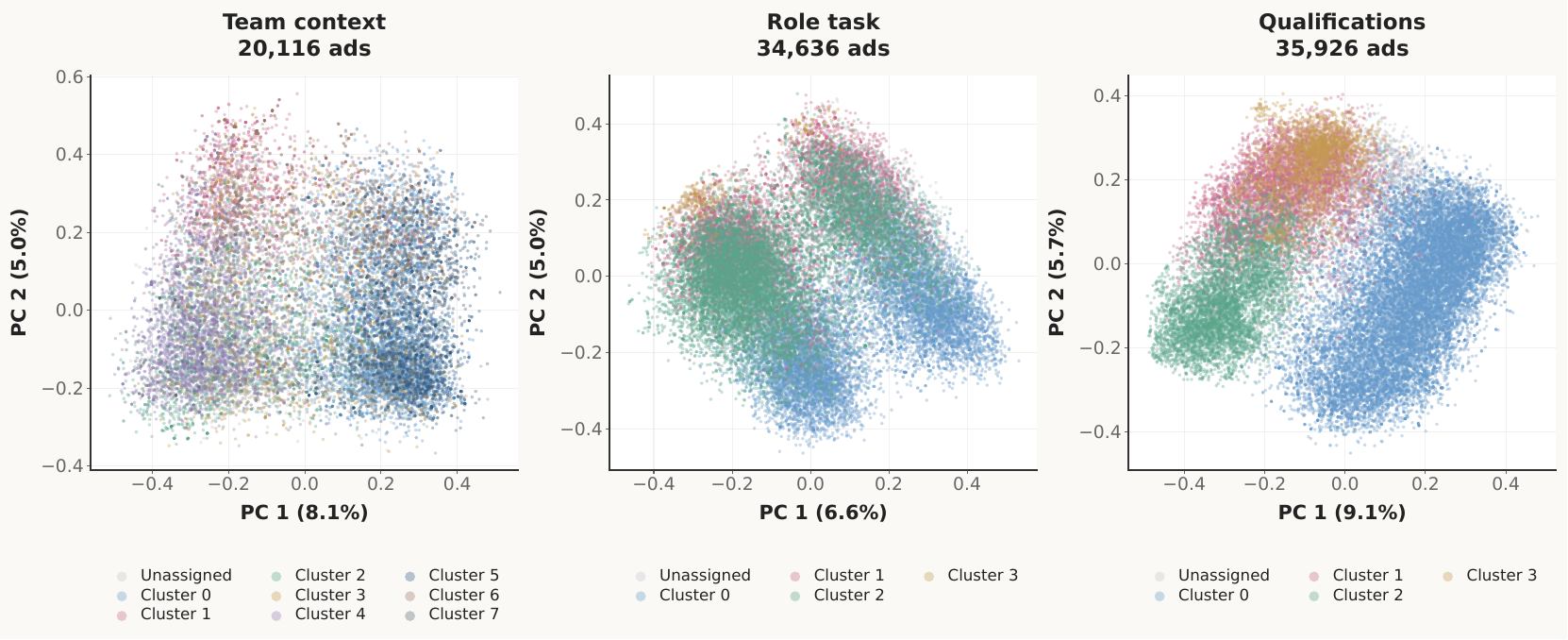}
\caption{A series of PCA scatter plots of PERC clusters for Team context, Role task and Qualifications. Each point represents one job advertisement and each color represents a cluster, with gray representing unassigned advertisements. We can see the clearest colored regions in Qualifications, with the blue cluster concentrated on the right, the green cluster at the lower left and the pink and gold clusters above. Role task also has broad colored regions but more overlap around their boundaries, whereas Team context has several colors mixed through its center.  The first two PCA components explain 11.6\%--14.8\% of the variance in the original job-ad vectors.}
\label{fig:perc_cluster_maps}\end{figure*}

\subsection{Monthly and Occupational Distribution Analytics}
We calculate MMD between collections of original embedding vectors and then rank the comparisons to identify differences for GPT-6.1 Terra to inspect. We compare the same PERC cluster across months and different clusters within the same occupation and month. Each group must contain at least 30 job advertisements from five companies, and we retain up to three comparisons of each type per occupation and segmentation text input.

\begin{figure*}[!tb]\centering
\includegraphics[width=\textwidth,height=.62\textheight,keepaspectratio]{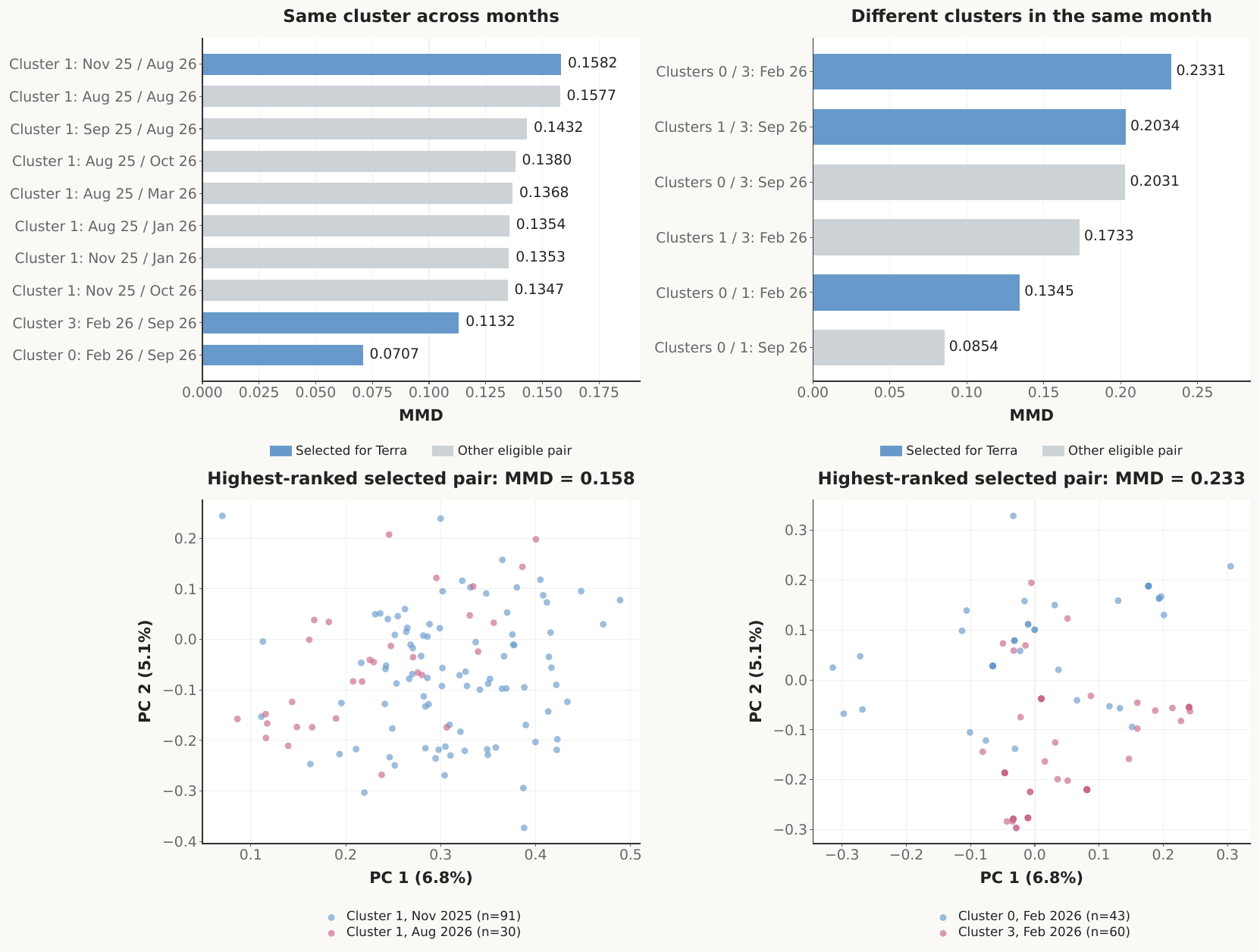}
\caption{A set of ranked bar charts and PCA scatter plots for Data Scientist cluster comparisons. The upper charts rank pairs by MMD and mark those inspected by GPT-6.1 Terra in blue. The lower plots show the highest-ranked selected pairs. The longest monthly-comparison bar belongs to Data Scientist cluster 1 between November 2025 and August 2026, with MMD 0.1582, and the longest same-month bar compares clusters 0 and 3 in February 2026, with MMD 0.2331. In the lower maps, the monthly samples overlap substantially and the two February clusters differ more visibly along PC 2.  Several long gray bars repeat cluster 1 across other month pairs, explaining why the selected monthly comparisons also include clusters 3 and 0.}
\label{fig:mmd_selected_clusters}\end{figure*}

For samples $X=\{x_i\}_{i=1}^{n}$ and $Y=\{y_j\}_{j=1}^{m}$ we measure vector similarity using the Gaussian kernel $k(x,y)=\exp(-\gamma\lVert x-y\rVert_2^2)$ with $\gamma>0$ set using vectors from training companies.

The squared MMD score is defined as
\begin{equation}
\begin{aligned}\widehat{\mathrm{MMD}}^2(X,Y)={}&\frac{1}{n^2}\sum_{i,i'=1}^{n}k(x_i,x_{i'})\\ &+\frac{1}{m^2}\sum_{j,j'=1}^{m}k(y_j,y_{j'})\\ &-\frac{2}{nm}\sum_{i=1}^{n}\sum_{j=1}^{m}k(x_i,y_j).
\end{aligned}
\end{equation}

The first two terms measure the average similarity between job-description vectors within each group, while the third term subtracts twice the average similarity between vectors from the two groups. A larger score indicates that the groups differ more in their embedding distributions. We use these scores to select comparisons for GPT-6.1 Terra to inspect, without treating them as tests of statistical significance. We report the squared score as “MMD” in the figures. Appendix~\ref{app:mmd} provides more information about why we use this form of the MMD equation, how we set the kernel bandwidth and how sample size affects the squared MMD score.

After selecting the comparisons, we give GPT-6.1 Terra the full job descriptions from the two groups being compared. Each description is labeled with its cluster and posting month. We ask Terra to explain differences in purpose, applications, responsibilities, outputs and qualifications, using exact quotations from advertisements in each group. This analysis covers DS, DA, MLE and BA. The following comparison of PERC clusters, SAE features and text alone also includes SWE.

\subsection{SAE Feature Extraction}
\begin{figure*}[!tb]\centering
\includegraphics[width=\textwidth,height=.62\textheight,keepaspectratio]{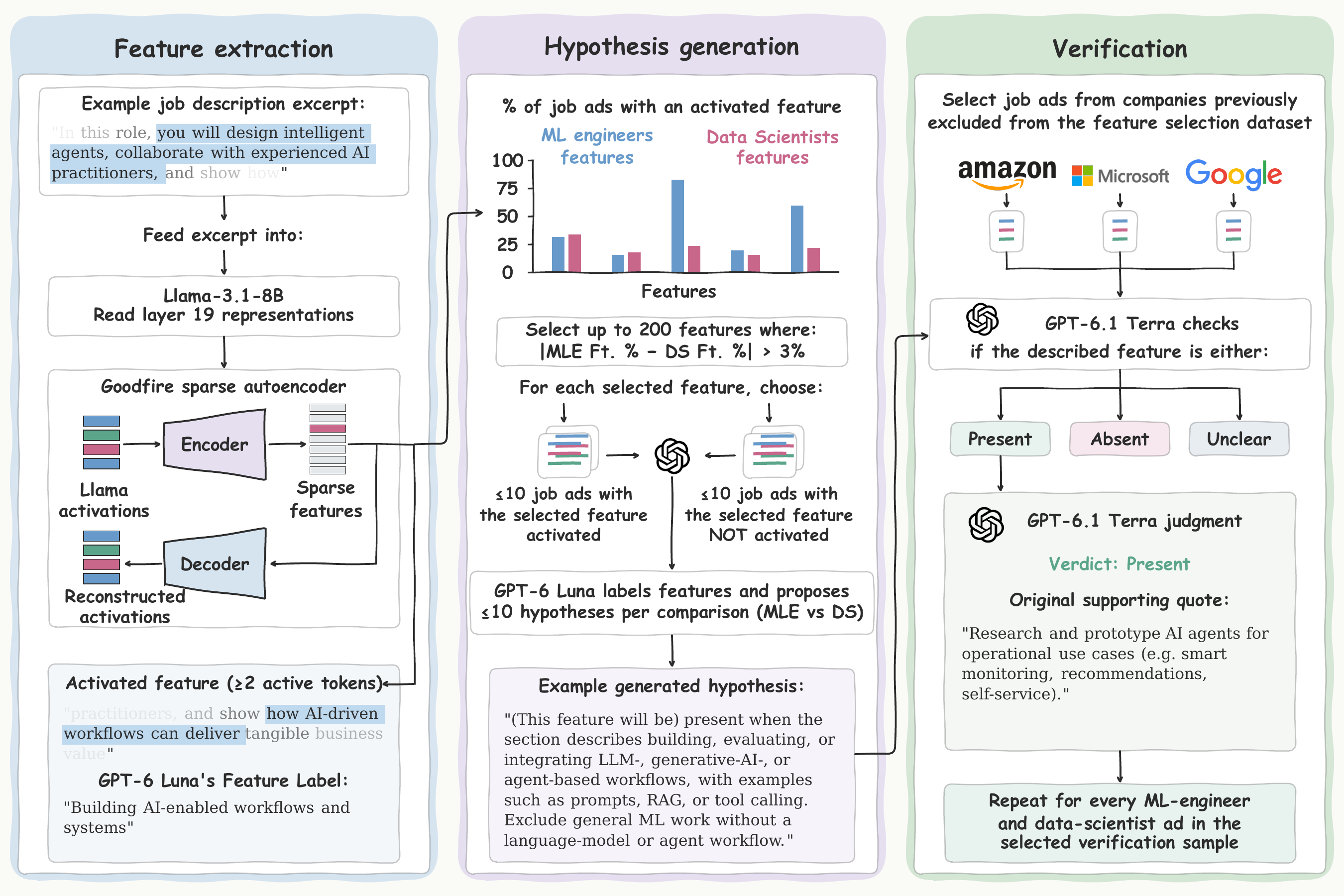}
\caption{A sparse autoencoder feature-extraction workflow diagram that showcases our design for feature extraction, hypothesis generation and verification on held-out companies' advertisements via another LLM verifier. The example shown in the figure connects a job excerpt about \textit{``building AI-workflows''} to a sparse feature within Llama-8B's latent space, and subsequently produces a hypothesis about \textit{``LLM and agent development''}, supported by the quote about prototyping AI agents in GPT-6.1 Terra's verification panel. The text examples are recorded outputs. The bars and company logos illustrate the procedure.}
\label{fig:sae_pipeline}\end{figure*}

For the SAE analysis, we use the pretrained Llama-3.1-8B-Instruct model and its matching Goodfire SAE. The SAE encoder maps each layer-19 hidden state into sparse feature activations, and its decoder reconstructs that state. Several features can activate within the same description so one posting can contain patterns associated with ``developing machine learning models,'' ``reporting'' and ``collaboration''. We use the pretrained weights without fitting a new SAE to the jobs. In the following discussion, ``Developing machine learning models'' refers to the feature abbreviated as ``Model development'' in the figures.

A feature counts as active when it activates on at least two source tokens in the complete section. We compare the percentage of postings activating each feature across occupations or months and then select features with large frequency differences for interpretation. Figure~\ref{fig:sae_pipeline} shows the extraction and interpretation process.

\begin{figure*}[!tb]\centering
\includegraphics[width=\textwidth,height=.62\textheight,keepaspectratio]{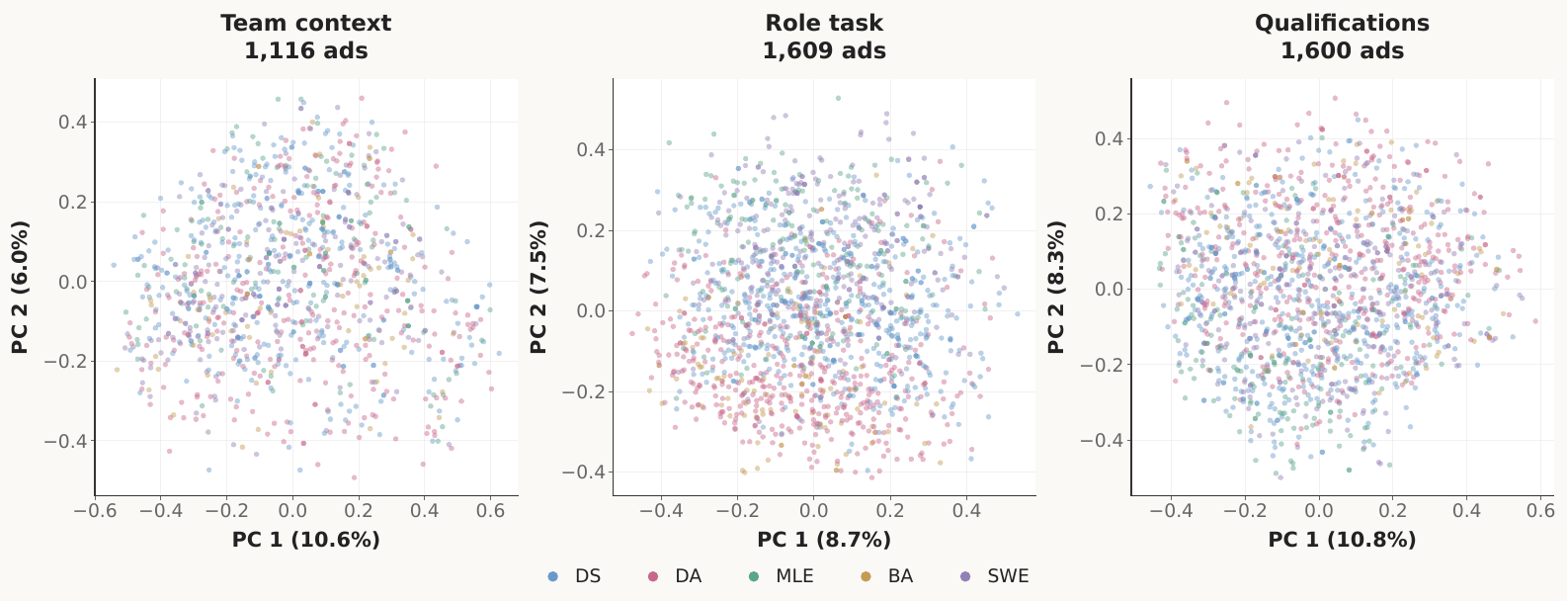}
\caption{A series of PCA scatter plots of job-ad vectors formed from Llama SAE feature activations. Each point represents one advertisement, and its color represents the occupation. These advertisements provide the text for hypotheses about occupational or monthly differences. Role task places more Data Analyst advertisements toward the bottom and more Data Scientist advertisements toward the upper right. Team context and Qualifications show greater overlap. The first two PCA components explain 16.2\%--19.1\% of the variance in the job-ad vectors.}
\label{fig:llama_sae_ad_maps}\end{figure*}

\begin{figure*}[!tb]\centering
\includegraphics[width=\textwidth,height=.62\textheight,keepaspectratio]{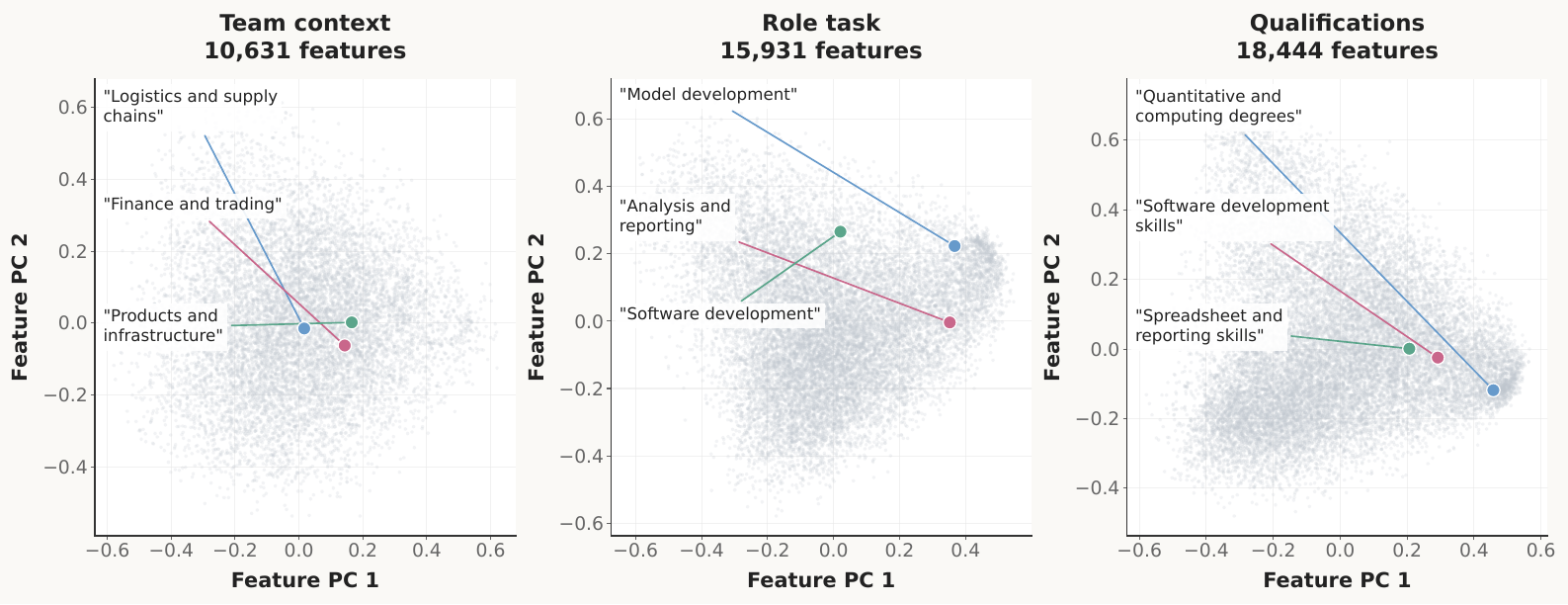}
\caption{A series of PCA scatter plots of SAE features represented by their activation patterns across advertisements. Colored points highlight the selected features. In Role task, ``Developing machine learning models'' and ``Analysis and reporting'' lie farther right than ``Software development.'' In Qualifications, ``Quantitative and computing degrees'' lies farther right than ``Software development skills'' and ``Spreadsheet and reporting skills.'' The highlighted Team context features lie closer together.}
\label{fig:llama_sae_feature_maps}\end{figure*}

\subsection{Using GPT-6 Luna to Label Features and Generate Hypotheses}
For each comparison, we select up to 200 features whose activation frequencies differ by more than 3\%. For each feature, GPT-6 Luna reads up to ten activating and ten nonactivating job-ad sections. We mark the activating text in the active examples so it can describe what the feature appears to capture. We remove unclear labels and labels about syntax or website boilerplate.

\begin{figure*}[!tb]\centering
\includegraphics[width=\textwidth,height=.72\textheight,keepaspectratio]{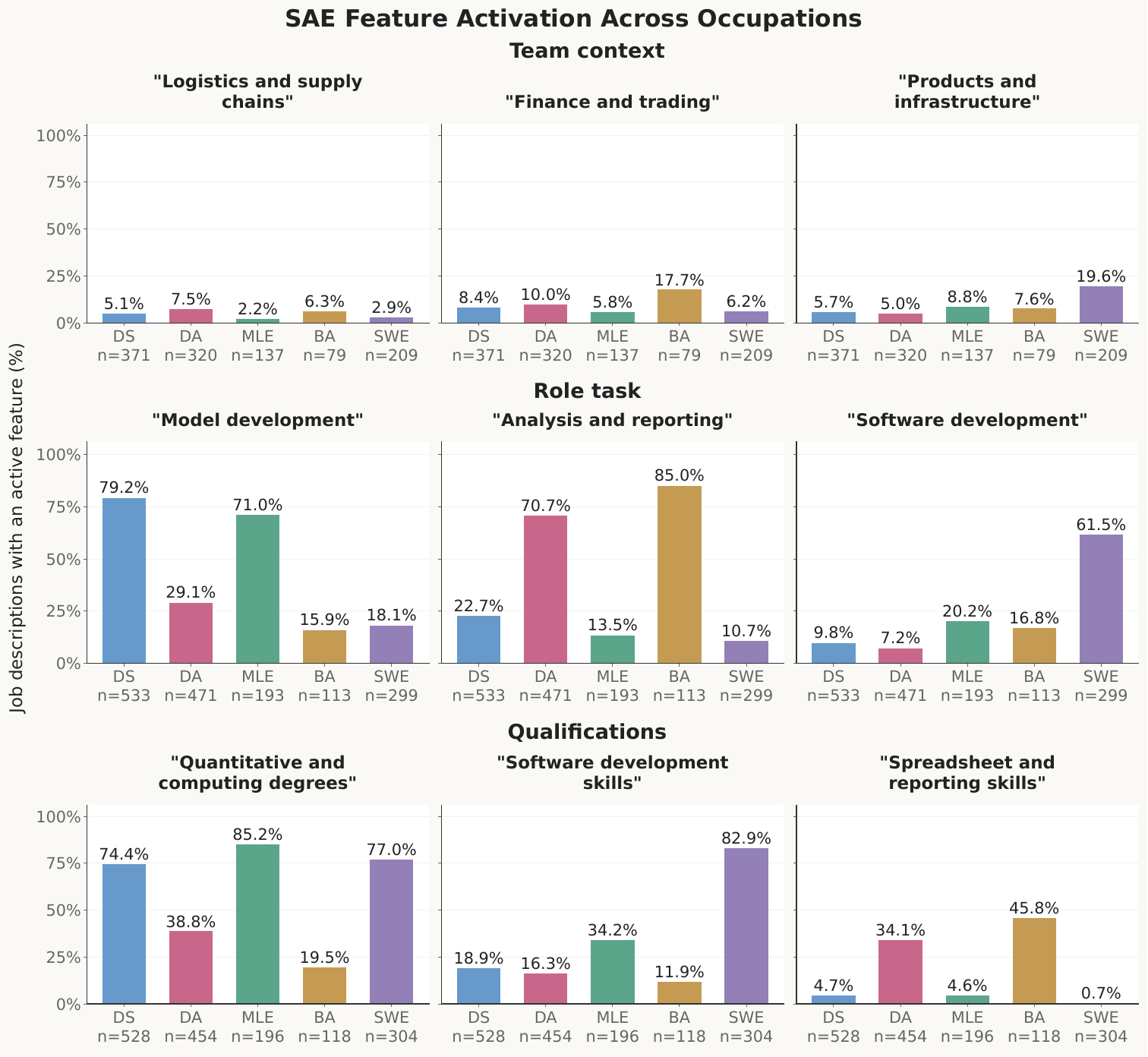}
\caption{A series of bar charts that compares SAE feature activation rates across five occupations. Each row represents a segmented job-description section, and each individual chart represents one sparse feature. Each bar represents an occupation and gives the percentage of its advertisements in which the feature activates. The label $n$ denotes the number of advertisements. ``Developing machine learning models'' peaks in Data Scientist and Machine Learning Engineer at 79.2\% and 71.0\%, ``Analysis and reporting'' in Business Analyst at 85.0\% and ``Software development'' in Software Engineer at 61.5\%. ``Software development skills'' activates most often in Software Engineer and ``Spreadsheet and reporting skills'' in Data Analyst and Business Analyst. Team context rates remain below 20\%.}
\label{fig:sae_feature_rates}\end{figure*}

\begin{figure*}[!tb]\centering
\includegraphics[width=\textwidth,height=.72\textheight,keepaspectratio]{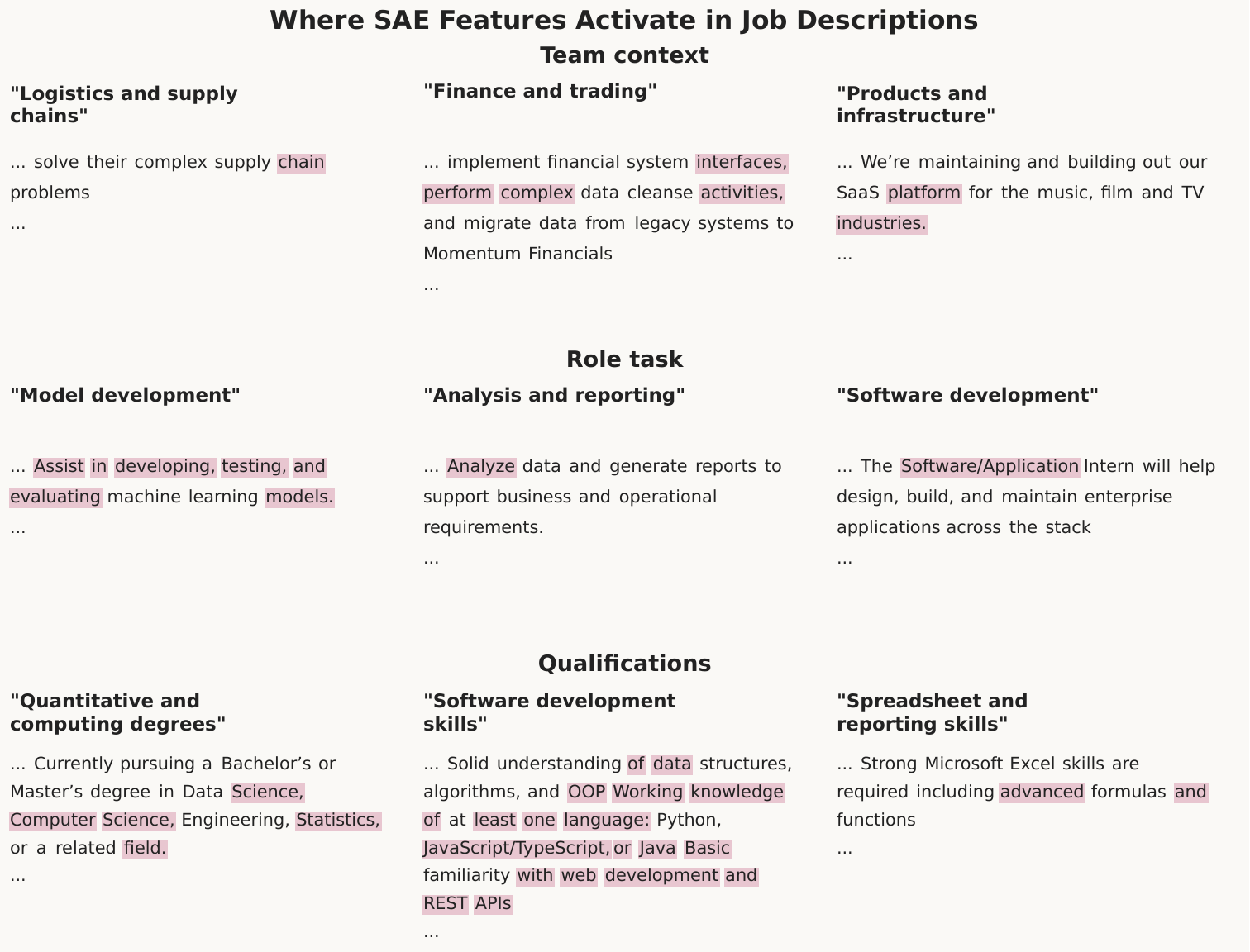}
\caption{A collection of job-ad excerpts that shows where the nine SAE features shown here activate. Pink highlights words containing activating tokens and ellipses mark omitted text. The highlighted words connect the feature names to specific passages: ``Developing machine learning models'' activates around developing, testing and evaluating ``machine learning models'', and the ``Quantitative and computing degrees'' feature activates around fields such as computer science and statistics. Other highlights fall on broader wording such as ``interfaces'' and ``activities'' in the ``Finance and trading'' example or ``and'' in the ``Spreadsheet and reporting skills'' example. The highlighted passages shown in the figure show why we interpret each activation in its surrounding text.}
\label{fig:sae_activating_passages}\end{figure*}

After this, GPT-6 Luna combines related labels into at most ten hypotheses per comparison. A hypothesis is a claim proposed by GPT-6 Luna about the differences in the Role task, Team context or Qualifications described by two occupations. For monthly comparisons, the claim describes differences between two posting months within the same occupation. For example, ``LLM and agent workflow development is more common in MLE than DS tasks'' covers building or evaluating those workflows and excludes general ML work. We keep each definition unchanged when checking descriptions from held-out companies.

\subsection{Comparing the Methods and Checking the Proposed Differences}
To compare the methods fairly, GPT-6 Luna reads the same complete job-ad sections under three conditions. In the PERC condition, it also receives cluster assignments and distribution scores, and in the SAE condition it receives feature labels and activating examples. In the text-only condition, it reads the descriptions without either representation. GPT-6 Luna generates the hypotheses under all three conditions using the same hypothesis limit and maximum generation budget.

We then ask GPT-6.1 Terra to check whether each proposed characteristic is present, absent or unclear in descriptions from held-out companies and provide an exact quote for each positive judgment. This tells us whether the pattern extends beyond the postings used to propose it. We compare the proportions judged present after selecting advertisements with similar formats and section lengths, using the same posting month for occupation comparisons. We also remove duplicate text between the two stages.

We report company-aware confidence intervals and adjust for testing many hypotheses using Benjamini--Hochberg (BH) and the more conservative Benjamini--Yekutieli (BY) procedures. (Appendix~\ref{app:verification} provides more information about how we sample the descriptions and test the proposed differences.) These checks support differences in the selected samples, with the source quotes explaining what those differences mean about the jobs.

\section{Evaluation}
\label{sec:evaluation}
To evaluate the two approaches, we first examine how PERC groups job advertisements and then inspect the SAE features that activate within those groups. After this, we compare the claims that GPT-6 Luna generates when it reads descriptions with PERC clusters, with SAE features or as text alone, and ask GPT-6.1 Terra to check those claims in advertisements from held-out companies.

\subsection{Selected PERC Frameworks}
After comparing the seven segmentation text inputs, we select a PERC configuration for each input using the criteria described in Section~\ref{sec:methodology}. Table~\ref{tab:frameworks} shows that Role task and Role task + Qualifications use UMAP with K-means, whereas Team context uses PCA with hierarchical agglomerative clustering. Role task + Qualifications has the highest TF-IDF cosine silhouette at 0.064, compared with 0.008 for Team context + Qualifications, so these selected clusters differ more in their word use when we combine tasks with qualifications.

To examine whether these clusters also separate in the original embedding space, we calculate vector silhouette scores and obtain scores between 0.014 and 0.091. These small positive scores indicate that there is substantial overlap between descriptions, which we also observe in the two-dimensional maps in Figure~\ref{fig:perc_cluster_maps}. We therefore use the clusters to examine related work within the advertisements without assuming that each cluster represents a separate profession.

We also compare cluster assignments across repeated runs using Adjusted Rand Index (ARI). The four configurations using hierarchical agglomerative clustering have an ARI of one because they repeat the same assignments on the same data, whereas Role task has an ARI of 0.781. This comparison measures changes across runs on the same advertisements, and the evaluation on held-out companies examines whether the selected configurations also group advertisements from different employers. (Appendix~\ref{app:ablations} provides the full configuration results and the measurements on held-out companies.)

\begin{table*}[!tb]\centering\footnotesize
\caption{Selected PERC configurations and validation results for the seven text inputs. Role task + Qualifications achieves the highest word separation (0.064), while Team context + Role task has the highest embedding-space silhouette (0.091). The silhouette score here measures separation in the original embeddings. Word separation (sep.) reports the TF-IDF cosine silhouette score, which measures how similar word use is within each cluster compared with the nearest other cluster, and ARI (or Adjusted Rand Index) measures how consistently the same advertisements are grouped across repeated runs and after accounting for chance. Assigned \% gives the percentage of validation advertisements assigned to a cluster, which measures how much of the validation sample the clustering method covers.}\label{tab:frameworks}
\setlength{\tabcolsep}{3pt}
\begin{tabularx}{\textwidth}{>{\raggedright\arraybackslash}p{1.25in}>{\raggedright\arraybackslash}Xrrrr}
\toprule
Segmentation text input & Selected framework & Silhouette & Word sep. & ARI & Assigned (\%) \\
\midrule
Team context & MPNet, PCA 50\newline Hierarchical agglomerative clustering (8) & 0.014 & 0.012 & 1.000 & 93.4 \\
Role task & text-embedding-3-small, UMAP 50\newline K-means (4) & 0.069 & 0.044 & 0.781 & 93.4 \\
Qualifications & text-embedding-3-small, PCA 10\newline Hierarchical agglomerative clustering (4) & 0.082 & 0.023 & 1.000 & 94.0 \\
Team context + Role task & MPNet, PCA 10\newline Spectral (4) & 0.091 & 0.012 & 0.999 & 93.8 \\
Team context + Qualifications & MiniLM, PCA 50\newline Hierarchical agglomerative clustering (4) & 0.032 & 0.008 & 1.000 & 93.9 \\
Role task + Qualifications & BGE, UMAP 10\newline K-means (4) & 0.070 & 0.064 & 0.976 & 90.8 \\
All three sections & MiniLM, PCA 50\newline Hierarchical agglomerative clustering (4) & 0.072 & 0.012 & 1.000 & 93.8 \\
\bottomrule\end{tabularx}\end{table*}

\subsection{SAE Features Within Selected PERC Clusters}
\label{sec:featureclusters}
To examine what the PERC clusters contain, we apply the SAE to the complete selected section of every advertisement in three comparisons chosen by MMD. These comparisons cover Data Scientist Team context in February 2026, Business Analyst Role task in April versus September 2026, and Data Scientist Qualifications in December 2025.

\begin{figure*}[!tb]\centering
\includegraphics[width=\textwidth,height=.70\textheight,keepaspectratio]{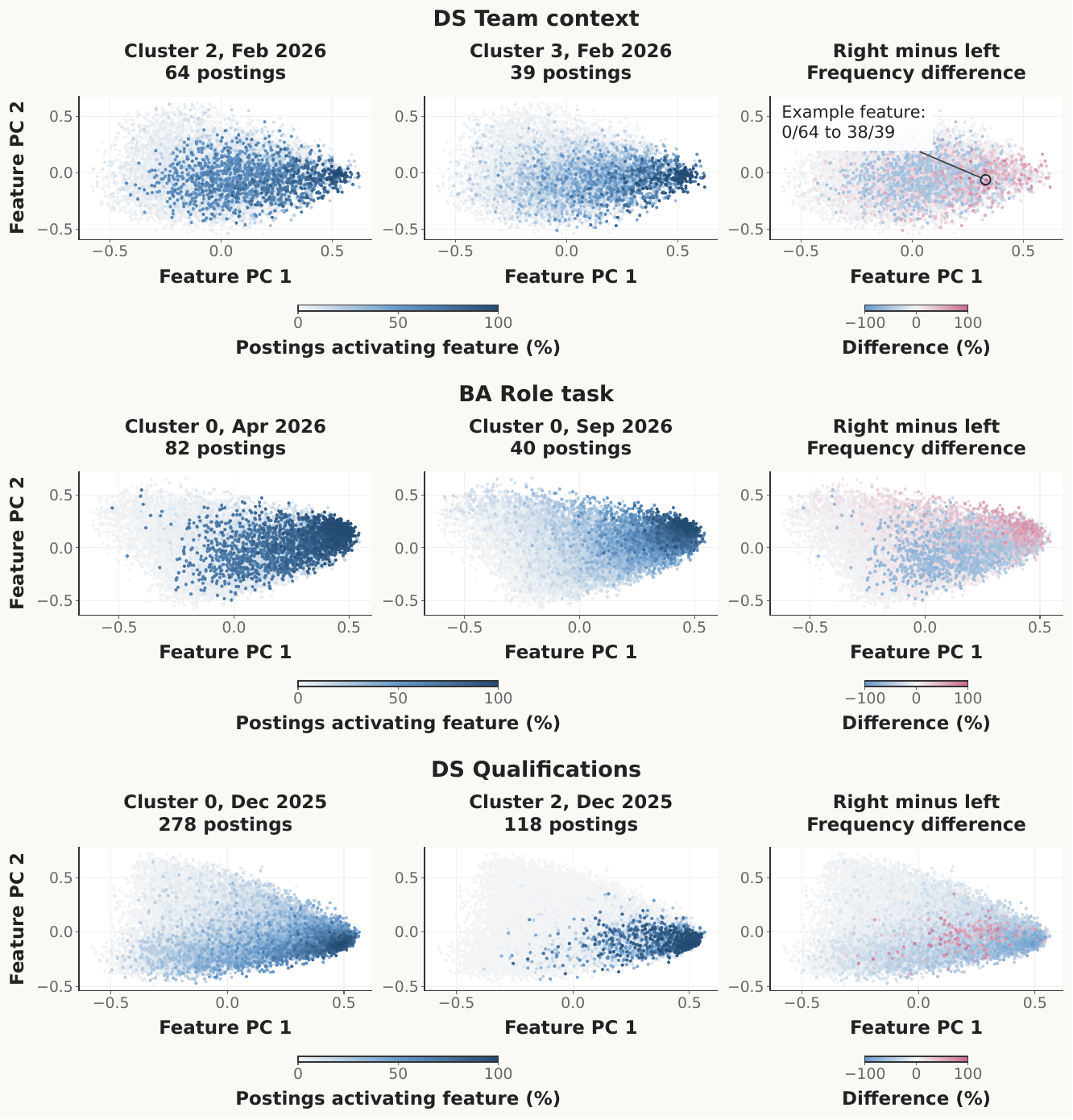}
\caption{A set of PCA feature maps for different occupations and segmentation text inputs. Each row compares extracted SAE features between two PERC cluster samples from the same occupation and job-description section. Each point represents an individual feature. For the first two columns, the greater the blue intensity, the greater the percentage of job advertisements in which the feature activates as Llama-3.1-8B reads the section. The last column is the difference between the percentages of jobs in which each feature activated in one cluster versus the other cluster, with the pink color scheme representing a higher activation of features in the second cluster compared to the first cluster. In the Team context of Data Scientist roles, the marked feature activates in 38/39 advertisements in cluster 3 and none of 64 in cluster 2. In the Role task section of Business Analyst roles, we compare the same cluster in April and September, where the pink and blue regions represent features that activate in a greater or smaller percentage of September advertisements. In the Qualifications of Data Scientist roles, the blue regions represent features that activate in a smaller percentage of cluster 2 advertisements than in cluster 0 advertisements.}\label{fig:sae_selected_cluster_profiles}
\end{figure*}

For each comparison, Figure~\ref{fig:sae_selected_cluster_profiles} places the SAE features at the same PCA coordinates so that we can compare their activation percentages between the two PERC cluster samples. Each point represents one feature, and its color represents the percentage of advertisements in which that feature activates. We also use K-means to group features that activate in similar descriptions, then average their activation percentages within each feature group to examine whether the differences extend across related features. (Appendix~\ref{app:featureclusters} provides the feature-group maps and explains how we calculate these averages.)

In the Team context of Data Scientist roles, the feature labeled ``health- and insurance-sector work'' activates in 38/39 advertisements in cluster 3 and none of 64 in cluster 2. When we read the descriptions in cluster 3, we find references to genomics research, clinical operations and drug development, which connect this activation pattern to the healthcare work described in GPT-6.1 Terra's PERC narrative.

For the Role task section of Business Analyst roles, we compare the same PERC cluster in April and September. In Figure~\ref{fig:sae_selected_cluster_profiles}, the pink and blue regions show that some features activate in a higher percentage of September advertisements and others in a lower percentage. When we average these percentages within each group of related SAE features, however, the April and September averages differ by at most 3.1\%. These averages can therefore look similar even when individual features activate in different percentages of advertisements across the two months.

For the Qualifications of Data Scientist roles, all ten feature groups have a lower average activation percentage in cluster 2 than in cluster 0. To interpret this difference, we also compare the length of the qualification sections and find a median of 55 tokens in cluster 2 compared with 140 in cluster 0. The lower activation percentages therefore occur in the cluster with shorter descriptions, so the comparison includes a difference in how much qualification text the SAE reads.

\subsection{SAE Features Across Occupations and Posting Periods}
After examining the selected clusters, we compare features across the five occupations in Figure~\ref{fig:sae_feature_rates}. For example, ``Developing machine learning models'' activates in 79.2\% of the selected Data Scientist advertisements and 71.0\% of the selected Machine Learning Engineer advertisements, and ``Analysis and reporting'' activates in 85.0\% of the selected Business Analyst advertisements. Because several features can activate in one description, we can examine overlapping characteristics of the work without assigning every advertisement to only one feature.

\begin{figure}[!tb]\centering
\includegraphics[width=\linewidth,height=.62\textheight,keepaspectratio]{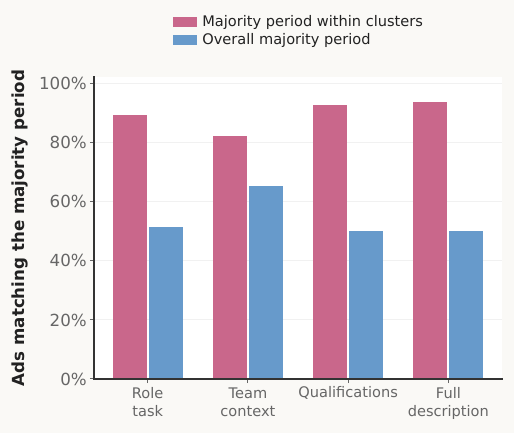}
\caption{A grouped bar chart that compares posting periods within clusters formed from Gemma SAE features. For each job-description section, pink represents the percentage of advertisements matching their cluster's most common posting period, and blue represents the percentage in the dataset's most common posting period. We separate the periods at January 2026. Pink exceeds blue in every section, reaching about 93\% compared with 50\% for Qualifications and full descriptions, so the clusters separate earlier and later advertisements. These periods also differ in description format and length.}
\label{fig:sae_source_period}\end{figure}

\begin{figure*}[!tb]\centering
\includegraphics[width=\textwidth,height=.62\textheight,keepaspectratio]{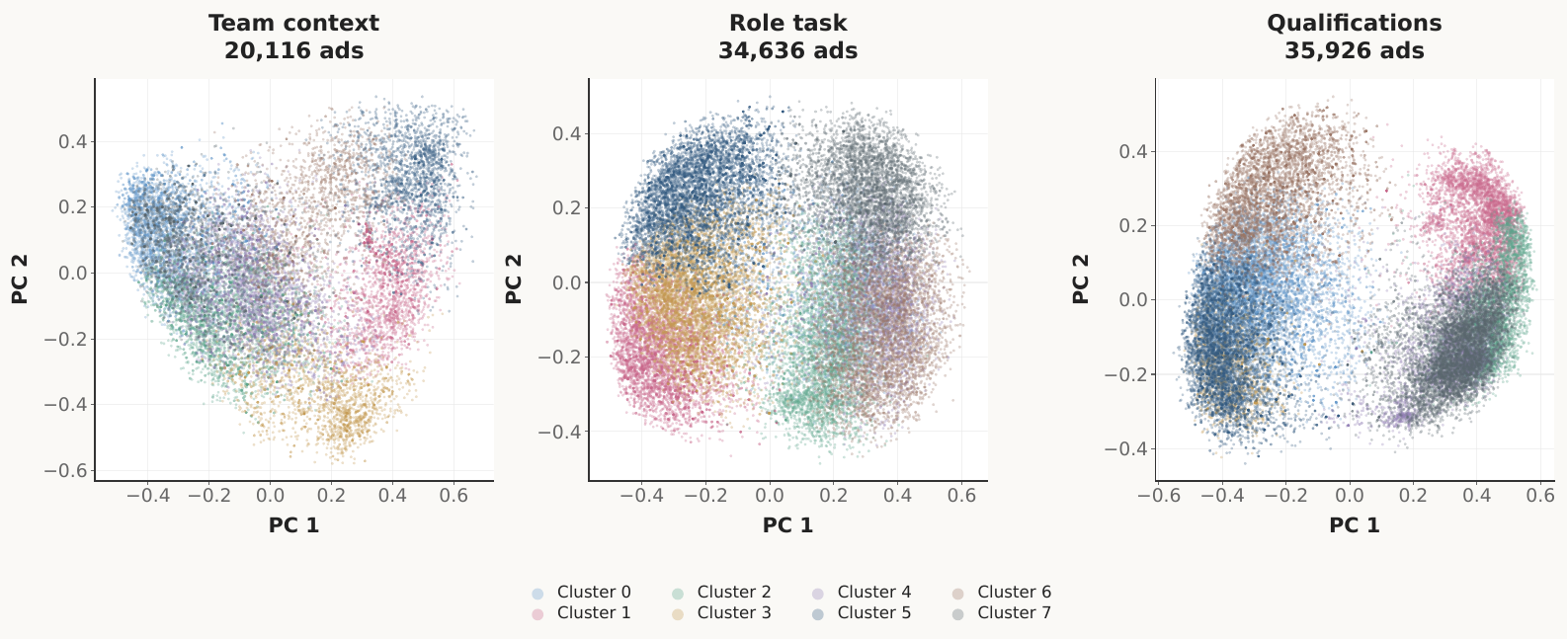}
\caption{A series of PCA scatter plots of job-ad vectors formed from Gemma SAE feature activations. Each point represents one advertisement and its color represents the cluster. The Gemma SAE projection separates Qualifications into two broad sides, with blue and brown clusters on the left and pink, green and gray clusters on the right. Role task forms a more continuous cloud with different colors concentrated in its corners, and Team context shows several regions joined through an overlapping center.  Together with the posting-period comparison in Figure~\ref{fig:sae_source_period}, these maps show that visible grouping can reflect description format and length as well as work content.}
\label{fig:gemma_sae_cluster_maps}\end{figure*}

To examine whether SAE activations also respond to how descriptions are written, we inspect the separate Gemma SAE analysis of the full dataset in Figures~\ref{fig:sae_source_period} and~\ref{fig:gemma_sae_cluster_maps}.

In this analysis, the Role task clusters largely separate advertisements posted before January 2026 from those posted in January 2026 or later, with 89.2\% of advertisements matching the most common posting period in their cluster. Two analytics clusters also differ in median description length, at 39 and 160 tokens, so the visible separation also needs to be understood alongside differences in format and text length. For the subsequent method comparison, we therefore match advertisements with similarly formatted descriptions and comparable section lengths to reduce these differences. (Appendix~\ref{app:sae} provides more information about the Gemma SAE analysis.)

\subsection{Comparison of the Insights}
To compare what the two frameworks add to language-model reading, we give GPT-6 Luna the same job-ad sections under three conditions: accompanied by PERC clusters, accompanied by SAE features or without either representation. A hypothesis is a claim proposed by GPT-6 Luna about the differences in the Role task, Team context or Qualifications described by two occupations. For monthly comparisons, GPT-6 Luna makes the same kind of claim about two posting months within one occupation.

With SAE features, GPT-6 Luna generates 26 occupational claims for the Data Scientist--Data Analyst and Machine Learning Engineer--Data Scientist comparisons, compared with 20 using PERC clusters and 22 using text alone. For the Software Engineer comparisons, PERC inputs lead to 29 claims compared with 24 under each remaining condition. These counts show that the number of candidate differences depends on both the input and the occupations compared. More claims give us more possibilities to investigate, but several claims can overlap or fail to appear in held-out advertisements, so the counts alone cannot establish which method produces better insights.

After generating the hypotheses, GPT-6.1 Terra reads advertisements from held-out companies and checks whether the characteristic described in each claim appears in each description. For example, when GPT-6 Luna uses PERC clusters to propose a claim about ``developing machine learning models,'' GPT-6.1 Terra finds that characteristic in 60/98 Data Scientist task sections and 7/98 Data Analyst task sections. When GPT-6 Luna uses SAE features, the corresponding counts are 48/98 and 5/98 under its resulting definition. Both comparisons support the same occupational contrast, although the definitions describe somewhat different sets of tasks. Section~\ref{sec:insights} explains these differences using the source descriptions and advertisement counts.

The Business Analyst and monthly comparisons also contain examples of different work, but none passes the correction for multiple tests. We therefore report the observed characteristics and their sample sizes without treating those examples as established differences across the wider job market.

The three conditions also differ in API cost. Across the Data Scientist, Data Analyst and Machine Learning Engineer comparisons, labeling SAE features and generating hypotheses costs \$17.50, compared with \$0.28 when GPT-6 Luna uses PERC clusters and \$0.31 when it reads the text without either representation. These amounts exclude shared verification and local computation, so they compare the API cost of preparing and generating the hypotheses. (Appendix~\ref{app:verification} provides the costs and statistical test results for each comparison study.)

\section{What the Job Advertisements Tell Us About the Work}
\label{sec:insights}
Employers often ask for similar programming tools but expect applicants to use them for different purposes. Their descriptions ask applicants to develop ``machine learning models'', maintain reports or build applications, so we examine Team context, Role task and Qualifications together. Most of our advertisements describe internships, so these findings primarily describe the work offered to interns at the companies in our dataset.

\subsection{Data Scientists Build Machine Learning Models and Data Analysts Track Business Performance}
Data Scientist advertisements more often describe developing ``machine learning models'' to answer a question or predict an outcome, whereas Data Analyst advertisements more often describe reports and dashboards that help a business understand its performance. The difference is therefore in what the employer expects the analysis to produce, even when both jobs involve programming and working with data.

Reporting also appears in Data Scientist work, particularly when the applicant is expected to explain the results of an analysis. This helps distinguish preparing a report to communicate a finding from maintaining reports that a team uses regularly. In these descriptions, the main distinction is whether the employer needs a new analytical result or dependable information about its ongoing operations.

The qualifications follow a similar pattern. Data Scientist advertisements place more emphasis on ``statistical and machine learning methods'', and Data Analyst advertisements more often ask for ``spreadsheet, reporting and visualization skills''. Both roles can require programming, so a list of languages alone gives an incomplete picture of the job.

\subsection{Machine Learning Engineers Put Machine Learning Models into Working Applications}
\begin{figure*}[!tb]\centering
\includegraphics[width=\textwidth,height=.62\textheight,keepaspectratio]{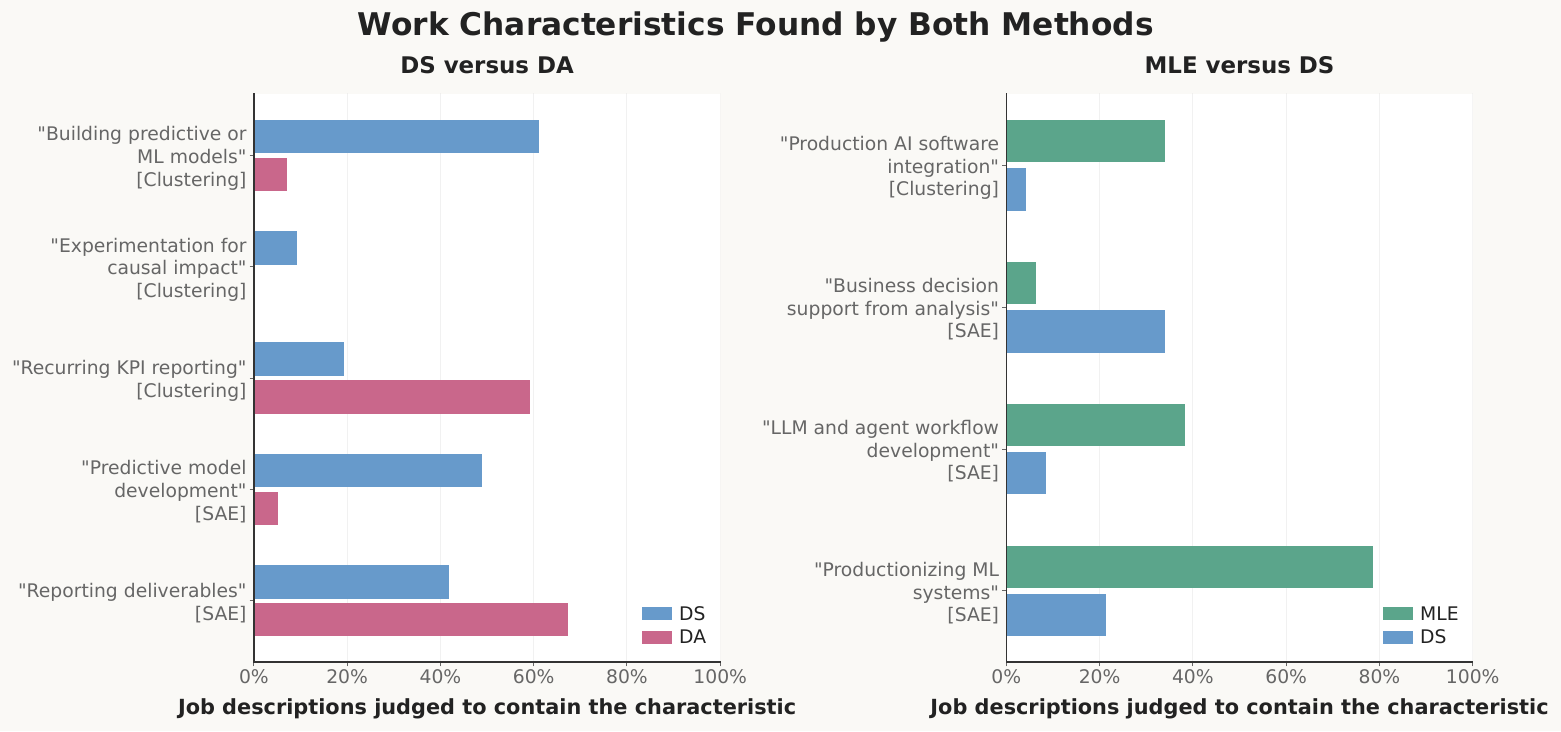}
\caption{A pair of bar charts comparing the work described in Data Scientist and Data Analyst advertisements on the left, and Machine Learning Engineer and Data Scientist advertisements on the right. Each bar represents the percentage of checked advertisements describing the given activity, and the labels distinguish findings from the PERC and SAE comparisons. Data Scientist advertisements more often describe ``developing machine learning models,'' and Data Analyst advertisements more often describe ``reporting.'' The right panel shows a stronger emphasis on ``building and operating AI applications'' in Machine Learning Engineer advertisements, compared with ``analysis used to inform business decisions'' in Data Scientist advertisements.}
\label{fig:verified_work_characteristics}\end{figure*}

Machine Learning Engineer advertisements describe work that extends from developing ``machine learning models'' to making them usable in an application. Employers ask applicants to connect ``machine learning models'' to software, build the processes that prepare their data and check how the ``machine learning models'' behave once they are in use. This places more emphasis on operating an AI application than we observe in Data Scientist advertisements focused on answering analytical questions.

Some descriptions also involve applications of language models and AI agents, including retrieving information, connecting language models to tools and testing how these systems work. These examples show that the engineering work around ``machine learning models'' can be a substantial part of the role, giving these jobs a strong software-engineering component alongside developing ``machine learning models''.

\subsection{Business Analysts Also Build Dashboards and Reports}
\begin{figure*}[!tb]\centering
\includegraphics[width=\textwidth,height=.62\textheight,keepaspectratio]{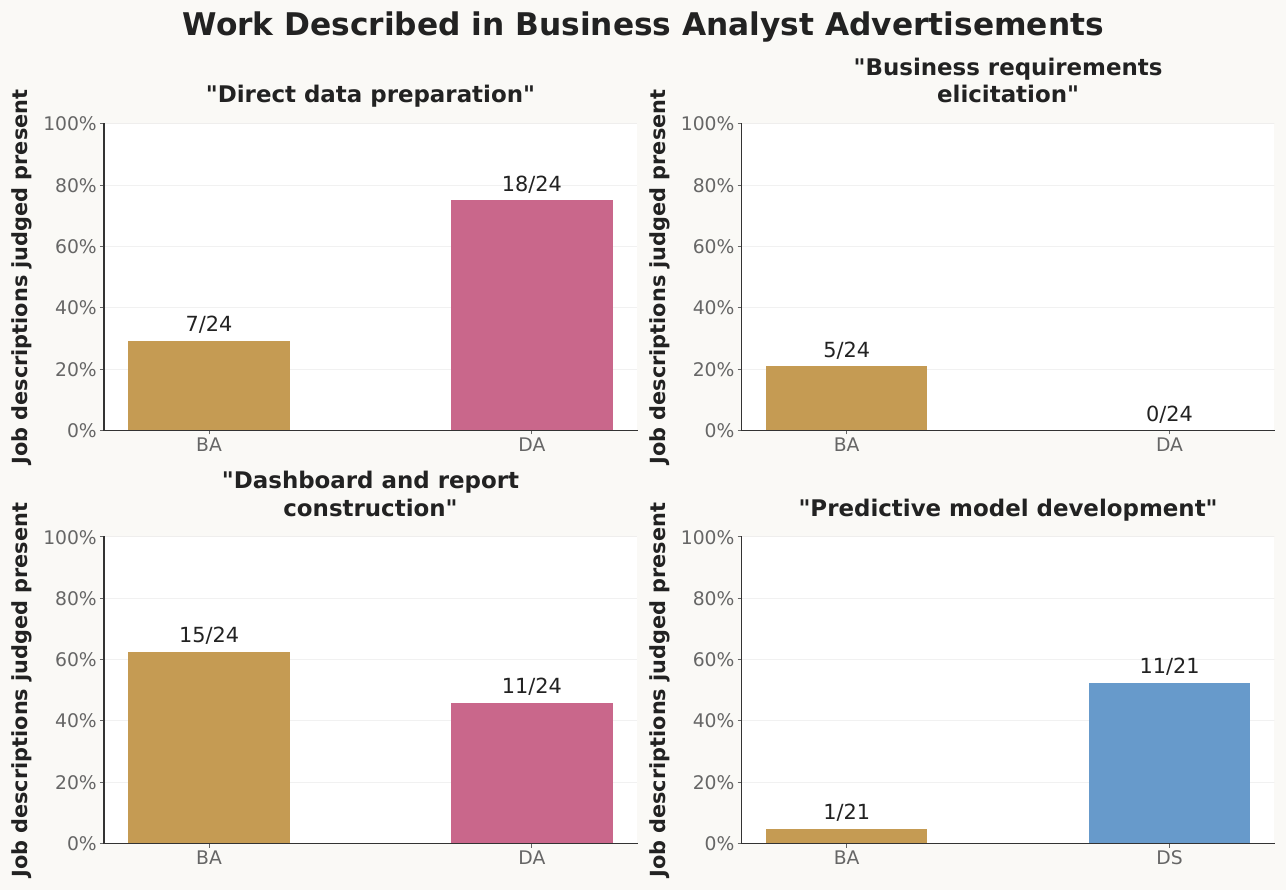}
\caption{A set of bar charts comparing the work described in Business Analyst advertisements with Data Analyst and Data Scientist advertisements. Each bar represents the percentage of checked advertisements describing the given activity. Business Analyst descriptions include ``gathering business requirements'' as well as ``building dashboards'', whereas Data Analyst descriptions more often emphasize ``preparing data'' and Data Scientist descriptions more often emphasize ``developing predictive machine learning models''. These examples show that while there exists overlap between the two 'Analyst' titles, they are still fundamentally different in general of as one focuses more on aligning with business requirements while the other focuses more on untangling data to find raw data insights.}
\label{fig:ba_work_characteristics}\end{figure*}

In our small Business Analyst sample, several descriptions connect understanding a business request with building something that meets it. Employers ask applicants to gather requirements, explain how a process should work and specify how a team will know whether the result meets its needs. Several of those same descriptions also include building reports or dashboards, so the role can involve carrying a business request through to a working analytical tool.

One advertisement connects gathering business requirements with building Power BI reports, checking data quality and testing AI applications. The reporting work overlaps with Data Analyst responsibilities, so employers in these examples use the Business Analyst title for roles that also include practical data work. The descriptions accept both business and computing or quantitative study backgrounds, consistent with work that connects business needs to technical implementation.

\subsection{The Same Job Title Can Mean Very Different Work}
The setting in which a team works changes the questions an applicant would be expected to answer. Within Data Science, we find descriptions about improving manufacturing quality alongside descriptions about disease research and healthcare operations. Forecasting demand for spare parts and studying genetic changes can involve similar analytical methods, but the data, decisions and background knowledge involved are different.

The same distinction appears in Data Analyst advertisements. Some focus on retail inventory and business performance, and others examine healthcare costs, patient outcomes or clinical data. Meanwhile, Business Analyst advertisements differ in whether the applicant would maintain reporting tools or use their results to help plan a business activity. At TikTok, for example, one description focuses on building dashboards and another on analyzing seller activity to inform incentive decisions.

Entry requirements also vary within a shared title. Some Data Scientist internships accept programming experience gained through coursework, whereas others require enrollment in a related graduate program. For an applicant, the team's purpose, the work described and the study requirements therefore give a more useful picture of a role than its title alone. (Appendix~\ref{app:cases} provides the company examples and source passages for these comparisons.)

\subsection{The Projects Described Differ Across Months}
When we compare advertisements across months, we observe differences in the projects employers describe within the same occupation. For Data Scientist roles, the February 2026 examples include infrastructure planning and search-performance reporting, and the September examples include analyzing vineyard images, working with satellite data and using AI to assist work. These descriptions involve different uses of data science, even though the occupation name remains the same.

Machine Learning Engineer advertisements show a similar difference between financial-software work in November 2025 and projects involving finding relevant information, processing data and using ``machine learning models'' to predict vehicle movement in December. For Data Analyst roles, October 2025 reporting descriptions contrast with March 2026 examples involving product experiments and AI agents that query analytical reports. AI-related work already appears in the earlier advertisements, so these examples show different uses of AI across the two months.

For Business Analyst roles, April 2026 examples focus on reporting and operational planning, and September examples include client consulting, predictive analysis and examining possible business scenarios. Taken together, these comparisons show how the applications of a role can differ across the advertisements available at different times. The employers and positions also differ between months, so we are observing different advertised projects rather than following the same employees as their responsibilities change. (Appendix~\ref{app:cases} provides the monthly examples and their source descriptions.)

\subsection{Reporting and Experiment Design Vary Across Months}
\begin{figure*}[!tb]\centering
\includegraphics[width=\textwidth,height=.62\textheight,keepaspectratio]{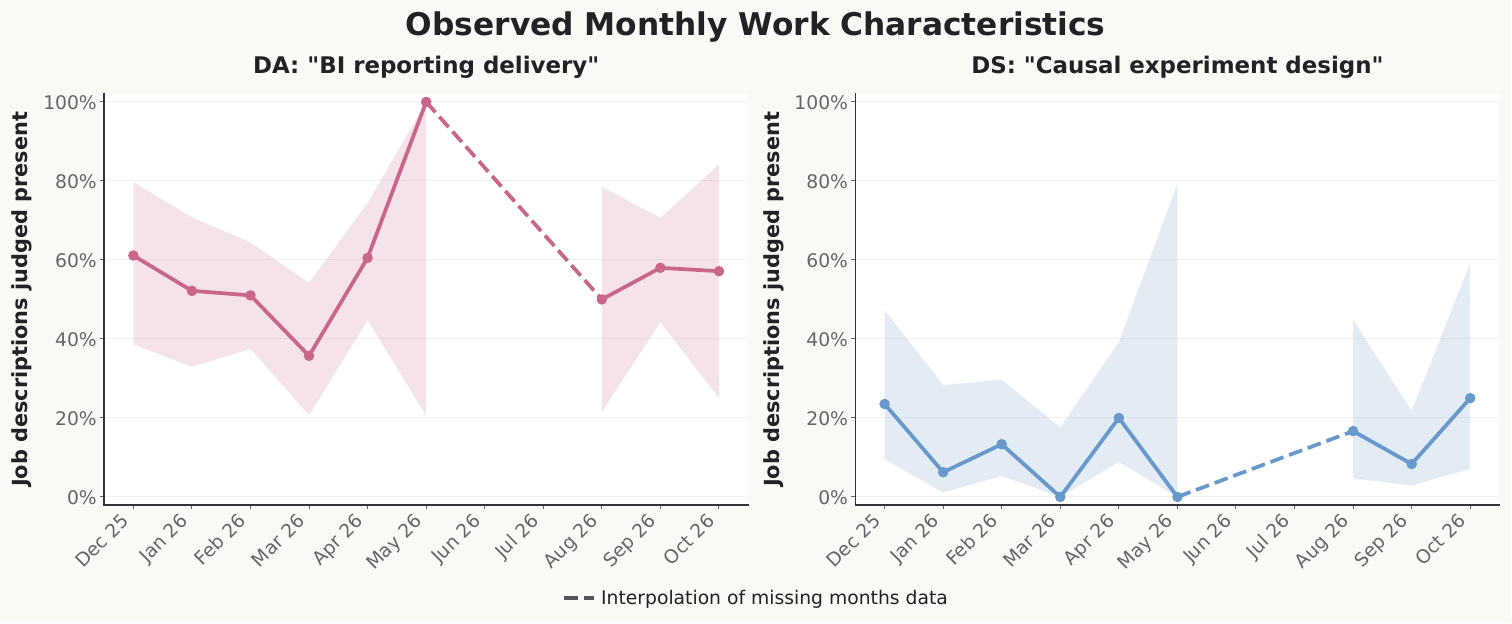}
\caption{A pair of line charts showing how often checked Data Analyst advertisements describe reporting work and Data Scientist advertisements describe designing experiments across posting months. Each line represents the percentage of advertisements containing the given activity, and the shaded areas represent uncertainty around those estimates. The percentage of jobs that requires ``Business intelligence reporting work'' fluctuates across the months, while ``causal experimental design work'' demand fluctuates less but with generally lower in demand. Due to technical errors in our data collection process, we exhibit a gap across june and july that we instead construct an interpolation estimate for.}
\label{fig:verified_monthly_characteristics}\end{figure*}

When we trace the same work characteristics across months, the percentage of Data Analyst advertisements describing ``reporting'' falls and then rises again, and the percentage of Data Scientist advertisements describing ``experiment design'' also fluctuates. These patterns show that the emphasis of the collected descriptions varies across months, so a comparison of two selected months can look different when we examine the months in between.

We also observe changes in narrower parts of the work. Some Data Scientist comparisons differ in their emphasis on ``preparing data'' or ``building data pipelines'', and Software Engineer comparisons differ in whether they mention ``organizing data and presenting it visually''. These examples help describe the projects represented in a particular month, but the small and uneven monthly samples do not establish a consistent change across the wider job market.

\subsection{Software Engineers Are Hired to Build Applications and Services}
\begin{figure*}[!tb]\centering
\includegraphics[width=.85\textwidth,height=.62\textheight,keepaspectratio]{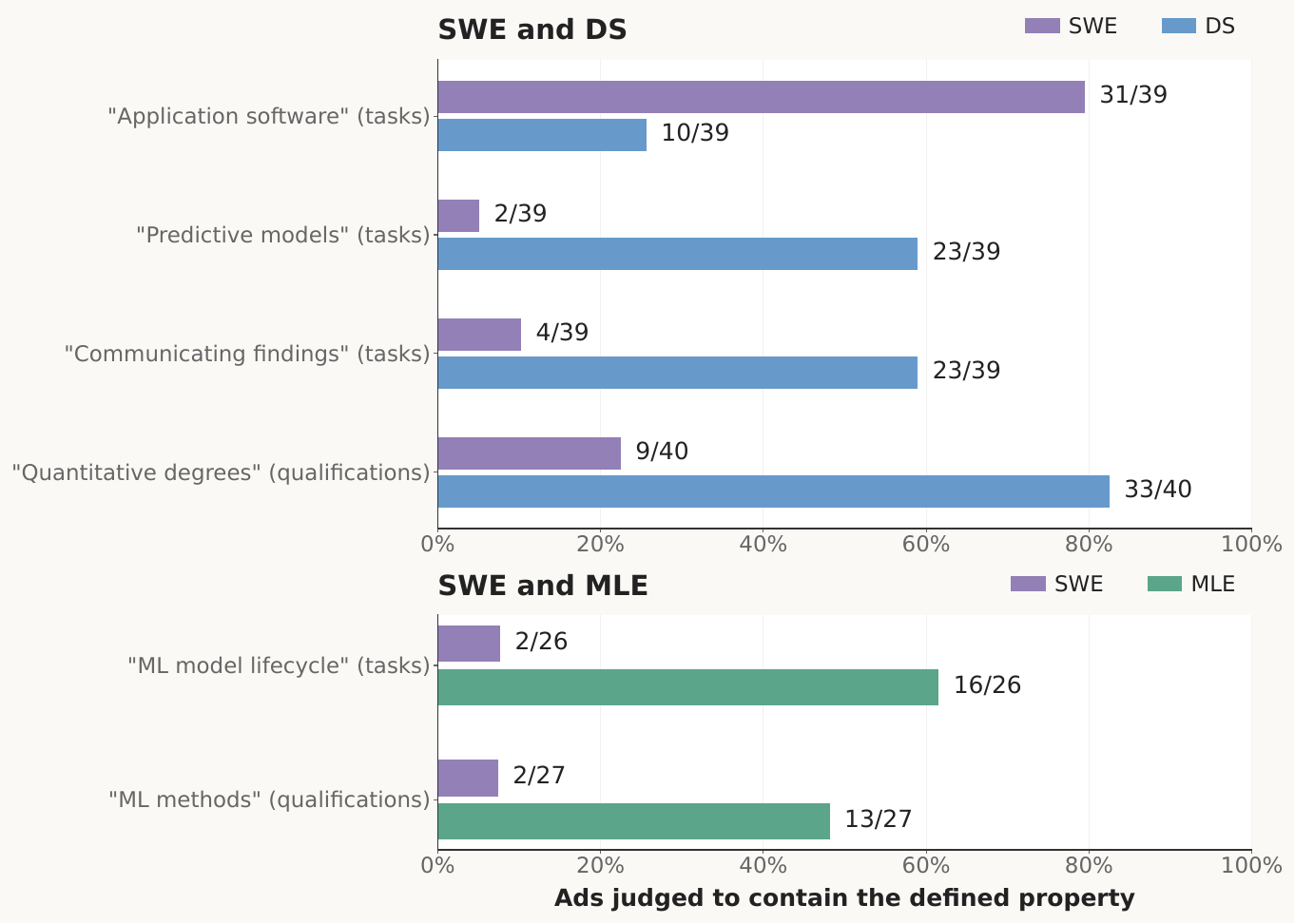}
\caption{A pair of horizontal bar charts comparing Software Engineer advertisements with Data Scientist advertisements above and Machine Learning Engineer advertisements below. Each bar represents the percentage of checked advertisements describing the given task or qualification. Software Engineer descriptions more often focus on building applications, and Data Scientist descriptions more often focus on developing ``machine learning models'' and communicating analytical findings. The lower panel shows that Machine Learning Engineer advertisements place more emphasis on developing and maintaining ``machine learning models'' and on the methods needed for that work.}
\label{fig:swe_verified_work}\end{figure*}

Software Engineer advertisements more often describe building applications and software services, whereas Data Scientist advertisements more often describe developing ``machine learning models'' and explaining analytical findings. One software-engineering advertisement asks applicants to build Python services that give staff access to data and automate internal tasks. A Data Scientist advertisement instead asks for an executive presentation of an analysis, showing how the expected result can differ even when both roles use programming.

The comparison with Machine Learning Engineer roles adds another distinction. Both roles involve software development, but the machine-learning descriptions more often focus on developing, training and evaluating ``machine learning models''. Their qualifications also place more emphasis on ``machine learning methods'', and Data Scientist qualifications more often name statistical methods and quantitative study backgrounds.

\section{Discussion}
PERC and SAE provide different ways to inspect the same job postings. PERC groups descriptions so that GPT-6.1 Terra can explain differences in Team context, Role task and Qualifications. SAE finds patterns that can appear together within one posting, making it useful for examining combinations such as ``developing machine learning models,'' ``reporting'' and ``deploying machine learning models''. The feature maps connect these views by showing which patterns activate within a selected PERC cluster.

The findings explain why jobs using the same tools can involve different work. A Data Scientist may develop ``machine learning models'' and explain their results, a Data Analyst may maintain the reports used to monitor a business, and a Software Engineer may build the application through which that work is used. Machine Learning Engineer advertisements put more emphasis on making ``machine learning models'' work within software, and Business Analyst advertisements can include building reports as well as defining what a business needs. Individual jobs can combine these activities, which is why we examine their descriptions alongside their titles.

We compare what these workflows reveal from segmented job descriptions and where each method falls short. GPT-6 Luna also finds several of the broad occupational differences, so the results do not establish that SAE is generally better or that its findings are unique discoveries.

\subsection{Limitations}
The data primarily describe internships and cover employers and months unevenly. Repeated recruitment, formatting of the original descriptions and inaccurate occupation or passage labels can influence the groups. We do not have independent human judgments for every label or hypothesis. Checking a quotation confirms that it appears in the source advertisement, but does not establish that Terra interpreted it correctly.

The method comparison uses descriptions with similar formats and lengths from separate company sets, though some test postings appeared in earlier exploration. The seven segmentation text inputs contain different samples, so their ablations do not isolate the benefit of segmentation on a common set of full descriptions. Different language models and incomplete accounting of computational costs also prevent a general cost-quality ranking. Monthly samples are often small.

\section{Conclusion}
In this study, we examine how PERC and pretrained SAE analysis can help us understand the work described beneath job titles using 36,081 early-career job postings. By separating each description into Team context, Role task and Qualifications, we examine what applicants would do, which teams they would join and what backgrounds employers request. PERC groups descriptions with similar meaning and uses MMD to select differences for GPT-6.1 Terra to explain. Meanwhile, SAE analysis lets us examine features that can appear together within one description, with GPT-6 Luna generating hypotheses and GPT-6.1 Terra checking those claims against advertisements from held-out companies.

Across the five occupations, we observe differences that skill counts alone would not fully explain. Data Scientist advertisements place more emphasis on developing ``machine learning models'', Data Analyst advertisements on reporting business performance, and Machine Learning Engineer advertisements on making ``machine learning models'' work within applications. Business Analyst descriptions connect business requirements with practical reporting work, and Software Engineer descriptions focus more on building applications and services. These responsibilities also overlap, which helps explain why two jobs requesting the same programming language can involve different work.

Our monthly comparisons further show differences in the projects being advertised. For example, the selected Data Scientist descriptions shift from infrastructure planning and reporting in February 2026 to remote sensing and agricultural analysis in September. Business Analyst descriptions similarly differ between operational reporting in April and consulting and scenario analysis in September. These findings describe changes within our collected advertisements, although uneven monthly coverage limits how far we can extend them to the wider job market. GPT-6 Luna also recovers several broad occupational differences through direct reading, so the additional value of SAE analysis depends on which characteristics we want to examine.

For future work, we would first expand collection across consecutive months and include more full-time positions, since our current findings primarily describe internships. Following the same companies over time would also help us examine whether the work they advertise changes, rather than attributing differences entirely to a changing set of employers.

After this, we would compare all seven segmentation text inputs using the same advertisements to examine whether separating Team context, Role task and Qualifications improves the insights obtained from complete descriptions. We would also compare PERC, SAE analysis and direct language-model reading under the same computational and API budget, with independent reviewers assessing whether the findings are supported by the descriptions and useful for understanding the roles. This would help establish when the additional feature analysis provides enough detail to justify its cost.

\section*{References}

{\small\raggedright \leavevmode\hypertarget{ref-1}{}[1] Mantas Lukauskas, Viktorija Šarkauskaitė, Vaida Pilinkienė, Alina Stundžienė, Andrius Grybauskas, and Jurgita Bruneckienė. 2023. \href{https://www.mdpi.com/2076-3417/13/10/6119}{Enhancing Skills Demand Understanding through Job Ad Segmentation Using NLP and Clustering Techniques}. Applied Sciences, 13(10), 6119. DOI: 10.3390/app13106119.\par}

{\small\raggedright \leavevmode\hypertarget{ref-2}{}[2] Sven Michalczyk, Mario Nadj, Alexander Maedche, and Christoph Gröger. 2021. \href{https://aisel.aisnet.org/ecis2021_rp/115/}{Demystifying Job Roles in Data Science: A Text Mining Approach}. ECIS 2021 Research Papers, 115.\par}

{\small\raggedright \leavevmode\hypertarget{ref-3}{}[3] Jonas Gunklach, Mario Nadj, Sven Michalczyk, Katharina Jacob, Christoph Gröger, and Alexander Mädche. 2026. \href{https://link.springer.com/article/10.1007/s12599-025-00954-2}{Beyond the Unicorn? Job Roles in Data Science}. Business \& Information Systems Engineering, 68, 871--895. Published online in 2025.\par}

{\small\raggedright \leavevmode\hypertarget{ref-4}{}[4] Tim Schlippe and Koen Bothmer. 2023. \href{https://online-journals.org/index.php/i-jac/article/view/34779}{Skill Scanner: An AI-Based Recommendation System for Employers, Job Seekers and Educational Institutions}. International Journal of Advanced Corporate Learning, 16(1), 55--64. DOI: 10.3991/ijac.v16i1.34779.\par}

{\small\raggedright \leavevmode\hypertarget{ref-5}{}[5] Fatih Gurcan, Ahmet Soylu, and Akif Quddus Khan. 2025. \href{https://www.mdpi.com/2071-1050/17/20/9293}{Towards a Sustainable Workforce in Big Data Analytics: Skill Requirements Analysis from Online Job Postings Using Neural Topic Modeling}. Sustainability, 17(20), 9293. DOI: 10.3390/su17209293.\par}

{\small\raggedright \leavevmode\hypertarget{ref-6}{}[6] Alessia De Santo, Lorenzo Malandri, Fabio Mercorio, Mario Mezzanzanica, and Navid Nobani. 2026. \href{https://aclanthology.org/2026.eacl-industry.65/}{SkiLLens: Recognising and Mapping Novel Skills from Millions of Job Ads Across Europe Using Language Models}. EACL Industry Track, 877--885.\par}

{\small\raggedright \leavevmode\hypertarget{ref-7}{}[7] Stephen Meisenbacher and Peter Norlander. 2026. \href{https://arxiv.org/abs/2605.21029}{Building a Custom Taxonomy of AI Skills and Tasks from the Ground Up with Job Postings}. arXiv:2605.21029, accepted to CustomNLP4U 2026.\par}

{\small\raggedright \leavevmode\hypertarget{ref-8}{}[8] Emilio Colombo, Simone D'Amico, Fabio Mercorio, and Mario Mezzanzanica. 2026. \href{https://boa.unimib.it/handle/10281/594429}{VEUCTOR: Training and selecting best vector space models from online job ads for European countries}. Information Sciences, 741, 123274. DOI: 10.1016/j.ins.2026.123274.\par}

{\small\raggedright \leavevmode\hypertarget{ref-9}{}[9] Janne Kauttonen, Umair Ali Khan, Lili Aunimo, Antti Nyqvist, and Aarne Klemetti. 2024. \href{https://www.frontiersin.org/journals/education/articles/10.3389/feduc.2024.1322774/full}{Topic mining for theses and job ads in ICT sector: can higher education institutes respond to job market demands?}. Frontiers in Education, 9, 1322774. DOI: 10.3389/feduc.2024.1322774.\par}

{\small\raggedright \leavevmode\hypertarget{ref-10}{}[10] Marie Labussière and Thijs Bol. 2026. \href{https://sociologicalscience.com/download/volume-13/april/SocSci_v13_362to407.pdf}{Are Occupations ``Bundles of Skills''? Identifying Latent Skill Profiles in the Labor Market Using Topic Modeling}. Sociological Science, 13, 362--407. DOI: 10.15195/v13.a16.\par}

{\small\raggedright \leavevmode\hypertarget{ref-11}{}[11] Arthur Gretton, Karsten M. Borgwardt, Malte J. Rasch, Bernhard Schölkopf, and Alexander Smola. 2012. \href{https://jmlr.org/papers/v13/gretton12a.html}{A Kernel Two-Sample Test}. Journal of Machine Learning Research, 13, 723--773.\par}

{\small\raggedright \leavevmode\hypertarget{ref-12}{}[12] Nils Reimers and Iryna Gurevych. 2019. \href{https://aclanthology.org/D19-1410/}{Sentence-BERT: Sentence Embeddings using Siamese BERT-Networks}. EMNLP-IJCNLP, 3982--3992.\par}

{\small\raggedright \leavevmode\hypertarget{ref-13}{}[13] Hyunjin Cho, Woo-Sung Jung, Jaehyuk Park, and Jisung Yoon. 2026. \href{https://papers.ssrn.com/sol3/papers.cfm?abstract_id=6138547}{Extracting Contextual Substitutes for Workplace Skills using Masked Language Models}. KDI School of Public Policy and Management Working Paper WP26-02. DOI: 10.2139/ssrn.6138547.\par}

{\small\raggedright \leavevmode\hypertarget{ref-14}{}[14] Nick Jiang, Xiaoqing Sun, Lisa Dunlap, Lewis Smith, and Neel Nanda. 2026. \href{https://arxiv.org/abs/2512.10092}{Interpretable Embeddings with Sparse Autoencoders: A Data Analysis Toolkit}. ICML 2026, arXiv:2512.10092v2.\par}

{\small\raggedright \leavevmode\hypertarget{ref-15}{}[15] Tom Lieberum et al. 2024. \href{https://arxiv.org/abs/2408.05147}{Gemma Scope: Open Sparse Autoencoders Everywhere All At Once on Gemma 2}. arXiv:2408.05147.\par}

{\small\raggedright \leavevmode\hypertarget{ref-16}{}[16] Leland McInnes, John Healy, and James Melville. 2018. \href{https://arxiv.org/abs/1802.03426}{UMAP: Uniform Manifold Approximation and Projection for Dimension Reduction}. arXiv:1802.03426, revised 2020.\par}

{\small\raggedright \leavevmode\hypertarget{ref-17}{}[17] Leland McInnes, John Healy, and Steve Astels. 2017. \href{https://joss.theoj.org/papers/10.21105/joss.00205}{hdbscan: Hierarchical density based clustering}. Journal of Open Source Software, 2(11), 205.\par}

\appendix
\section{Data and PERC Implementation}\label{app:data}

\subsection{Data Collection and Preprocessing}
We collect 41,040 records and retain 36,081 advertisements after checking their dates and text, removing duplicate advertisements posted on the same date, identifying company names and selecting early-career positions. These positions include internships, co-ops, junior and new-graduate roles, as well as descriptions requesting zero to two years of experience. Of the retained advertisements, 35,511 are internships and 570 are other early-career positions, which implies that the findings primarily describe internship recruitment.

\begin{table*}[!tb]\centering\small

\caption{The number of job advertisements retained and removed during preprocessing. We report how many advertisements each preprocessing step removes and how many remain. When we select early-career positions, we remove 4,238 advertisements, which is the largest reduction, and retain 36,081 of the original 41,040 records. Empty descriptions account for 319 removed advertisements, and same-day deduplication removes another 276.}

\begin{tabularx}{\linewidth}{>{\raggedright\arraybackslash}X>{\raggedright\arraybackslash}X>{\raggedright\arraybackslash}X}
\toprule

Filtering decision & Ads remaining & Ads removed at this decision \\

\midrule

Collected advertisements & 41,040 & 0 \\

Nonempty posting ID & 41,040 & 0 \\

Readable posting date & 41,025 & 15 \\

Nonempty description & 40,706 & 319 \\

Remove same-day copies & 40,430 & 276 \\

Identified company name & 40,319 & 111 \\

Early-career positions & 36,081 & 4,238 \\

\bottomrule\end{tabularx}\label{tab:original5}\end{table*}

To prepare the text, we remove HTML, normalize whitespace and exclude descriptions that contain only missing-value text, such as ``nan.'' We retain posting IDs containing letters or punctuation because these characters can be part of a valid identifier. We then remove duplicate advertisements with the same company, title, location, posting date and description. Advertisements posted on different dates remain separate observations at this stage.

We require a parsable posting date without applying a 90-day cutoff. The retained advertisements span 29 posting months from August 2023 to October 2026, although our scraper collected them between December 2025 and October 2026 with gaps in collection. The earlier posting dates belong to advertisements that were still available when the scraper retrieved them, so the dataset does not represent continuous collection throughout all 29 months.

We retain all job types during preprocessing and then classify occupations from job titles and advertised-role descriptions for the occupation comparisons. This gives 901 Machine Learning Engineer, 3,121 Data Scientist, 3,840 Data Analyst, 474 Business Analyst and 906 Software Engineer advertisements. Another 165 advertisements describe joint titles and 26,674 describe other or unspecified occupations, so we include them in the overall analysis without assigning them to one of the five occupation comparisons.

\subsection{Text Sections and Input Combinations}
A job description can discuss the team, the work and the applicant's background within the same paragraph. To examine these separately, we first divide the description into short passages and use GPT-6 Luna to classify each passage using the surrounding text for context. We retain passages labeled as Team context, Role task or Qualifications and exclude company promotion, employment terms, boilerplate and uncertain content.

When a passage mixes retained content with excluded material, GPT-6 Luna marks the boundaries using exact prefixes from the source text. We use these boundaries to retain the original text spans without paraphrasing. A passage containing more than one retained category receives the best-fitting label. Finally, we group the retained passages by their labels to form the three sections.

The following excerpts come from a General Motors trajectory-generation internship, posting ID \nolinkurl{4345263826}:
\begin{itemize}
\item \textbf{Team context:} ``The trajectory generation team (TGX) is responsible for turning scene understanding into wholistic AV behavior.'' This describes the team's purpose.
\item \textbf{Role task:} ``This might involve implementing and benchmarking new neural network architectures for behavior prediction, improving data pipelines for training trajectory models, or conducting ablation studies on different learning approaches.'' This describes the work that the applicant would perform.
\item \textbf{Qualifications:} ``Currently enrolled in a Masters program in Computer Science, Machine Learning, Robotics, or a related STEM field.'' This describes the background requested by the employer.
\end{itemize}

We join all passages belonging to a selected section before embedding, giving each advertisement one vector for that section. For combinations, we join the complete selected sections from the same advertisement and produce one vector for the combined text. An advertisement must contain every selected section to enter that comparison.

\begin{figure}[!tb]\centering
\includegraphics[width=\linewidth,height=.62\textheight,keepaspectratio]{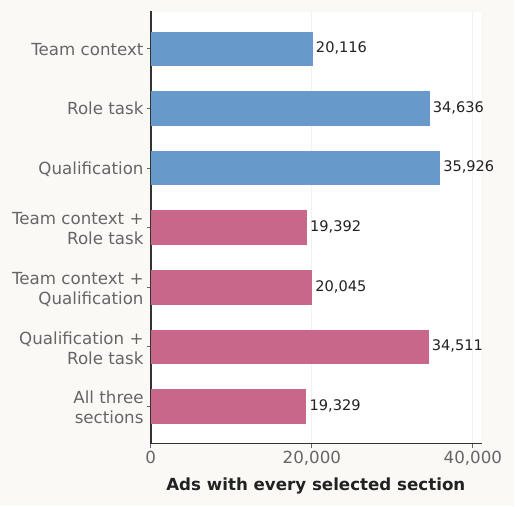}
\caption{A horizontal bar chart that counts advertisements containing every section required for each of the seven text inputs. Qualifications has the largest available sample at 35,926 job advertisements, followed by Role task at 34,636, and joining these two sections retains 34,511 advertisements. Team context is available in only 20,116 advertisements, so every combination containing it falls to around 19,000--20,000 advertisements. The all-three-sections input has the smallest sample at 19,329 advertisements. Thus, adding Team context changes both the text being compared and the number of advertisements available for the comparison.}
\label{fig:input_availability}\end{figure}

The seven segmentation text inputs are Team context, Role task, Qualifications, Team context + Role task, Team context + Qualifications, Role task + Qualifications and all three sections. We refer to the final combination as the full retained description because it contains all three selected sections. The original advertisement can also contain application instructions, benefits and company promotion, which are outside those sections.

\subsection{Split Dataset by Company Names}
We randomly assign 60\% of normalized company names to training, 20\% to validation and 20\% to testing using seed 42. All advertisements from one company name remain in the same set, so repeated employer wording cannot appear in both training and validation through that company. We mix months and occupations within the sets, obtaining 3,643 training companies, 1,214 validation companies and 1,214 test companies, containing 21,871, 7,524 and 6,686 advertisements respectively.

We fit dimensionality reduction and clustering on the training advertisements, select the PERC configuration using validation advertisements and evaluate the selected configuration once on test advertisements. This lets us examine whether clusters fitted to descriptions from one set of employers also group descriptions from held-out employers. We do not train the framework to predict the occupation labels.

Within each segmentation text input and set, identical joined text counts once per company. Every distinct training text has equal weight, so a company with more distinct descriptions contributes more observations. When measuring word separation, we instead select one advertisement per company using a reproducible sample. The number of test advertisements therefore depends on which sections are available and which duplicate descriptions have been removed.

After evaluating the framework, we use all job advertisements retained after preprocessing for the monthly and occupation analyses, with equal weight for each advertisement in MMD. These analyses are therefore not restricted to the 20\% of company names reserved for testing.

\subsection{Semantic Representation}\label{app:embeddings}
To represent the selected text, we embed each advertisement into one vector per available segmentation text input. We compare MiniLM-L6, MPNet-base, BGE-base-en-v1.5, E5-base-v2, EmbeddingGemma 2 and text-embedding-3-small. These embedding models let us compare compact sentence embeddings, retrieval embeddings and embedding models that accept longer descriptions under the inference settings listed below.

\begin{table*}[!tb]\centering\small

\caption{The settings used to turn job-ad text into vectors with six embedding models. Each row represents one embedding model, and the columns give its name, the number of values in its output vector, its maximum input tokens and its numerical precision during local inference. API output means that internal precision is unspecified, and we store those vectors as float32. MiniLM produces the shortest vectors at 384 dimensions and text-embedding-3-small the longest at 1,536, whereas EmbeddingGemma 2 and text-embedding-3-small accept much longer inputs than the four remaining embedding models.}

\begin{tabularx}{\linewidth}{>{\raggedright\arraybackslash}p{1.65in}>{\raggedright\arraybackslash}X>{\raggedright\arraybackslash}X>{\raggedright\arraybackslash}X}
\toprule

Embedding model & Dimensions & Token limit & Precision \\

\midrule

MiniLM-L6 & 384 & 256 & float32 \\

MPNet-base & 768 & 384 & float16 \\

BGE-base-en-v1.5 & 768 & 512 & float16 \\

E5-base-v2 & 768 & 512 & float16 \\

EmbeddingGemma 2, text only & 768 & 8,192 & bfloat16 \\

text-embedding-3-small & 1,536 & 8,191 & API output \\

\bottomrule\end{tabularx}\label{tab:original6}\end{table*}

When a description exceeds the embedding model's input limit, we divide it into chunks and embed each chunk separately. We normalize the chunk vectors, average them with weights proportional to their token counts and normalize the resulting vector. For E5, we add the prefix \nolinkurl{query:}, and for Gemma we add \texttt{task: clustering | query:}. The local chunk lengths leave room for these prefixes and the embedding model's special tokens.

Of the 19,329 advertisements containing all three sections, 65.6\% exceed MiniLM's input limit, 43.0\% exceed MPNet's, 22.0\% exceed BGE's and 22.2\% exceed E5's. Every description fits within the limits of EmbeddingGemma 2 and text-embedding-3-small. Consequently, the embedding model comparison also reflects differences in input length and the use of chunk averaging.

\subsection{Dimensionality Reduction and Clustering}\label{app:clustering}
For each embedding model, we compare the original vectors with PCA and UMAP at 10 and 50 components. We then apply K-means, diagonal Gaussian mixtures, hierarchical agglomerative clustering with Ward linkage and spectral clustering using four, six or eight clusters. We also test six HDBSCAN configurations, combining EOM or leaf selection with minimum cluster sizes of 10, 15 or 30 and a minimum of five samples. These methods group descriptions using cluster centers, Gaussian distributions, hierarchical distances, neighbor connections or local density \hyperlink{ref-16}{[16]}, \hyperlink{ref-17}{[17]}.

PCA uses full singular-value decomposition without whitening. UMAP uses 30 neighbors, a minimum distance of zero and cosine distance, with the same seed for fitting and transformation. K-means uses ten initializations. This produces 540 configurations per segmentation text input and 3,780 configurations across all seven inputs. Repeating the evaluation with seeds 42, 7, 19, 73 and 101 gives 18,900 records. We reuse repeated records for hierarchical agglomerative clustering and HDBSCAN when the original or PCA vectors are unchanged because those fits are deterministic. The selected framework must include dimensionality reduction, so the selection itself does not establish that reduction is necessary.

For hierarchical agglomerative clustering and spectral clustering, we construct a symmetric graph from the 30 nearest training neighbors. Neither method directly assigns a newly arriving advertisement to a fitted cluster, so we use a distance-weighted vote from its five nearest training neighbors in the selected representation. For spectral clustering, LOBPCG uses a tolerance of 0.00001 and seed 42, and we vary the seed of its K-means step. Gaussian mixtures use diagonal covariance, float64 fitting and three initializations. We check convergence and exclude numerical fits that fail.

To compare configurations, we measure assignment agreement across seeds using Adjusted Rand Index (ARI) and separation in the original embedding space using cosine silhouette on up to 1,000 validation texts. For K-means, Gaussian mixtures, hierarchical agglomerative clustering and spectral clustering, we calculate these scores before rejecting distant advertisements. For HDBSCAN, ARI includes the unassigned label and silhouette excludes unassigned advertisements.

We also measure word separation using TF-IDF cosine silhouette on one reproducibly selected advertisement per validation company, giving each cluster equal weight in the final score. The vocabulary and inverse document frequencies are fitted on training text only, excluding common career wording. This measures whether the clusters differ in their word use as well as their embedding vectors.

Under the criteria for selecting the right PERC framework configuration, we require a positive word-separation score, a mean ARI of at least 0.60 and cluster assignments for at least 60\% of validation texts. Each cluster must contain at least ten training companies, five validation companies and ten validation advertisements, and no training company can account for more than 20\% of that cluster. We then select the eligible configuration with the highest word-separation score, so a high score alone is insufficient when a cluster largely repeats one employer's descriptions.

For K-means, we reject assignments beyond the cluster's 95th percentile of training distances. For every method, we also reject advertisements below the cluster's fifth percentile of training cosine similarity in the original embedding space. These advertisements remain available for comparisons of the overall distributions. After selection, we evaluate the seed-42 framework on test companies without changing its configuration in response to those results.

\subsection{Distributional Comparison}\label{app:mmd}

\textbf{Population definition.} To define the distributional difference, let \(P\) and \(Q\) represent probability distributions over job-ad vectors in \(\mathbb{R}^d\). A positive-definite kernel \(k\) defines a reproducing kernel Hilbert space (RKHS) \(\mathcal{H}_k\) with feature map \(\phi(x)=k(x,\cdot)\). In this space, the kernel value is the inner product \(k(x,y)=\langle\phi(x),\phi(y)\rangle_{\mathcal H_k}\). We define the kernel means and Maximum Mean Discrepancy as

\begin{equation}\mu_P=\mathbb E_{x\sim P}[\phi(x)],\qquad \mu_Q=\mathbb E_{y\sim Q}[\phi(y)].\end{equation}

\begin{equation}\begin{aligned}\operatorname{MMD}_k(P,Q)&=\sup_{\lVert f\rVert_{\mathcal H_k}\leq1}\left|\mathbb E_P f-\mathbb E_Q f\right|\\&=\lVert\mu_P-\mu_Q\rVert_{\mathcal H_k}.\end{aligned}\end{equation}

MMD therefore measures the distance between the two kernel means. In its equivalent supremum definition, we restrict functions to an RKHS norm of at most one so that their values cannot be increased by arbitrary rescaling. The reproducing property gives \(\mathbb E_P f-\mathbb E_Q f=\langle f,\mu_P-\mu_Q\rangle_{\mathcal H_k}\), and the Cauchy--Schwarz inequality gives the norm expression above. When the means differ, equality is attained by \(f^*=(\mu_P-\mu_Q)/\lVert\mu_P-\mu_Q\rVert_{\mathcal H_k}\). With a Gaussian kernel, population MMD is zero exactly when the embedding distributions are equal. This equality describes the vectors produced by the embedding model, so it does not require the original wording of the advertisements to be identical \hyperlink{ref-11}{[11]}.

From this definition, the squared MMD is defined as the following sum of within-distribution and between-distribution similarities:

\begin{equation}\begin{aligned}\operatorname{MMD}_k^2(P,Q)={}&\langle\mu_P,\mu_P\rangle_{\mathcal H_k}+\langle\mu_Q,\mu_Q\rangle_{\mathcal H_k}\\&-2\langle\mu_P,\mu_Q\rangle_{\mathcal H_k}\\={}&\mathbb E_{x,x\prime\sim P}k(x,x\prime)\\&+\mathbb E_{y,y\prime\sim Q}k(y,y\prime)\\&-2\mathbb E_{x\sim P,y\sim Q}k(x,y).\end{aligned}\end{equation}

The first expectation measures similarity between two independent vectors from \(P\), and the second measures similarity between two independent vectors from \(Q\). The final expectation measures similarity between a vector from each distribution. We subtract twice this final term because \(\lVert u-v\rVert^2=\lVert u\rVert^2+\lVert v\rVert^2-2\langle u,v\rangle\). Thus, greater similarity between the distributions reduces the squared distance between their kernel means.

\textbf{Observed samples.} For the collected advertisements, we estimate each kernel mean by averaging the feature maps of the observed vectors. For samples \(X=\{x_i\}_{i=1}^n\) and \(Y=\{y_j\}_{j=1}^m\), these estimates are defined as

\begin{equation}\widehat\mu_X=\frac1n\sum_{i=1}^{n}\phi(x_i),\qquad\widehat\mu_Y=\frac1m\sum_{j=1}^{m}\phi(y_j).\end{equation}

Expanding the squared distance between these estimates gives the V-statistic reported in Section~\ref{sec:methodology}. We calculate it from kernel values without constructing the potentially infinite-dimensional feature map. The embedding model first produces each job-ad vector \(x\), and the radial basis function (RBF) kernel then measures similarity between those vectors. A linear kernel would compare only the ordinary vector means, but the Gaussian kernel can distinguish distributions whose ordinary means are equal.

\textbf{Kernel scale and computation.} To set the Gaussian kernel scale, we take at most 2,000 finite original embedding vectors from training-company advertisements, sampling with seed 42 when necessary. For these vectors \(\{z_r\}_{r=1}^{L}\) with \(L\leq2000\), we define

\begin{equation}\begin{aligned}\mathcal D&=\{\lVert z_r-z_s\rVert_2^2:r<s,\ z_r\ne z_s\},\\\gamma&=\bigl[\operatorname{median}(\mathcal D)\bigr]^{-1}.\end{aligned}\end{equation}

We exclude identical vector pairs when estimating the scale and require at least one positive squared distance. With this choice, a pair at the median squared distance has kernel similarity \(e^{-1}\), equivalent to \(\sigma^2=\operatorname{median}(\mathcal D)/2\) in the usual Gaussian-kernel notation. We keep the same scale for every month, occupation and cluster pair within a segmentation text input. MMD values from different embedding models, inputs or bandwidths therefore should not be interpreted as sharing one common effect-size scale.

We calculate kernel values in blocks of at most 512 vectors from each sample to reduce memory use, evaluating every pair, including comparisons of each vector with itself. The three kernel matrices contain \(n^2+m^2+nm\) entries in total, and each distance uses all \(d\) coordinates. Dividing the computation into blocks leaves the score unchanged. The resulting monthly matrix is symmetric, with zero on its diagonal because each sample is compared with itself.

\textbf{Finite-sample bias.} Including comparisons of each vector with itself makes the V-statistic nonnegative but also introduces finite-sample bias. The following relationship assumes that the two samples are independent, observations within each sample are independently and identically distributed, and the kernel scale is fixed. Define \(a_P=\mathbb E_{x,x\prime\sim P}k(x,x\prime)\) and \(a_Q=\mathbb E_{y,y\prime\sim Q}k(y,y\prime)\). Because \(k(x,x)=1\), counting the diagonal and off-diagonal terms gives

\begin{equation}\begin{aligned}\mathbb E\!\left[\widehat{\operatorname{MMD}}_{V,k}^2\right]&=\operatorname{MMD}_k^2(P,Q)\\&\quad+\frac{1-a_P}{n}+\frac{1-a_Q}{m}.\end{aligned}\end{equation}

For the RBF kernel, \(0\leq a_P,a_Q\leq1\), so the bias is nonnegative and becomes smaller as the sample sizes increase. Two finite samples from the same distribution can therefore have a positive expected squared MMD even when population MMD is zero. This relationship assumes independent observations. Our advertisements can contain repeated descriptions, and we select comparisons with large scores, so we use the ranking to choose descriptions for inspection without treating it as evidence of statistical significance.

If we exclude comparisons of each vector with itself and adjust the denominators, the corresponding U-statistic for \(n,m>1\) is defined as

\begin{equation}\begin{aligned}\widehat{\operatorname{MMD}}_{U,k}^2={}&\frac{1}{n(n-1)}\sum_{i\ne i\prime}k(x_i,x_{i\prime})\\&+\frac{1}{m(m-1)}\sum_{j\ne j\prime}k(y_j,y_{j\prime})\\&-\frac{2}{nm}\sum_{i=1}^{n}\sum_{j=1}^{m}k(x_i,y_j).\end{aligned}\end{equation}

The U-statistic is unbiased under the stated sampling assumptions but can be negative for a finite sample. We use the V-statistic in the reported matrices and narrative rankings. Requiring at least 30 advertisements from five company names in each group gives each narrative comparison more observations, although it does not remove the V-statistic's bias or make its ranking a statistical test.

To visualize the distributions, we fit two PCA components on training advertisements and project every month using the same components. The histograms use common bins and each density integrates to one. The plotted distributions can overlap even when MMD is large because the score uses every original embedding dimension and the figure shows only two PCA components.

\subsection{Cluster Selection and Narrative Analysis}
For each occupation and segmentation text input, we calculate MMD between eligible months within the same cluster. We retain the month pair with the largest MMD for each cluster and then select up to three distinct clusters. Meanwhile, to compare different clusters, we calculate MMD between cluster pairs within the same month, retain the largest month-specific difference for each pair and select up to three pairs.

Each side of a comparison must contain at least 30 advertisements and five recorded company names. We count the company names available in the dataset, which can still include employer aliases or publishing services. The occupation comparisons use the clusters fitted across all job types for that segmentation text input, so we do not fit a new clustering framework for each occupation.

After selecting the comparisons, we give GPT-6.1 Terra every original job description from the two selected groups, together with the passages used for clustering. We label each advertisement with its occupation, cluster, posting month, company and posting ID so Terra can identify which group it belongs to. We send each comparison in one request without sampling advertisements or dividing the clusters into batches, and ask for differences and similarities in Team context, Role task and Qualifications.

For each finding, GPT-6.1 Terra must quote an advertisement from each group. We check these quotations and their cluster membership against the source descriptions. Across 81 responses, three quotations and three inline source IDs required corrections in five responses, and we retain both the original and corrected versions. These checks confirm where a statement comes from, although they do not measure how common it is across the cluster. Two additional comparisons exceed GPT-6.1 Terra's context limit and remain unrun.

\section{Complete Configuration Ablations}\label{app:ablations}

\subsection{Selected Configurations and Held-Out Evaluation}
Across the seven configuration searches, we record 18,900 results covering five random seeds, with no numerical fit failures. Of the 3,780 configurations, 542 meet the criteria for selecting the right PERC framework configuration, with 54 to 102 eligible configurations per segmentation text input. We exclude the original-vector configurations from selection because the selected framework is required to use dimensionality reduction. Table~\ref{tab:original7} lists the highest-ranked eligible configuration for each input.

\begin{table*}[!tb]\centering\small

\caption{This table demonstrates the selected PERC configurations for the seven segmentation text inputs. Each row represents one input, and the columns give its section combination, embedding model, reduction method and clustering setting. The number after PCA or UMAP represents the reduced vector dimension, and the number in parentheses represents the number of clusters. Five inputs use PCA and two use UMAP, with hierarchical agglomerative clustering selected for four inputs. The full retained job description joins Team context, Role task and Qualifications.}

\begin{tabularx}{\linewidth}{>{\raggedright\arraybackslash}p{1.65in}>{\raggedright\arraybackslash}X>{\raggedright\arraybackslash}X>{\raggedright\arraybackslash}X}
\toprule

Segmentation text input & Embedding model & Reduction & Clustering setting \\

\midrule

Team context & MPNet & PCA 50 & Hierarchical agglomerative clustering (8 clusters) \\

Role task & text-embedding-3-small & UMAP 50 & K-means (4 clusters) \\

Qualifications & text-embedding-3-small & PCA 10 & Hierarchical agglomerative clustering (4 clusters) \\

Team context + Role task & MPNet & PCA 10 & Spectral (4 clusters) \\

Team context + Qualifications & MiniLM & PCA 50 & Hierarchical agglomerative clustering (4 clusters) \\

Qualifications + Role task & BGE & UMAP 10 & K-means (4 clusters) \\

Full retained job description & MiniLM & PCA 50 & Hierarchical agglomerative clustering (4 clusters) \\

\bottomrule\end{tabularx}\label{tab:original7}\end{table*}

We select PCA for five inputs and UMAP for two, together with hierarchical agglomerative clustering for four inputs, K-means for two and spectral clustering for one. MPNet, MiniLM and text-embedding-3-small are each selected for two inputs, and BGE is selected for the remaining input. E5 and Gemma also produce eligible configurations but do not obtain the highest eligible score for any input.

\begin{table*}[!tb]\centering\small

\caption{The validation measurements for the selected PERC configurations, where each row represents one segmentation text input. The columns give the input, Adjusted Rand Index (ARI) across repeated runs, the percentage of advertisements assigned to a cluster, TF-IDF cosine silhouette as Word separation, and silhouette in the original embeddings as Vector silhouette. Role task + Qualifications has the highest word separation at 0.0641, and Team context + Role task has the highest vector silhouette at 0.0906, so the strongest separation depends on which representation we measure. Values are averaged across the recorded runs.}

\begin{tabularx}{\linewidth}{>{\raggedright\arraybackslash}p{1.65in}>{\raggedright\arraybackslash}X>{\raggedright\arraybackslash}X>{\raggedright\arraybackslash}X>{\raggedright\arraybackslash}X}
\toprule

Segmentation text input & ARI & Assigned (\%) & Word separation & Vector silhouette \\

\midrule

Team context & 1.0000 & 93.4 & 0.0115 & 0.0143 \\

Role task & 0.7806 & 93.4 & 0.0443 & 0.0686 \\

Qualifications & 1.0000 & 94.0 & 0.0227 & 0.0818 \\

Team context + Role task & 0.9993 & 93.8 & 0.0116 & 0.0906 \\

Team context + Qualifications & 1.0000 & 93.9 & 0.0077 & 0.0321 \\

Qualifications + Role task & 0.9765 & 90.8 & 0.0641 & 0.0702 \\

Full retained job description & 1.0000 & 93.8 & 0.0124 & 0.0720 \\

\bottomrule\end{tabularx}\label{tab:original8}\end{table*}

The four selected hierarchical agglomerative configurations have an ARI of one because PCA and the clustering algorithm repeat the same assignments on unchanged data. For spectral clustering, agreement across seeds reflects its K-means step on unchanged graph coordinates. Role task has the lowest selected ARI at 0.7806. These scores describe sensitivity to random initialization on the same sample, so we also evaluate the selected configurations on advertisements from held-out companies.

The original-vector silhouettes range from 0.0143 to 0.0906 and the word-separation scores are also small, indicating substantial overlap between the descriptions. The clusters therefore help us organize related advertisements for reading, although the measurements do not establish sharply separated professions.

\begin{table*}[!tb]\centering\small

\caption{The selected PERC configurations evaluated on advertisements from held-out companies, where each row represents one segmentation text input. Test ads counts distinct job-description texts separately for each company, Test companies counts company names, Assigned gives the percentage of advertisements assigned to a cluster, and Word separation is TF-IDF cosine silhouette. Between 91.2\% and 94.9\% of these advertisements receive cluster assignments, and Role task + Qualifications has the highest word-separation score at 0.0765. Thus, this input also produces the strongest separation in word use among the selected configurations on held-out companies.}

\begin{tabularx}{\linewidth}{>{\raggedright\arraybackslash}p{1.65in}>{\raggedright\arraybackslash}X>{\raggedright\arraybackslash}X>{\raggedright\arraybackslash}X>{\raggedright\arraybackslash}X}
\toprule

Segmentation text input & Test ads & Test companies & Assigned (\%) & Word separation \\

\midrule

Team context & 2981 & 920 & 94.0 & 0.0137 \\

Role task & 5551 & 1200 & 94.0 & 0.0599 \\

Qualifications & 5776 & 1203 & 91.2 & 0.0188 \\

Team context + Role task & 2983 & 909 & 94.2 & 0.0061 \\

Team context + Qualifications & 3120 & 910 & 94.1 & 0.0067 \\

Qualifications + Role task & 5645 & 1191 & 92.6 & 0.0765 \\

Full retained job description & 3016 & 901 & 94.9 & 0.0117 \\

\bottomrule\end{tabularx}\label{tab:original9}\end{table*}

On the test companies, 91.2\% to 94.9\% of advertisements receive cluster assignments and every input has positive word separation. However, Team context + Role task falls from 0.0116 on validation companies to 0.0061 on test companies despite near-perfect agreement across seeds. This shows why repeating assignments on the same data and separating descriptions from held-out companies are different aspects of the evaluation. Neither measurement represents accuracy in predicting occupation labels.

\subsection{How Embedding, Reduction and Clustering Affect the Results}
To examine the effect of the embedding model, we first use PCA with ten components and K-means with four clusters for every embedding model. Under this configuration, MiniLM has the highest Role task word-separation score at 0.0056 and text-embedding-3-small scores near zero. After searching the reduction and clustering configurations for each embedding model, text-embedding-3-small reaches 0.0443 with UMAP at 50 dimensions and becomes the selected Role task embedding model, compared with MiniLM's best eligible score of 0.0104. The embedding model ranking therefore changes when we configure the later stages separately for each embedding model.

\begin{figure*}[!tb]\centering
\includegraphics[width=\textwidth,height=.62\textheight,keepaspectratio]{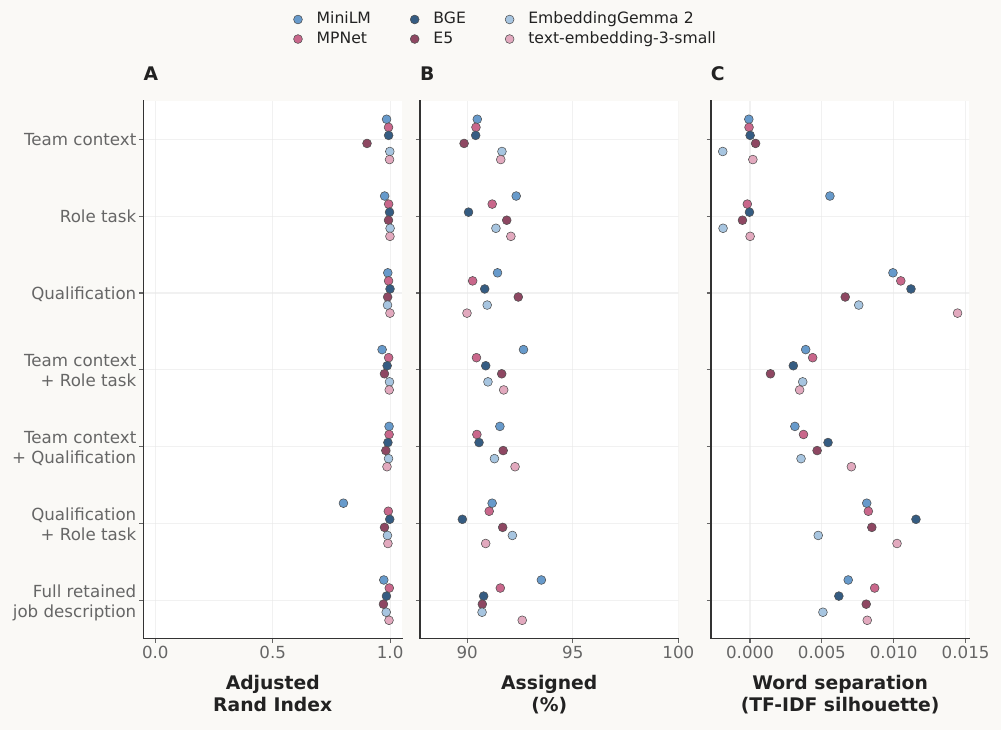}
\caption{A series of dot plots comparing six embedding models under PCA at 10 dimensions and four K-means clusters. Panels A, B and C report agreement across runs, the percentage of advertisements assigned to clusters and TF-IDF cosine silhouette. Panel C shows MiniLM leading on Role task at 0.0056, whereas the other embedding models score near zero or below. For Qualifications, text-embedding-3-small leads at 0.0144 and E5 scores lowest at 0.0066. Team context scores remain close to zero for every embedding model. Panels A and B show generally high agreement across runs and assignment rates around 90\%--94\%, so these measures vary less than the separation of word use in panel C.}
\label{fig:expanded_embedding_ablation}\end{figure*}

\begin{figure*}[!tb]\centering
\includegraphics[width=\textwidth,height=.62\textheight,keepaspectratio]{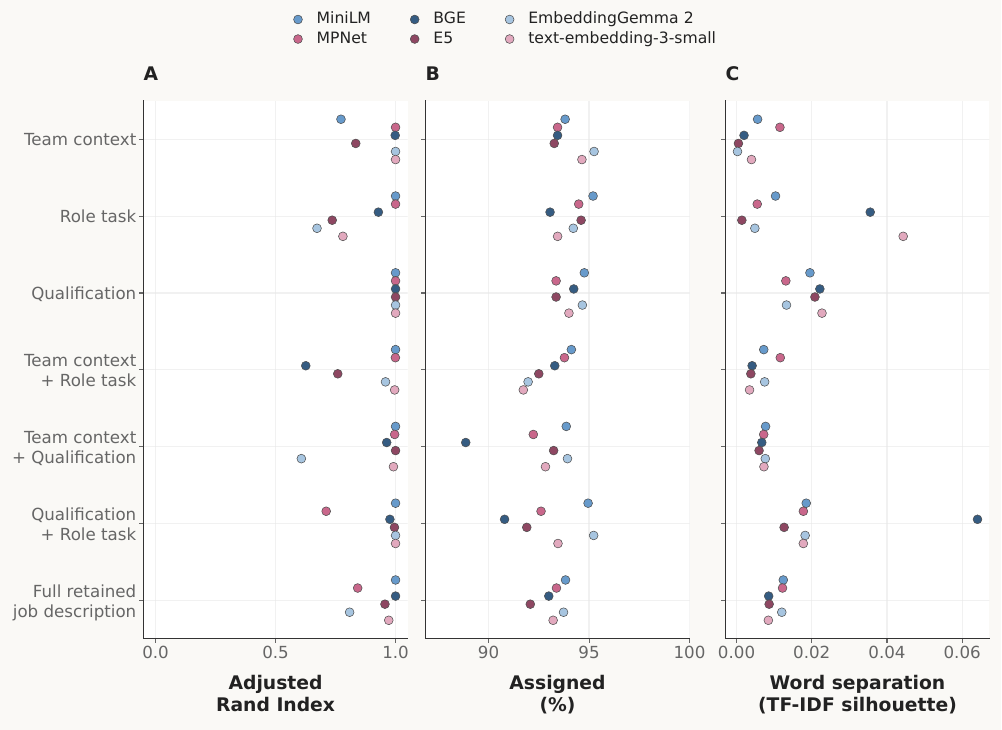}
\caption{A comparison of the best eligible configuration for each embedding model and text input, presented as three dot plots. Panels A, B and C report agreement across runs, the percentage of advertisements assigned to clusters and TF-IDF cosine silhouette. Text-embedding-3-small leads on Role task at 0.0443, followed by BGE at 0.0355, whereas MiniLM reaches 0.0104. BGE produces the largest score in the figure at 0.0641 for Qualifications + Role task. MPNet leads on Team context and Team context + Role task, whereas MiniLM narrowly leads when all three sections are joined. The different winners show that the strongest separation depends on the text input and its accompanying reduction and clustering methods.}
\label{fig:expanded_tuned_embeddings}\end{figure*}

For all three joined sections, Gemma reaches a best eligible word-separation score of 0.0120 compared with MiniLM's 0.0124. Although Gemma accepts the complete descriptions without chunking, this longer input limit does not give it the highest score in this comparison. These measurements describe the resulting clusters and do not establish how much information each embedding model preserves from every description.

To examine dimensionality reduction, we use MPNet and four K-means clusters throughout the comparison. For Team context, word separation increases from near zero with PCA at ten dimensions to 0.0041 with UMAP at 50 dimensions. For all three joined sections, the direction reverses, with PCA at ten dimensions scoring 0.0087 and UMAP at 50 dimensions scoring 0.0077. Thus, the reduction method that gives greater separation depends on which parts of the advertisement we embed.

\begin{figure*}[!tb]\centering
\includegraphics[width=\textwidth,height=.62\textheight,keepaspectratio]{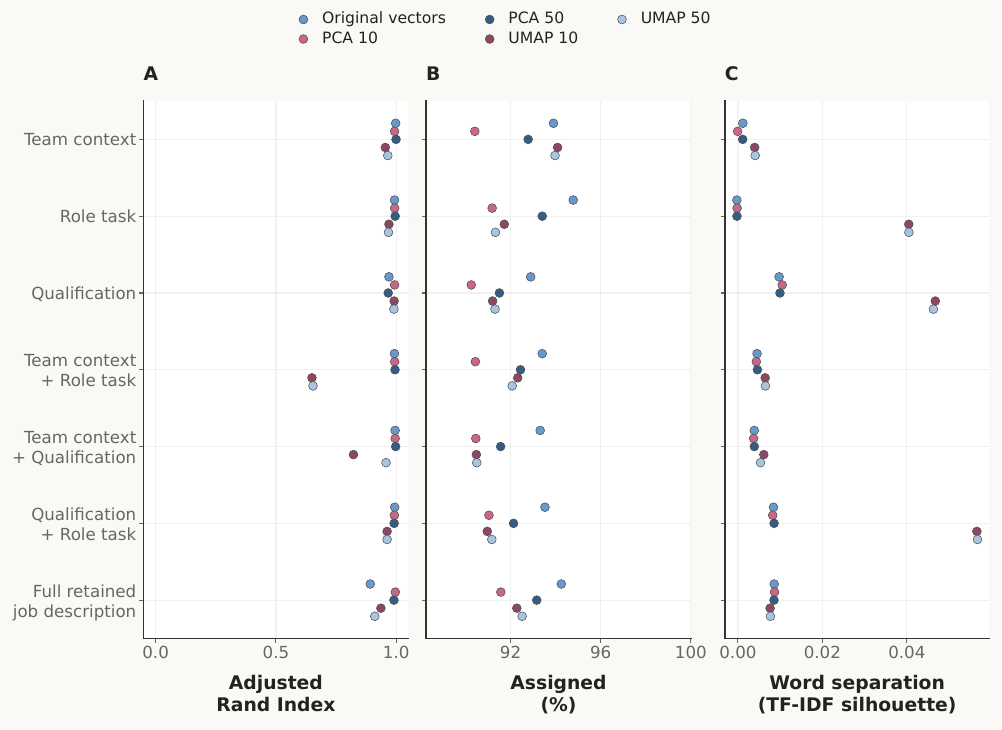}
\caption{A three-panel dot plot comparison of the original MPNet vectors, PCA and UMAP under four K-means clusters. Panel A shows agreement across runs, B shows the percentage of advertisements assigned to clusters and C shows TF-IDF cosine silhouette. Panel C shows the largest increases in TF-IDF cosine silhouette with UMAP for Role task, Qualifications and their combination, which reaches about 0.0567 compared with about 0.0085 for PCA or the original vectors. Increasing UMAP from 10 to 50 dimensions adds little. Joining all three sections reverses the ranking slightly, with PCA near 0.0086 and UMAP near 0.0077. Panel A also shows lower agreement for UMAP on Team context + Role task, and panel B shows that these increases in cosine silhouette do not consistently increase the percentage of advertisements assigned to clusters.}
\label{fig:expanded_reduction_ablation}\end{figure*}

We also check whether a high word-separation score comes from advertisements across several companies. For example, MPNet with UMAP at 50 dimensions and four clusters scores 0.0405 on Role task, but one company accounts for 26.1\% of a training cluster, exceeding our 20\% limit. We therefore exclude this configuration from selection even though its plotted score is high.

Finally, we compare clustering methods using MPNet and PCA at 50 dimensions, showing the highest word-separation score among the tested configurations for each method. The complete search selects hierarchical agglomerative clustering, K-means and spectral clustering, and retains Gaussian mixtures and HDBSCAN as comparison methods. These results apply to the cluster counts, neighbor graphs, assignment procedures and company requirements tested here.

\begin{figure*}[!tb]\centering
\includegraphics[width=\textwidth,height=.62\textheight,keepaspectratio]{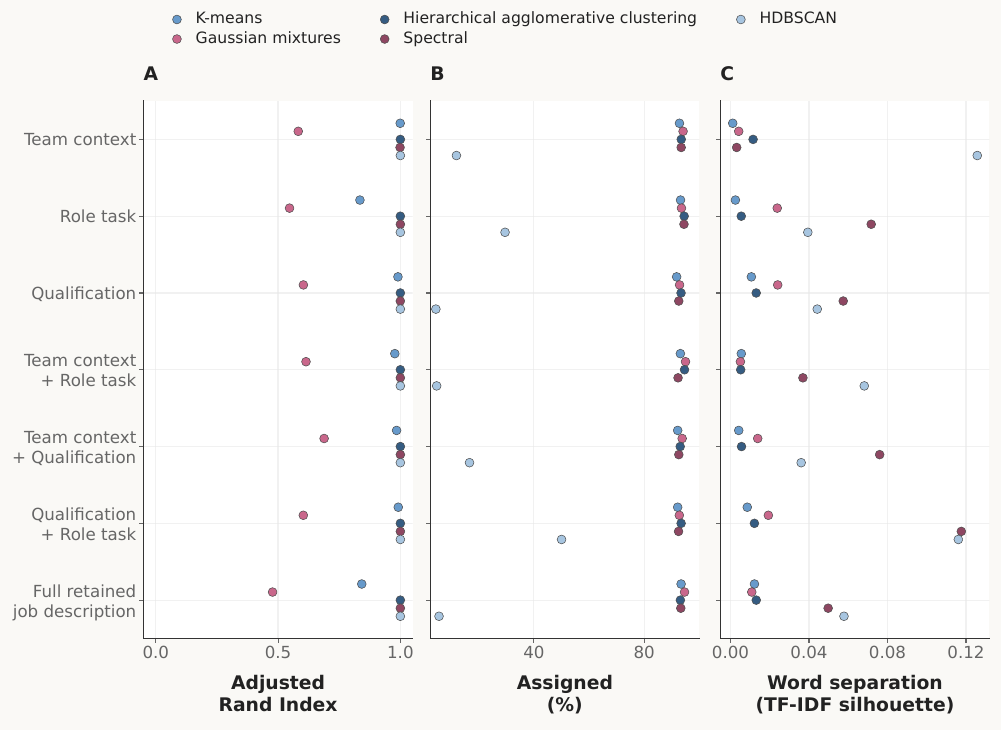}
\caption{A three-panel comparison of clustering methods presented as dot plots using MPNet and 50-dimensional PCA. Panels A, B and C report agreement across runs, the percentage of advertisements assigned to clusters and TF-IDF cosine silhouette. Each point shows that method's highest-scoring tested configuration. HDBSCAN reaches a Team context silhouette of 0.126 but assigns only 11.9\% of advertisements. Spectral clustering separates task and qualification wording more strongly than K-means or hierarchical agglomerative clustering and assigns about 93\% of advertisements. Gaussian mixtures have the lowest agreement across runs at ARI 0.48--0.69.}
\label{fig:expanded_clustering_ablation}\end{figure*}

Together, the ablations show that text selection, embedding, reduction and clustering affect one another. A higher word-separation score measures more distinct word use, although it does not directly measure whether the resulting interpretation is useful. The seven segmentation text inputs also contain different advertisements because their section requirements differ. In the separate comparison of PERC, SAE and text-only inputs, we use the same advertisements for all three conditions, but we do not compare all seven segmentation inputs on one common sample.

\section{Complete-Cluster Case Studies}\label{app:cases}

\subsection{Comparing the Selected Clusters}
We apply the seven selected PERC configurations in Table~\ref{tab:original7} to the monthly and occupation analyses, calculating MMD from the original embedding vectors and using the fitted clusters to organize the descriptions. GPT-6.1 Terra reads 81 comparisons, including 49 comparisons of the same cluster across months and 32 comparisons of different clusters within one month. Of these, 80 describe Data Scientist, Data Analyst, Machine Learning Engineer or Business Analyst advertisements and one combines all occupations. Table~\ref{tab:original10} gives the counts for each occupation.

\begin{figure*}[!tb]\centering
\includegraphics[width=\textwidth,height=.62\textheight,keepaspectratio]{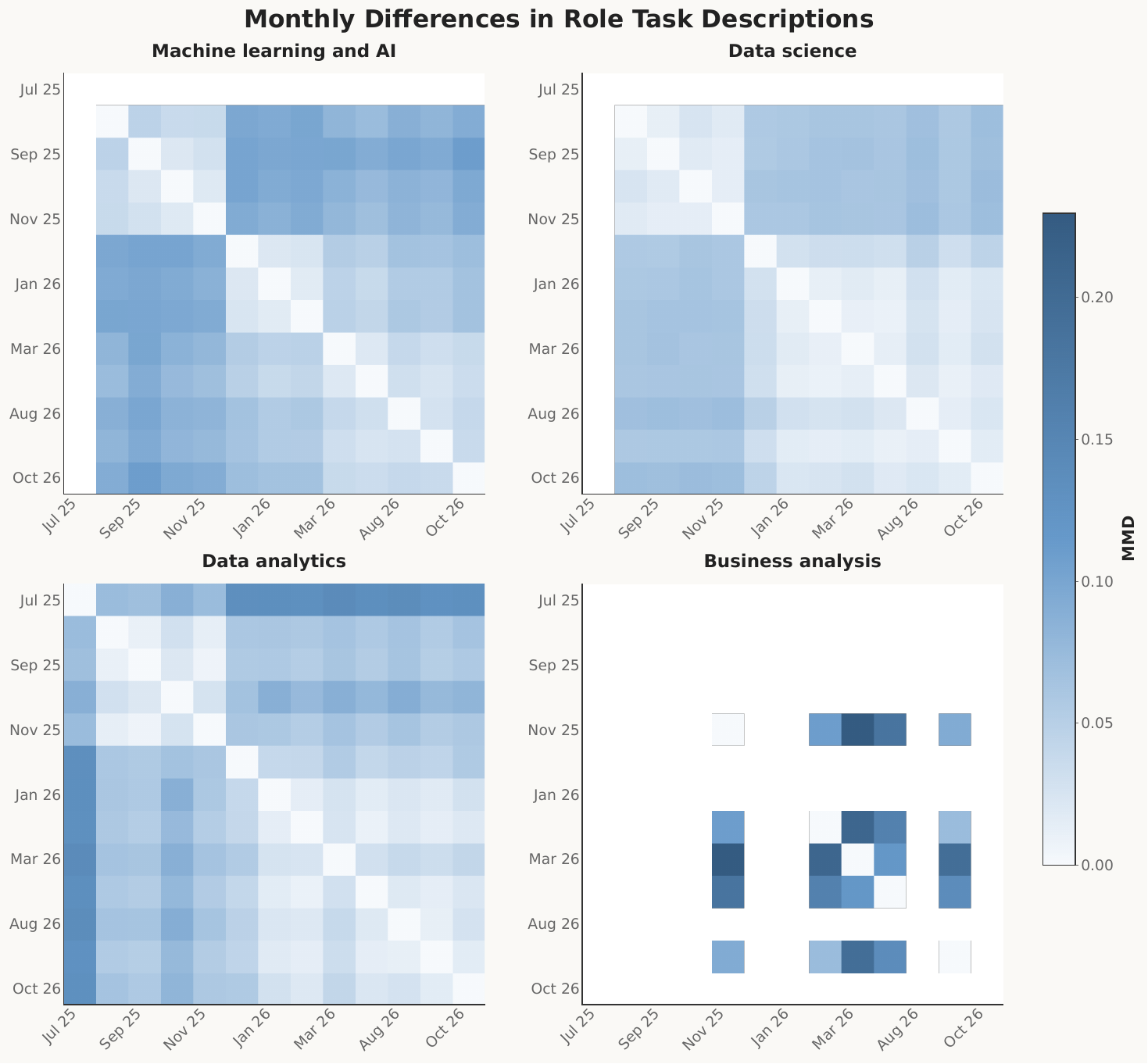}
\caption{A set of MMD matrices comparing monthly Role task vector distributions within each occupation. Darker cells indicate larger MMD scores and blank cells mark months with fewer than 30 advertisements. The darker bands in the Machine Learning Engineer and Data Scientist panels separate August--November 2025 descriptions from later descriptions, whereas Data Analyst also shows a strong contrast between July 2025 and later months. The largest displayed difference occurs in Business Analyst between November 2025 and March 2026, with MMD 0.230. Business Analyst has far fewer available month pairs than the other occupations.}
\label{fig:monthly_mmd}\end{figure*}

\begin{table*}[!tb]\centering\small

\caption{The completed GPT-6.1 Terra cluster comparisons by occupation. For each occupation and the combined sample, we count comparisons across months within the same cluster and comparisons between clusters within the same month. The final row sums these counts, giving 49 monthly and 32 same-month comparisons. Data Scientist and Data Analyst advertisements account for 63 of the 81 comparisons, so most of these analyses of all advertisements in the selected groups describe work within those two occupations.}

\begin{tabularx}{\linewidth}{>{\raggedright\arraybackslash}p{1.65in}>{\raggedright\arraybackslash}X>{\raggedright\arraybackslash}X>{\raggedright\arraybackslash}X}
\toprule

Occupation & Within cluster, different months & Different clusters, same month & Total \\

\midrule

Machine learning and AI & 9 & 0 & 9 \\

Data science & 16 & 14 & 30 \\

Data analytics & 17 & 16 & 33 \\

Business analysis & 6 & 2 & 8 \\

All occupations together & 1 & 0 & 1 \\

Total & 49 & 32 & 81 \\

\bottomrule\end{tabularx}\label{tab:original10}\end{table*}

Across these comparisons, GPT-6.1 Terra produces 347 findings with source quotations at a recorded token cost of \$21.52. Some findings recur across segmentation text inputs, so the total does not represent 347 independent discoveries. The completed all-occupation comparison uses Team context + Role task. The additional comparisons using Team context + Qualifications and all three sections exceed the context limit and remain unrun.

The following examples explain how the selected clusters differ in Team context, Role task and Qualifications. We also examine repeated recruitment and differences in description detail because these can affect the embedding distributions. MMD ranks the observed differences without establishing their statistical significance.

\begin{figure}[!tb]\centering
\includegraphics[width=\linewidth,height=.62\textheight,keepaspectratio]{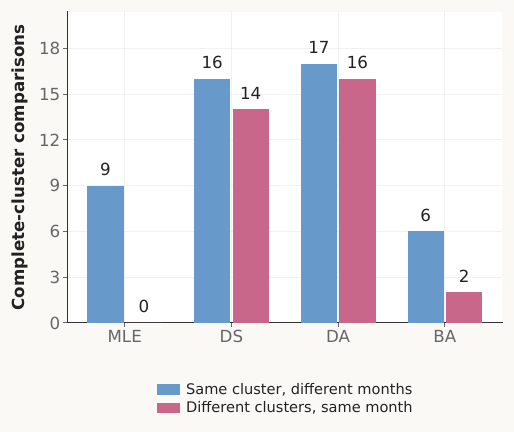}
\caption{A grouped bar chart that compares how many comparisons use all advertisements in the selected groups for each occupation, where blue represents comparisons across months and pink represents comparisons between different clusters within one month. For Data Analyst advertisements, we complete 17 monthly and 16 same-month comparisons, followed by 16 and 14 for Data Scientist advertisements. For Machine Learning Engineer advertisements, we complete nine monthly comparisons, and for Business Analyst advertisements we complete six monthly and two same-month comparisons.}
\label{fig:narrative_coverage}\end{figure}

\subsection{Monthly Differences Within the Same Cluster}

\begin{figure*}[!tb]\centering
\includegraphics[width=\textwidth,height=.62\textheight,keepaspectratio]{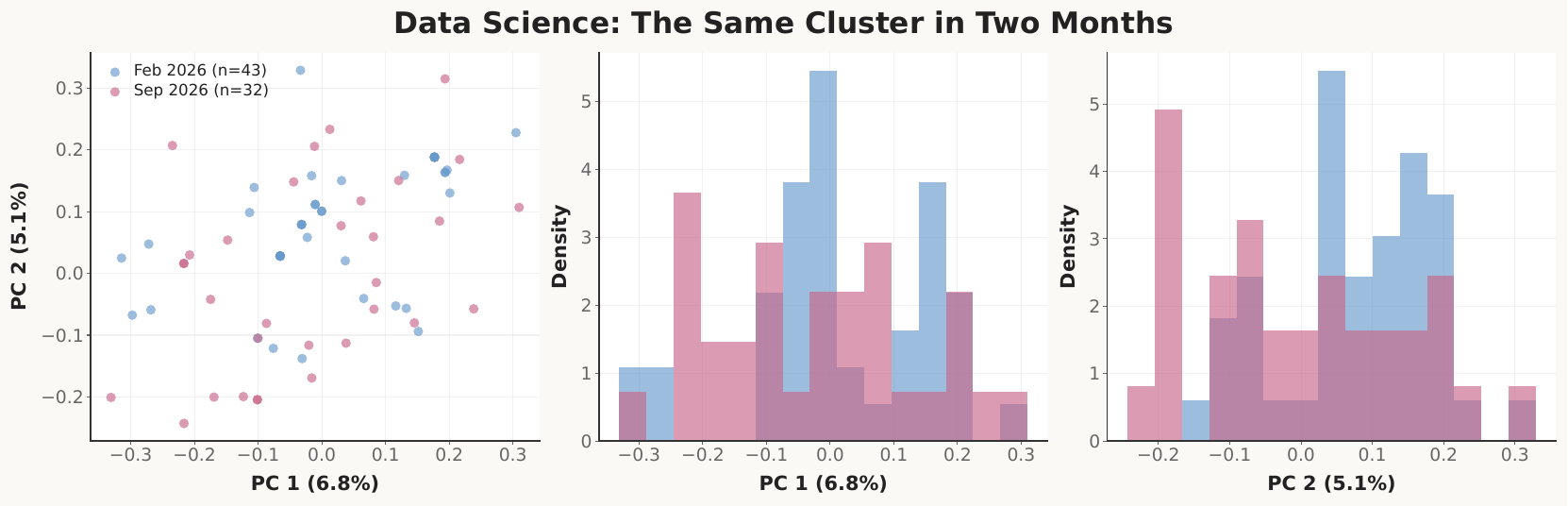}
\caption{A PCA scatter plot and two histograms comparing Data Scientist all-sections cluster 0 in February and September 2026. Blue represents February and pink represents September, and the plots use the same PCA components. The February and September advertisements overlap in the scatter plot, but their histograms show different concentrations. September has more density near $-0.2$ on both axes, whereas February is more concentrated near zero on PC 1 and above zero on PC 2. We therefore see changes in the distribution within the same Data Scientist cluster without a complete separation of the points. MMD is 0.0707 across all original embedding dimensions, and the first and second PCA components explain 6.8\% and 5.1\% of the variance in the job-ad vectors used to fit PCA.}
\label{fig:ds_full_distributions}\end{figure*}

\begin{figure*}[!tb]\centering
\includegraphics[width=\textwidth,height=.62\textheight,keepaspectratio]{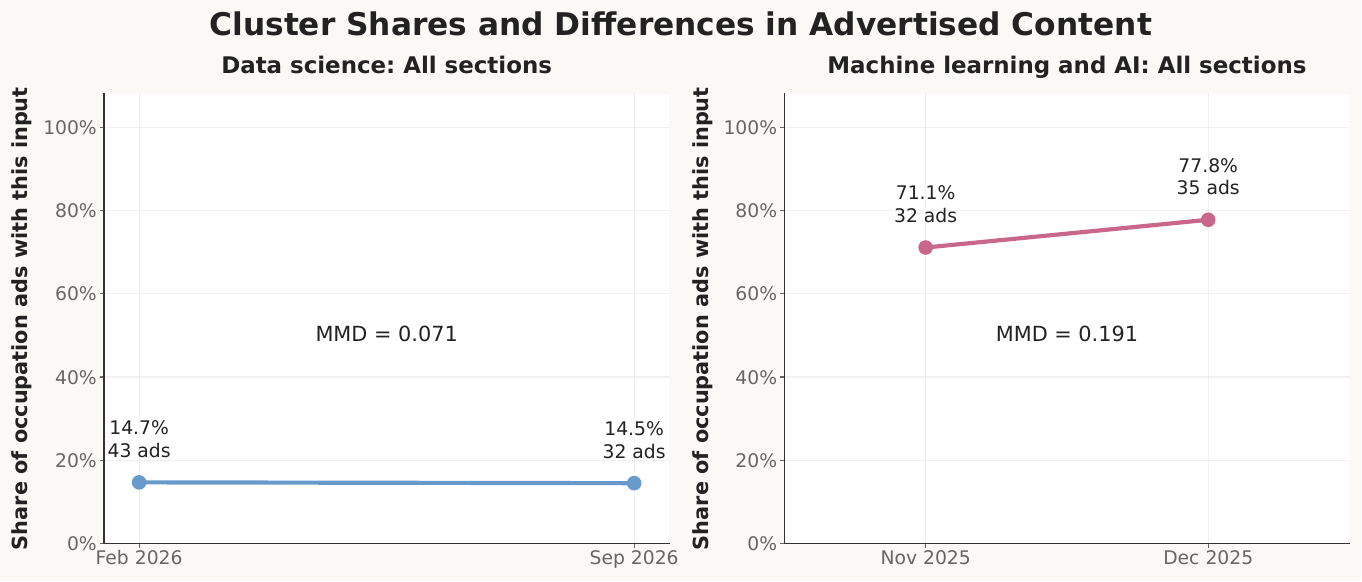}
\caption{A pair of line charts that tracks the share of advertisements in a selected Data Scientist or Machine Learning Engineer cluster across two months. Counts appear beside the points, and MMD measures differences between the all-sections embedding distributions. The Data Scientist line stays almost flat at 14.7\% in February and 14.5\% in September even as the number of advertisements falls from 43 to 32. Its MMD of 0.071 records a difference in description vectors that the nearly unchanged share does not express. The Machine Learning Engineer line rises from 71.1\% to 77.8\% between November and December alongside MMD 0.191. The Data Scientist and Machine Learning Engineer comparisons shown in the two panels distinguish changes in a cluster's share of advertisements from changes in the descriptions within that cluster.}
\label{fig:temporal_shares}\end{figure*}

For Data Scientist advertisements with all three sections, cluster 0 contains 43 postings from 27 company names in February 2026 and 32 postings from 29 company names in September. Its share of the available Data Scientist advertisements stays close to the same level, at 14.7\% and 14.5\%, although MMD is 0.0707. As shown in Figures~\ref{fig:ds_full_distributions} and~\ref{fig:temporal_shares}, the descriptions can differ even when the cluster contains a similar share of the occupation's advertisements.

In February, HNTB describes infrastructure inventories for maps and planning, and Publicis describes search-performance reporting. In September, Gallo describes satellite and drone data for vineyard analysis, Los Alamos describes remote-sensing research and Corteva describes AI-assisted workflow recommendations. Planning and work with ``machine learning models'' appear in both months, but the advertisements apply them to different problems. The source postings are \nolinkurl{4370805700}, \nolinkurl{4368139547}, \nolinkurl{4462926667}, \nolinkurl{4462718150} and \nolinkurl{4460766364}.

For Machine Learning Engineer advertisements, all-sections cluster 1 contains 32 November 2025 postings from nine company names and 35 December postings from 16 company names. Its share increases from 71.1\% to 77.8\% and MMD is 0.1912. November includes repeated OneStream descriptions of forecasting, anomaly detection and financial-software development. December includes Kensho's retrieval and financial-information applications, MITRE's data pipelines and General Motors' use of ``machine learning models'' for trajectory prediction with driving data and simulation. These descriptions show different advertised applications, although they do not follow the same positions over time. The source postings include \nolinkurl{692acf05508b54b5d8d2e695}, \nolinkurl{4329583848}, \nolinkurl{4325487180} and \nolinkurl{4345263826}.

\subsection{Data Scientist Clusters Differ in Their Use of Machine Learning Models}
For February 2026, Data Scientist Team context clusters 2 and 3 contain 64 and 39 advertisements from 17 and 13 company names respectively, with MMD 0.4155. Cluster 2 describes engineering, business and research applications, and cluster 3 places greater emphasis on healthcare and life-science work.

For example, Impinj's advertisement in cluster 2 describes algorithms for product quality and manufacturing yield, whereas AbbVie's advertisement in cluster 3 describes annotating sequencing variants to study their functional effects in disease (\nolinkurl{4361074947}, \nolinkurl{4290822999}). KLA describes spare-parts demand forecasting and Ascension describes healthcare throughput, access and labor management (\nolinkurl{4347725262}, \nolinkurl{4368133772}). Although these roles can use similar statistical and machine learning methods, the descriptions differ in the domain data and decisions that the statistical and ``machine learning models'' support.

The clusters also overlap in their commercial and operational tasks. Otsuka's healthcare advertisement describes prescription-pattern analysis for product adoption, and Ascension describes service operations. We therefore interpret the separation as a difference in the emphasis of the Team context rather than evidence of two exclusive professions.

\subsection{Data Analyst Clusters Include Business and Healthcare Applications}
For February 2026, Data Analyst all-sections clusters 0 and 3 contain 93 and 39 advertisements from 44 and 11 company names respectively, with MMD 0.2100. Cluster 0 covers several business and operational settings, and cluster 3 contains a greater emphasis on healthcare, life science and medical data.

For example, Altar'd State describes inventory management and forecasting, whereas CareFirst connects reporting to care costs, provider performance and clinical outcomes (\nolinkurl{4368022621}, \nolinkurl{4377578541}). EBSCO describes interface experiments to inform product decisions, and Johnson \& Johnson describes examining clinical characteristics, population-level effects and patient-level predictions (\nolinkurl{4373949976}, \nolinkurl{4372168521}). Reporting and experiments can appear in both clusters, but the descriptions apply them to different analytical questions.

\begin{figure*}[!tb]\centering
\includegraphics[width=\textwidth,height=.62\textheight,keepaspectratio]{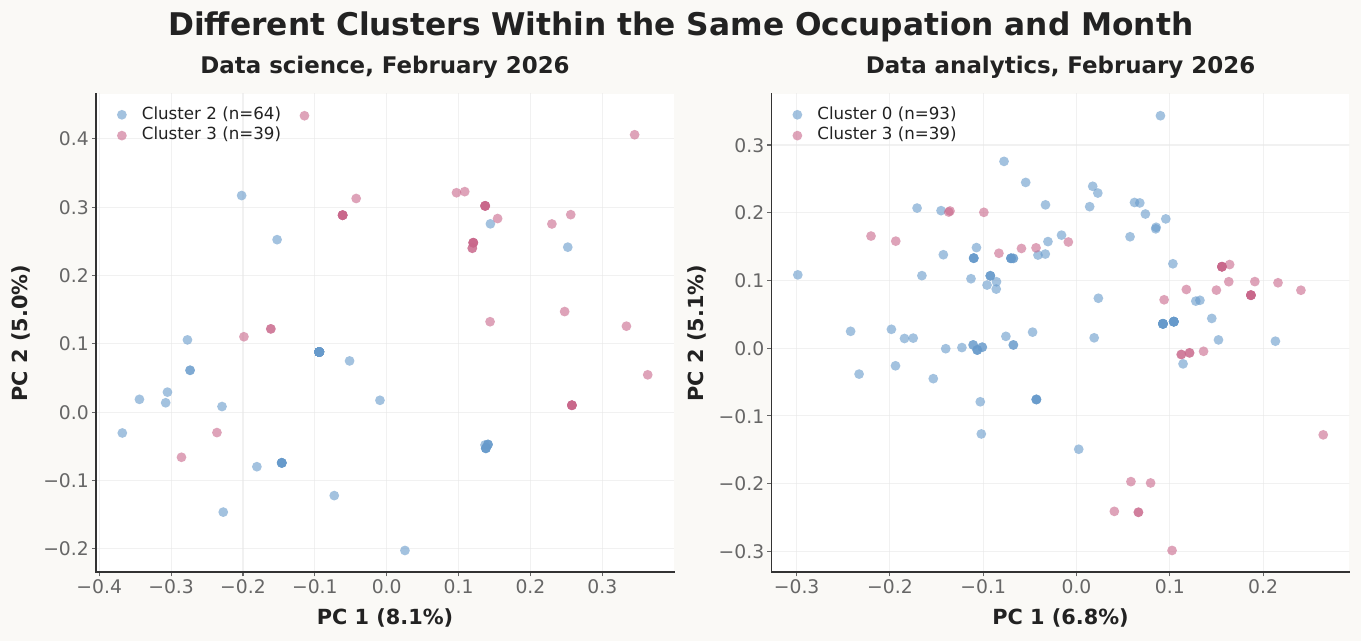}
\caption{A pair of PCA scatter plots comparing two Data Scientist Team context clusters on the left and two Data Analyst all-sections clusters on the right. Each point represents a February 2026 advertisement. In the Data Scientist panel, cluster 3 is concentrated toward the upper right and cluster 2 extends farther left and downward, with some overlap between the groups. The Data Analyst panel also shows overlap but places more cluster 3 advertisements on the right. Reading the descriptions connects these differences to engineering and manufacturing versus healthcare research in Data Scientist advertisements, and business operations versus healthcare and medical-data work in Data Analyst advertisements.}
\label{fig:occupation_specializations}\end{figure*}

The Role task descriptions also differ in what the applicant would build. Thermo Fisher describes dashboards and automated business workflows, whereas Kaiser describes translating clinical-analysis programs into Databricks notebooks and checking consistency with earlier results (\nolinkurl{4357587093}, \nolinkurl{4368762099}). Healthcare and technical roles also occur in cluster 0, so the clusters shown in Figure~\ref{fig:occupation_specializations} represent overlapping descriptions with different emphases.

\subsection{Business Analyst Advertisements Describe Different Uses of Reporting}
For Business Analyst Role task, MMD selects cluster 0 in April and September 2026, containing 82 and 40 advertisements from 22 and 28 company names respectively, with a score of 0.2335. Repeated Red Cross advertisements in April describe reporting and planning for training services. September includes Plante Moran's client consulting and U.S. Bank's predictive and scenario analysis (\nolinkurl{4400338944}, \nolinkurl{4460652621}, \nolinkurl{4464666363}). September also includes Vylor's AI-agent workflows, although April already includes AI-related work at Teledyne (\nolinkurl{4462547606}, \nolinkurl{4398870254}). These examples show differences in the advertised assignments without establishing that AI work first appears in September.

For Team context + Role task, April clusters 0 and 2 contain 31 and 70 advertisements from 18 and 12 companies respectively, with MMD 0.4343. Imperial Supplies describes maintaining reporting systems and the Red Cross describes analytics for operational planning (\nolinkurl{4377025865}, \nolinkurl{4400338944}). Two TikTok advertisements show a similar distinction, with one describing measurement frameworks and KPI dashboards and the other describing seller analysis for incentive and resource decisions (\nolinkurl{4366433763}, \nolinkurl{4382816786}). Both clusters therefore contain work on analytical systems and their use in business decisions.

\subsection{Qualification Clusters Differ in Entry Requirements}
For December 2025, Data Scientist Qualifications clusters 0 and 2 contain 278 and 118 advertisements from 97 and 11 company names respectively, with MMD 0.3842. Cluster 2 contains repeated Centene graduate-intern descriptions, which give graduate eligibility greater weight in that cluster. Cluster 0 includes more employers and entry routes.

For example, IBM's advertisement in cluster 0 accepts programming knowledge gained through coursework without full-time experience, whereas a Centene advertisement in cluster 2 requests completed undergraduate coursework and enrollment in a related graduate program (\nolinkurl{4344857454}, \nolinkurl{692ec3e823af210b6328b82f}). Graduate roles also appear in cluster 0 and some undergraduate routes appear in cluster 2. The distinction therefore reflects differences in eligibility together with repeated recruitment and wording.

\subsection{Monthly Comparisons Across All Occupations Also Reflect the Mix of Jobs}
The completed all-occupation comparison uses Team context + Role task cluster 2, containing 37 advertisements from 18 company names in July 2025 and 505 advertisements from 348 company names in September 2026, with MMD 0.1230. July includes nonprofit employment services, education support and program administration, and September includes a wider range of corporate analytics, finance and operations work.

For example, the International Institute of New England describes job-readiness training and administrative support, whereas Circle K describes portfolio strategy, site-selection analytics and executive reporting (\nolinkurl{68ac3a66501e80814a16c5eb}, \nolinkurl{4370458029}). September also includes operational data standardization and actuarial pricing. Because this comparison combines different occupations, its MMD can reflect changes in the mix of job types as well as differences within a job title. We therefore also examine each occupation separately and interpret this monthly example in light of its unequal sample sizes and collection periods.

\section{SAE Pilots, Evidence and Reproduction}\label{app:sae}

\subsection{What We Learned from the SAE Pilots}
Before the final comparison, we tested GPT-2 and Gemma language models, trained SAEs on sentence embeddings and examined changes in feature labeling, pooling, sparsity and domain interpretation. These experiments led us to check description format and verify proposed characteristics in held-out advertisements. We report them separately because they use different samples, definitions and selection criteria from the comparison of PERC, SAE and text-only inputs.

In the section-embedding pilot, we compare Top-k SAEs with dictionaries of 512 or 1,024 features, eight or sixteen active features and five seeds. The selected 1,024-feature configuration with $k=8$ has 79 stable features, of which 77 activate in at least 20 advertisements. A separate embedding-model pilot obtains 96 stable features with the same dictionary size and $k=8$, including 88 that activate in at least 20 advertisements. These counts describe features that can be aligned across runs, so they do not by themselves establish meaningful occupational characteristics. In the higher-capacity experiment, a 1,024-feature dictionary with $k=32$ gives 11 stable features and $k=64$ gives none under that experiment's feature-matching criteria.

Independent checks find some useful labels. Five of eight labels pass the work-domain confirmation criteria, including ``sports,'' ``transportation,'' ``manufacturing analytics,'' ``healthcare'' and ``pricing.'' A separate confirmation finds five of eight passing labels for ``sustainability,'' ``continuous improvement,'' ``financial close,'' ``data governance'' and ``recruiting.'' These results describe the selected features and advertisements, and the artifact catalog also records the unsuccessful and incomplete runs. They do not establish that the trained SAE performs better than the final pretrained SAE or PERC.

The complete Gemma experiment contains 126,759 observations across Role task, Team context, Qualifications and original descriptions, representing all 36,081 prepared advertisements. Its eight-cluster fits have reduced-space cosine silhouettes of 0.130, 0.139, 0.175 and 0.224 for those inputs respectively, and mean ARIs across initializations of 0.629, 0.905, 0.946 and 0.915. Because these silhouettes use the reduced representation, they cannot be ranked directly against PERC's original-vector silhouettes. The posting-period comparison in the main text also shows that greater separation can reflect description format and length rather than more distinct kinds of work.

\subsection{Which Hypotheses Enter Each Statistical Comparison}
The original Llama analysis contains 110 hypotheses from 15 paired comparisons. The comparison with PERC retains separate sets of tests for the original paired-advertisement study and the expanded monthly study. The expanded monthly set contains 1,867 tests, including 1,271 checks of earlier definitions, so it contains more tests than newly generated monthly hypotheses. Business Analyst comparisons have a separate set of 176 hypotheses, and Software Engineer comparisons have a separate set of 246 hypotheses, comprising 77 occupational and 169 monthly claims.

The combined results contain 1,171 hypotheses generated by GPT-6 Luna across 127 comparisons. We retain the original statistical corrections for each set rather than recomputing a correction across all recorded experiments, and reused advertisements do not count as independent replications.

A feature label describes a pattern in activating text, a hypothesis is GPT-6 Luna's claim about how a characteristic differs between occupations or months, and a verification judgment records whether GPT-6.1 Terra finds that characteristic in an advertisement. Several features can contribute to one hypothesis, and GPT-6 Luna can generate similar claims from PERC clusters, SAE features or text alone. We therefore distinguish feature activation, verified characteristics and cluster measurements throughout the results. Feature IDs are specific to each SAE checkpoint, so an ID in Gemma does not represent the same feature as that ID in Llama.

\subsection{Code and Result Locations}
The numbered experiment folders contain code outputs from recorded runs. For example, \nolinkurl{artifacts/perc_expanded_001/} contains the PERC configuration search and distribution analyses, and the prepared advertisements are stored separately. Table~\ref{tab:original11} connects each analysis to its code and recorded results.

\begin{table*}[!tb]\centering\small

\caption{The files used to reproduce the experiments in this paper. Each row represents an experiment, with the first column naming the analysis, Code giving its implementation path and Results giving the folder containing its recorded outputs. The rows separate PERC configuration searches and interpretation of selected job clusters from SAE extraction and the occupation comparisons, so each analysis can be traced to its code and results.}

\begin{tabularx}{\linewidth}{>{\raggedright\arraybackslash}X>{\raggedright\arraybackslash}X>{\raggedright\arraybackslash}X}
\toprule

Experiment & Code & Results \\

\midrule

Data preparation & \nolinkurl{src/latent_job_market/preparation.py} & \nolinkurl{artifacts/validity_preparation_001/} \\

Expanded PERC search & \nolinkurl{experiments/expanded_ablation/} & \nolinkurl{artifacts/hpc_migration_001/results/} \\

Interpretation of selected job clusters & \nolinkurl{src/latent_job_market/cluster_insights.py} & \nolinkurl{artifacts/perc_expanded_001/} \\

SAE pilots and full Gemma extraction & \nolinkurl{experiments/sae_pilot/} & \nolinkurl{artifacts/sae_full_dataset_001/} \\

Llama feature discovery and verification & \nolinkurl{experiments/sae_dataset_comparison/} & \nolinkurl{artifacts/sae_dataset_comparison_001/} \\

Monthly feature discovery and tracing & \nolinkurl{experiments/sae_dataset_comparison/monthly.py} & \nolinkurl{artifacts/sae_monthly_trace_001/} \\

PERC, SAE and text-only comparison & \nolinkurl{experiments/clustering_comparison/} & \nolinkurl{artifacts/clustering_sae_comparison_001/} \\

BA comparisons & \nolinkurl{experiments/ba_comparison/} & \nolinkurl{artifacts/ba_methods_001/} \\

SWE comparisons & \nolinkurl{experiments/swe_comparison/} & \nolinkurl{artifacts/swe_methods_001/} \\

\bottomrule\end{tabularx}\label{tab:original11}\end{table*}

The prepared dataset has SHA-256 checksum \nolinkurl{7035e4dde69db58b899edab5ba8531202fc962c1d216a37e042e3ce745bb0e81}. Each figure manifest lists its numerical source files, selected activation examples and checksums. The file \nolinkurl{paper/evidence/all_method_results.csv} contains the 1,171 hypotheses, \nolinkurl{artifact_catalog.csv} lists local result directories and \nolinkurl{code_catalog.csv} lists the active Python sources. The artifact catalog includes completed and incomplete runs so their records remain distinguishable.

For the reading and execution order, see \nolinkurl{paper/EXPERIMENTS.md}. The code archive contains scripts, notebook sources without saved outputs, dependency files and aggregate results. It excludes credentials, downloaded machine learning model weights and raw advertisement request payloads. We rebuild figures and the paper from completed outputs without making further paid API calls, although reproducing the underlying experimental results requires running those experiments again.

\section{Additional Method Details and Feature Maps}
\label{app:additional}
\subsection{Selecting Advertisements and Checking Hypotheses}
\label{app:verification}
To compare what PERC clusters, SAE features and text alone add to hypothesis generation, we give GPT-6 Luna the same selected job-ad sections under all three conditions. We use LinkedIn advertisements collected within seven days of posting and pair advertisements from different occupations within the same 2026 posting month, matching the formatting of their descriptions. Each pair must have a section character-length ratio no greater than 1.25. We first select the largest possible set of eligible pairs and then minimize their differences in log character length, allowing at most one advertisement per company in each comparison group. Before verification, we remove test-company sections or full descriptions that duplicate the text used to generate hypotheses after normalization.

For the PERC condition, we refit the selected configuration on the descriptions used to generate hypotheses. In the Business Analyst and Software Engineer comparisons, we pool these descriptions by section when an individual sample cannot support the selected dimensionality reduction. We do not apply the 30-advertisement threshold used when Terra compares all descriptions in selected groups here because this experiment gives the same selected sections to all three conditions.

After GPT-6 Luna generates the claims, GPT-6.1 Terra receives the complete held-out sections, anonymous text identifiers and the unchanged hypothesis definitions. We omit separate occupation, company, month, method and activation information, although employer details can remain in the descriptions themselves. To mark a characteristic as present, GPT-6.1 Terra must provide an exact quotation from the advertisement. We include unclear judgments in the total number of advertisements checked and repeat the calculation with unclear judgments counted as present to examine their effect on the result.

To calculate confidence intervals, we resample companies connected through the selected advertisement pairs together, keeping repeated links between companies in the same resampling unit. We also use a sign-flip test that swaps the two group labels within each connected component. We count a difference as statistically supported when its adjusted p-value is at most 0.05, its absolute frequency difference exceeds 0.01 and its direction agrees with GPT-6 Luna's claim. We calculate adjusted p-values using both Benjamini--Hochberg (BH) and Benjamini--Yekutieli (BY) correction. The test assumes that swapping the two group labels within each connected component is valid under the null hypothesis, and the observational comparisons do not establish a causal change in hiring.

We apply these corrections separately to the Data Scientist--Data Analyst and Machine Learning Engineer--Data Scientist study, the expanded monthly study, the Business Analyst study and the Software Engineer study. The expanded monthly set contains 1,867 tests, including checks of previously generated definitions. No monthly or Business Analyst hypothesis passes either correction. We also keep the occupational results from the Data Scientist and Software Engineer comparisons separate in the main tables because they belong to different sets of tests.

The recorded API cost of feature labeling and hypothesis generation is \$17.50 with SAE inputs, \$0.28 with PERC inputs and \$0.31 with text alone across 86 Data Scientist, Data Analyst and Machine Learning Engineer comparisons. The corresponding costs are \$3.64, \$0.042 and \$0.041 for Business Analyst comparisons, and \$5.231, \$0.080 and \$0.092 for Software Engineer comparisons. These amounts exclude inference with the language model used to extract SAE features, cached embeddings and shared verification. Although we allow the same maximum generation budget for each condition, the actual spending differs.

\subsection{Additional Verification Tables}
The following tables give the hypothesis counts, API costs and advertisement-level judgments behind the occupation and monthly comparisons. Each count of generated hypotheses describes a claim that GPT-6 Luna proposes, and each verification count describes advertisements in which GPT-6.1 Terra finds the characteristic described in that claim. We retain this distinction because generating more claims does not establish that more of them recur in held-out advertisements.

\begin{table*}[!tb]\centering\small

\caption{The occupational hypothesis counts and generation API costs for three GPT-6 Luna input conditions. We compare PERC clusters, SAE features and text alone. Occupational proposals counts GPT-6 Luna's claims across six section comparisons involving Data Scientist, Data Analyst and Machine Learning Engineer advertisements, and Generation cost includes all 86 occupational and monthly comparisons. Using SAE features costs \$17.50 including feature labeling, compared with \$0.28 using PERC clusters and \$0.31 using the text alone. These costs exclude shared verification and local computation.}

\begin{tabularx}{\linewidth}{>{\raggedright\arraybackslash}X>{\raggedright\arraybackslash}X>{\raggedright\arraybackslash}X}
\toprule

Method & Occupational proposals & Generation cost, all 86 comparisons \\

\midrule

PERC + GPT-6 Luna & 20 & \$0.28 \\

SAE + GPT-6 Luna & 26 & \$17.50 \\

GPT-6 Luna & 22 & \$0.31 \\

\bottomrule\end{tabularx}\label{tab:original1}\end{table*}

\begin{table*}[!tb]\centering\small

\caption{The number of hypotheses GPT-6 Luna generates for Software Engineer comparisons under each input condition. Each row represents one condition, the Method column names that condition, and Occupational proposals and Monthly proposals count claims across six occupational and 18 monthly section comparisons, respectively. With SAE features, GPT-6 Luna generates 82 monthly claims compared with 42 using PERC clusters and 45 using text alone. None of these monthly differences passes the statistical criteria described in Appendix~\ref{app:verification}.}

\begin{tabularx}{\linewidth}{>{\raggedright\arraybackslash}X>{\raggedright\arraybackslash}X>{\raggedright\arraybackslash}X}
\toprule

Method & Occupational proposals & Monthly proposals \\

\midrule

PERC + GPT-6 Luna & 29 & 42 \\

SAE + GPT-6 Luna & 24 & 82 \\

GPT-6 Luna & 24 & 45 \\

\bottomrule\end{tabularx}\label{tab:original2}\end{table*}

\begin{table*}[!tb]\centering\small

\caption{Qualification characteristics checked by GPT-6.1 Terra in held-out Data Scientist and Data Analyst advertisements. Each row represents one characteristic, with Method naming the input condition used by GPT-6 Luna and Qualification stating the characteristic to check. The Data Scientist and Data Analyst columns give the number of advertisements containing that characteristic out of 93 checked for each occupation. ``Machine learning and statistical methods'' and ``machine learning platform and deployment tools'' appear more often in Data Scientist advertisements, and ``Spreadsheet skills'' and ``business intelligence tools'' appear more often in Data Analyst advertisements. GPT-6 Luna produces different definitions under the two input conditions, so the spreadsheet rows describe different requirements.}

\begin{tabularx}{\linewidth}{>{\raggedright\arraybackslash}p{.8in}>{\raggedright\arraybackslash}Xrr}
\toprule

Method & Qualification & DS & DA \\

\midrule

Clustering & ``Machine learning platform and deployment tools'' & 35/93 & 15/93 \\

Clustering & ``Advanced Excel analysis'' & 2/93 & 18/93 \\

SAE & ``Machine learning and statistical methods'' & 56/93 & 17/93 \\

SAE & ``Programming languages and libraries'' & 82/93 & 62/93 \\

SAE & ``Excel-based data work'' & 3/93 & 27/93 \\

SAE & ``Business intelligence and visualization tools'' & 26/93 & 53/93 \\

\bottomrule\end{tabularx}\label{tab:original3}\end{table*}

\begin{table*}[!tb]\centering\small

\caption{Selected monthly characteristics checked by GPT-6.1 Terra in advertisements from held-out companies. Each row represents one characteristic in one month comparison. The columns name the occupation and months, the characteristic and job-description section, the input condition used by GPT-6 Luna, and the counts for the earlier and later month. Each fraction represents advertisements containing the characteristic out of all advertisements checked. In Data Scientist advertisements, ``Data preparation'' falls from 9/13 in December to 4/13 in January, and ``Pipeline construction'' rises from 2/15 in January to 6/15 in February. None of these monthly differences passes the statistical criteria described in Appendix~\ref{app:verification}.}

\begin{tabularx}{\linewidth}{>{\hsize=1.3\hsize\raggedright\arraybackslash}X>{\hsize=2.0\hsize\raggedright\arraybackslash}X>{\hsize=0.7\hsize\raggedright\arraybackslash}X>{\hsize=0.5\hsize\raggedright\arraybackslash}X>{\hsize=0.5\hsize\raggedright\arraybackslash}X}
\toprule

Occupation and months & Property and section & Method & Earlier & Later \\

\midrule

DA, Feb--Sep 2026 & ``Data operations'', qualifications & SAE & 4/39 & 12/39 \\

DS, Dec 2025--Jan 2026 & ``Data preparation'', tasks & SAE & 9/13 & 4/13 \\

DS, Jan--Feb 2026 & ``Pipeline construction'', tasks & SAE & 2/15 & 6/15 \\

DS, Mar--Apr 2026 & ``Designing and running experiments'', tasks & SAE & 1/17 & 5/17 \\

DS, Mar--Apr 2026 & ``Causal inference or experiments'', qualifications & Clustering & 0/17 & 3/17 \\

DS, Aug--Sep 2026 & ``Numeric experience thresholds'', qualifications & SAE & 5/12 & 2/12 \\

MLE, Aug--Sep 2026 & ``AI and data quality checks'', tasks & SAE & 2/6 & 5/6 \\

MLE, Aug--Sep 2026 & ``Repeatable AI regression tests'', tasks & Clustering & 1/6 & 0/6 \\

BA, Feb--Sep 2026 & ``AI-enabled work'', tasks & Clustering & 0/5 & 2/5 \\

BA, Feb--Sep 2026 & ``Enterprise system requirements'', tasks & SAE & 2/5 & 0/5 \\

\bottomrule\end{tabularx}\label{tab:original4}\end{table*}

\subsection{SAE Extraction and Feature Clustering}
\label{app:featureclusters}

To extract features, we run Llama-3.1-8B-Instruct in bfloat16 without quantization and the Goodfire SAE in float32. At layer 19, Llama-3.1-8B-Instruct produces a 4,096-dimensional residual state for each token. The SAE's linear encoder and ReLU transform this state into 65,536 feature activations, and its linear decoder reconstructs the residual state. We compare the converted SAE weights with the downloaded checkpoint and check the extracted residuals against a complete forward pass through Llama-3.1-8B-Instruct.

We count a feature as present when at least two source tokens have positive activation, excluding beginning-of-sequence and padding tokens. We also repeat the presence calculation using one token as a sensitivity check. When labeling a feature, GPT-6 Luna reads up to ten sections where it activates and ten sections where no tokens activate it. Sections with only one activating token do not count as inactive examples. GPT-6 Luna then combines related labels into at most ten hypotheses per comparison, with at least two supporting source descriptions per hypothesis.

To examine how features relate to one another, we combine unique job-ad sections used for hypothesis generation across the Data Scientist, Data Analyst, Machine Learning Engineer, Business Analyst and Software Engineer comparisons. We weight each binary activation by inverse feature frequency, normalize each advertisement's activation vector and fit truncated singular-value decomposition (SVD) with 50 components. Each feature is represented by its SVD loading scaled by the singular values and then normalized. We retain features that activate in 1\% to 99\% of the section sample, so the map compares features observed across multiple advertisements without including those active almost everywhere.

We apply K-means to the 50-dimensional feature vectors and compare four, six, eight, ten and twelve clusters. For each section, we calculate cosine silhouette on the same sample of 2,000 features and select the cluster count with the highest score. This gives twelve clusters for Team context, eight for Role task and ten for Qualifications, with sampled silhouettes of 0.071, 0.097 and 0.123 respectively. ARI between two initializations is 0.669, 0.863 and 0.816 for those sections. We then use PCA to display the features in two dimensions, keeping their coordinates unchanged across the selected PERC comparisons.

\begin{figure*}[!tb]\centering
\includegraphics[width=\textwidth,height=.85\textheight,keepaspectratio]{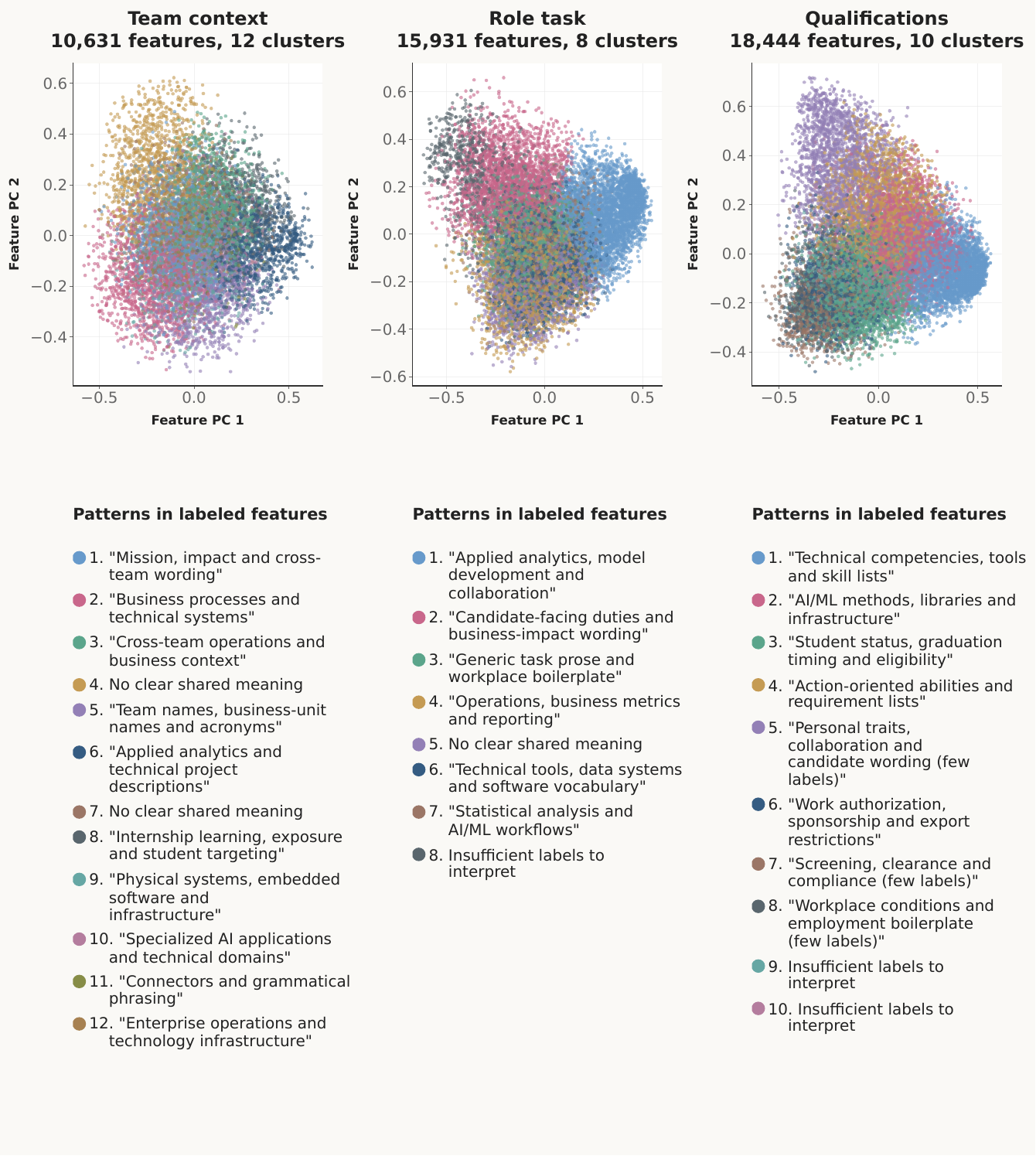}
\caption{A series of PCA scatter plots of SAE features, where each point represents one feature and each color represents a K-means cluster. The descriptions below each plot represent the patterns found among labeled features in each cluster. Role task and Qualifications show broad colored regions, whereas Team context has more colors mixed near its center. In Role task, the blue region on the right includes features related to ``applied analytics,'' ``developing machine learning models'' and ``collaboration'', and the gold region below includes ``Operations, business metrics and reporting''. In Qualifications, the right-hand blue region includes ``Technical competencies, tools and skill lists'' and the lower green region includes ``Student status, graduation timing and eligibility''. Other clusters contain grammar and recruiting language, showing that SAE features group writing patterns alongside work content.}\label{fig:llama_sae_feature_clusters}
\end{figure*}

GPT-6 Luna has labeled 10,581 of the 45,006 features represented in the section maps through the preceding occupation and monthly comparisons. Features appearing in more than one job-description section are counted separately for each section. To describe each feature cluster, we inspect labels near its center and their source quotations. Because these labels were selected for specific comparisons, their coverage varies across feature clusters. Some clusters contain grammatical or recruiting patterns, and we leave clusters with too few informative labels uninterpreted.

For the analysis within selected PERC clusters, we retain the membership of three completed MMD comparisons: Data Scientist Team context clusters 2 and 3 in February 2026 with 64 and 39 advertisements, Business Analyst Role task cluster 0 in April and September 2026 with 82 and 40 advertisements, and Data Scientist Qualifications clusters 0 and 2 in December 2025 with 278 and 118 advertisements. We extract each advertisement's complete selected section and reuse the activations when its text is identical to a section already processed. Repeated descriptions still count as separate postings in the activation percentages. We use the existing feature clusters to describe these PERC comparisons without fitting new groups or generating new claims about the wider job market.

The health-sector feature shown in Figure~\ref{fig:sae_selected_cluster_profiles} received different labels across earlier comparisons, including unclear wording and ``operational analytics.'' Its activation difference shows that the representation responds differently to the two PERC samples, although this feature has not been validated as a healthcare classifier.

Together, the three comparisons contain 621 job-description sections across the selected postings, containing 354 distinct texts. Each section matches the text used in its PERC comparison and is processed without truncation. We run Llama-3.1-8B-Instruct once per distinct text and then count repeated descriptions at the posting level when calculating activation rates.

For feature $f$ in PERC cluster sample $G$ the activation rate is defined as
\begin{equation}
p_{Gf}=\frac{1}{|G|}\sum_{a\in G}\mathbf{1}[c_{af}\geq2],
\end{equation}
where $c_{af}$ is the number of source tokens in advertisement $a$ that activate feature $f$, and $|G|$ is the number of advertisements in the sample. For samples $A$ and $B$, the maps display $p_{Af}$, $p_{Bf}$ and $p_{Bf}-p_{Af}$ at the same feature coordinates. We multiply these rates by 100 when displaying percentages. To summarize a feature cluster, we average the rates of its member features, so this average describes feature activation rather than the proportion of advertisements containing a characteristic verified by GPT-6.1 Terra.

\begin{figure*}[!tb]\centering
\includegraphics[width=\textwidth,height=.85\textheight,keepaspectratio]{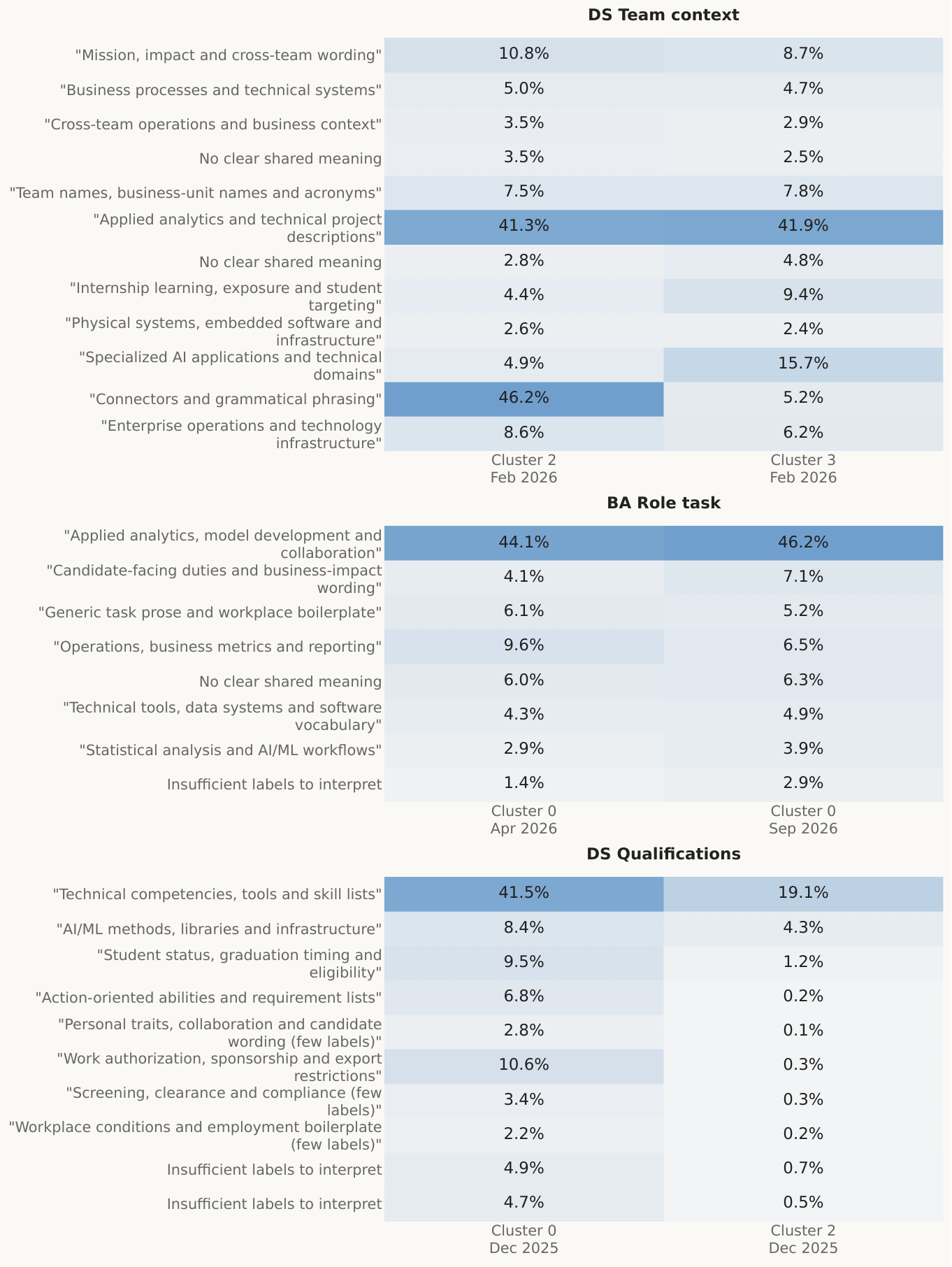}
\caption{A set of heatmaps comparing average activation rates for groups of related SAE features across selected PERC clusters or months. Darker blue indicates higher rates. Data Scientist Team context shows a large difference in ``Connectors and grammatical phrasing'', whose average activation falls from 46.2\% in cluster 2 to 5.2\% in cluster 3, whereas ``Applied analytics and technical project descriptions'' remains near 41\% in both clusters. Business Analyst task averages differ by at most 3.1\% between April and September. Every Data Scientist Qualifications group activates less often in cluster 2, including ``Technical competencies, tools and skill lists'' at 19.1\% versus 41.5\% in cluster 0.}\label{fig:sae_selected_feature_groups}
\end{figure*}

\end{document}